\documentclass[]{style}

\usepackage[toc,page,header]{appendix}
\usepackage{enumitem}
\usepackage{hyperref}

\usepackage{natbib}

\usepackage{CJKutf8}

\usepackage{xargs}  

\usepackage{todonotes}  

\usepackage{multirow}

\usepackage{cleveref}

\usepackage{amsmath}
\usepackage{dsfont}

\usepackage{svg}

\usepackage{mathrsfs}
\usepackage{adjustbox}
\usepackage{multirow}
\usepackage{multicol}
\usepackage{tcolorbox}
\usepackage{changepage}
\usepackage{enumitem}
\usepackage{graphicx}
\usepackage{amssymb}
\usepackage{xcolor}
\usepackage{float}
\usepackage{multirow}
\usepackage{threeparttable}
\usepackage{graphicx}
\usepackage{subcaption}
\usepackage{algorithm}
\usepackage{algpseudocode}
\usepackage{wrapfig}
\usepackage[table]{xcolor}  %
\usepackage{colortbl}       %
\usepackage{tabularx} %
\usepackage{makecell} %
\usepackage[dvipsnames]{xcolor}

\newcolumntype{g}{>{\columncolor{gray!10}}c} %

\definecolor{catgray}{gray}{0.9}
\definecolor{skyblue}{rgb}{0.53,0.81,0.92} %

\colorlet{skyblue!30}{skyblue!30!white} %

\definecolor{customblue}{RGB}{70,130,180}  %

\newtcolorbox{evolbox}[2][]{%
  enhanced,
  colframe=customblue,
  colback=white,
  coltitle=white,
  rounded corners,
  boxrule=1pt,
  titlerule=0pt,
  toptitle=1mm,
  bottomtitle=1mm,
  fonttitle=\bfseries,
  width=#2\textwidth, %
  #1
}

\usepackage{minitoc}
\usepackage{url}

\PassOptionsToPackage{table,xcdraw}{xcolor}
\usepackage{titletoc}
\usepackage{placeins}
\usepackage{pifont}

\definecolor{RowBlue}{HTML}{E9F2FB}
\definecolor{RowRed}{HTML}{F9EAEA}
\definecolor{Top1}{HTML}{50DB4B} %
\definecolor{Top2}{HTML}{A5FFA2} %
\definecolor{Top3}{HTML}{D9FFD9} %
\definecolor{Sub1}{HTML}{EAB8B8}
\definecolor{Sub2}{HTML}{E4E4E4}

\renewcommand{\emph}[1]{\textit{#1}}

\title{Open-HumanWAM: An Open Recipe for Scaling Human and Robot Experience for Embodied Foundation Models}

\author{
  Juncheng~Ma\texorpdfstring{$^{*}$}{} \hspace{0.05cm}
  Jiaxin~Li\texorpdfstring{$^{*}$}{} \hspace{0.1cm}
  Jiankai~Tu\texorpdfstring{$^{*}$}{} \hspace{0.1cm}
  Bo~Liang\texorpdfstring{$^{*}$}{}{} \hspace{0.1cm}
  Guodong~Lv\texorpdfstring{}{} \hspace{0.1cm}
  Yufan~Deng\texorpdfstring{}{} \hspace{0.1cm}
  Junzhe~Tang\texorpdfstring{}{} \hspace{0.1cm}
  Liangzhu~Yang\texorpdfstring{}{} \hspace{0.1cm}
  Chen~Bian\texorpdfstring{}{} \hspace{0.1cm}
  Shukai~Gong\texorpdfstring{}{} \hspace{0.1cm}
  Ye Huang\texorpdfstring{}{} \hspace{0.1cm}
  Yintianrun~Zhang\texorpdfstring{}{} \hspace{0.1cm}
  Zicong~Cao\texorpdfstring{}{} \hspace{0.1cm}
  Baohao~Zhu\texorpdfstring{}{} \hspace{0.1cm}
  Jianxin~Bi\texorpdfstring{}{} \hspace{0.1cm}
  Tengfei~Yan\texorpdfstring{}{} \hspace{0.1cm}
  Yinyuan~Zhang\texorpdfstring{}{} \hspace{0.1cm}
  Kewei~Zhang\texorpdfstring{}{} \hspace{0.1cm}
  Xinglong~Chen\texorpdfstring{}{} \hspace{0.1cm}
  Kaiqi~Chen\texorpdfstring{}{} \hspace{0.1cm}
  Duomin~Wang\texorpdfstring{}{} \hspace{0.1cm}
  Xiaotian~Tang\texorpdfstring{}{} \hspace{0.1cm}
  Liang~Tan\texorpdfstring{}{} \hspace{0.1cm}
  Li~Li\texorpdfstring{}{} \hspace{0.1cm}
  Yuan~Xu\texorpdfstring{}{} \hspace{0.1cm}
  Haosen~Yang\texorpdfstring{}{} \hspace{0.1cm}
  Yuqi~Wang\texorpdfstring{}{} \hspace{0.1cm}
  Jingfei~Xia\texorpdfstring{}{} \hspace{0.1cm}
  Songyan~Huang \hspace{0.1cm}
  Zhiyang~Dou \hspace{0.1cm}
  Zhen~Dong \hspace{0.1cm}
  Enze~Xie \hspace{0.1cm}
  Wojciech~Matusik
  \hspace{0.1cm}
  Tat-Seng~Chua \hspace{0.1cm}
  Daquan~Zhou\texorpdfstring{$^{\dagger}$}{}
}

\affiliation[]{PKU, NUS, MIT, UCSB, NVIDIA}
\contribution[*]{Equal Contribution}
\contribution[\dagger]{Corresponding Author}

\usepackage{xcolor}
\usepackage{xparse}

\NewDocumentCommand{\answerYes}{g}{%
  \textcolor{blue}{[Yes\IfValueT{#1}{ #1}]}%
}
\NewDocumentCommand{\answerNo}{g}{%
  \textcolor{orange}{[No\IfValueT{#1}{ #1}]}%
}
\NewDocumentCommand{\answerNA}{g}{%
  \textcolor{gray}{[N/A\IfValueT{#1}{ #1}]}%
}
\NewDocumentCommand{\answerTODO}{g}{%
  \textcolor{red}{\bfseries [TODO\IfValueT{#1}{ #1}]}%
}
\NewDocumentCommand{\justificationTODO}{g}{%
  \textcolor{red}{\bfseries [TODO\IfValueT{#1}{ #1}]}%
}

\usepackage{xcolor}
\usepackage[normalem]{ulem}
\newif\iftrackchanges
\trackchangestrue %

\definecolor{PekingRed}{RGB}{178,31,45}

\abstract{
Egocentric human video offers a scalable source of manipulation priors for robot learning, yet translating raw data scale into generalizable robot policies demands rigorous data curation and training recipes. We introduce \textbf{Open-HumanWAM}, the first fully open recipe for scaling egocentric pretraining to 50,000 hours, anchored by \textbf{Open-HumanWAM-50K}---a curated egocentric video corpus densely annotated with hand motion and natural language. Leveraging this corpus, we systematically investigate pure human pretraining via closed-loop simulation evaluations and uncover an empirical log-linear scaling law linking pretraining data hours to downstream robot success rates. Grounded in these findings, we develop \textbf{HumanWAM-Vanilla}, a world--action model pretrained exclusively on 50k hours of egocentric human video. Upon robot post-training, HumanWAM-Vanilla reaches 92.08\% success on RoboTwin 2.0 Clean2Clean and 68.88\% on Clean2Random, alongside a 14.17\% mean success rate and a 19.00 mean progress score on RoboDojo. To push the capability frontier further, we explore joint human--robot pretraining with \textbf{HumanWAM-Goal}, a visual-goal-conditioned world--action model. Hybrid data scaling unlocks broad zero-shot generalization across both simulation and physical benchmarks, propelling post-trained RoboTwin 2.0 performance to new state-of-the-art levels: 92.70\% on Clean2Clean and a substantial leap to 84.60\% on Clean2Random. On physical robots, HumanWAM-Goal further exhibits emergent embodied in-context learning, dynamically adapting its manipulation behavior to unseen visual goals without parameter updates. To support reproducible scaling research in embodied AI, we open-source the Open-HumanWAM-50K corpus, data curation and pretraining pipelines, alongside the pretrained and post-trained HumanWAM checkpoints, inference code, and simulation evaluation suites.
}

\checkdata[
\raisebox{-0.2em}{\includegraphics[width=0.025\linewidth]{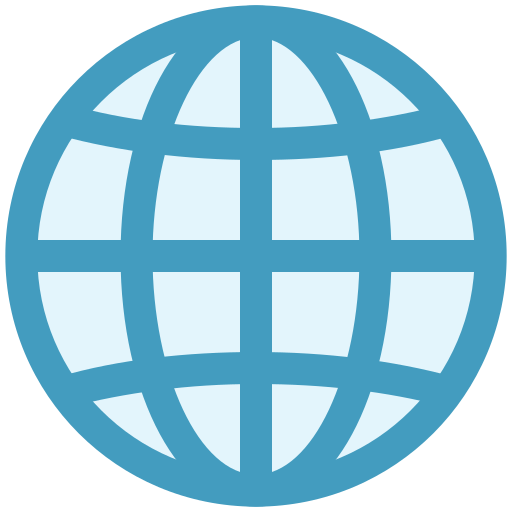}}~~Project Page]{\href{https://dagroup-pku.github.io/open-humanwam/}{\texttt{https://dagroup-pku.github.io/open-humanwam/}}
\\[-1.5ex]}

\checkdata[
\raisebox{-0.2em}{\includegraphics[width=0.025\linewidth]{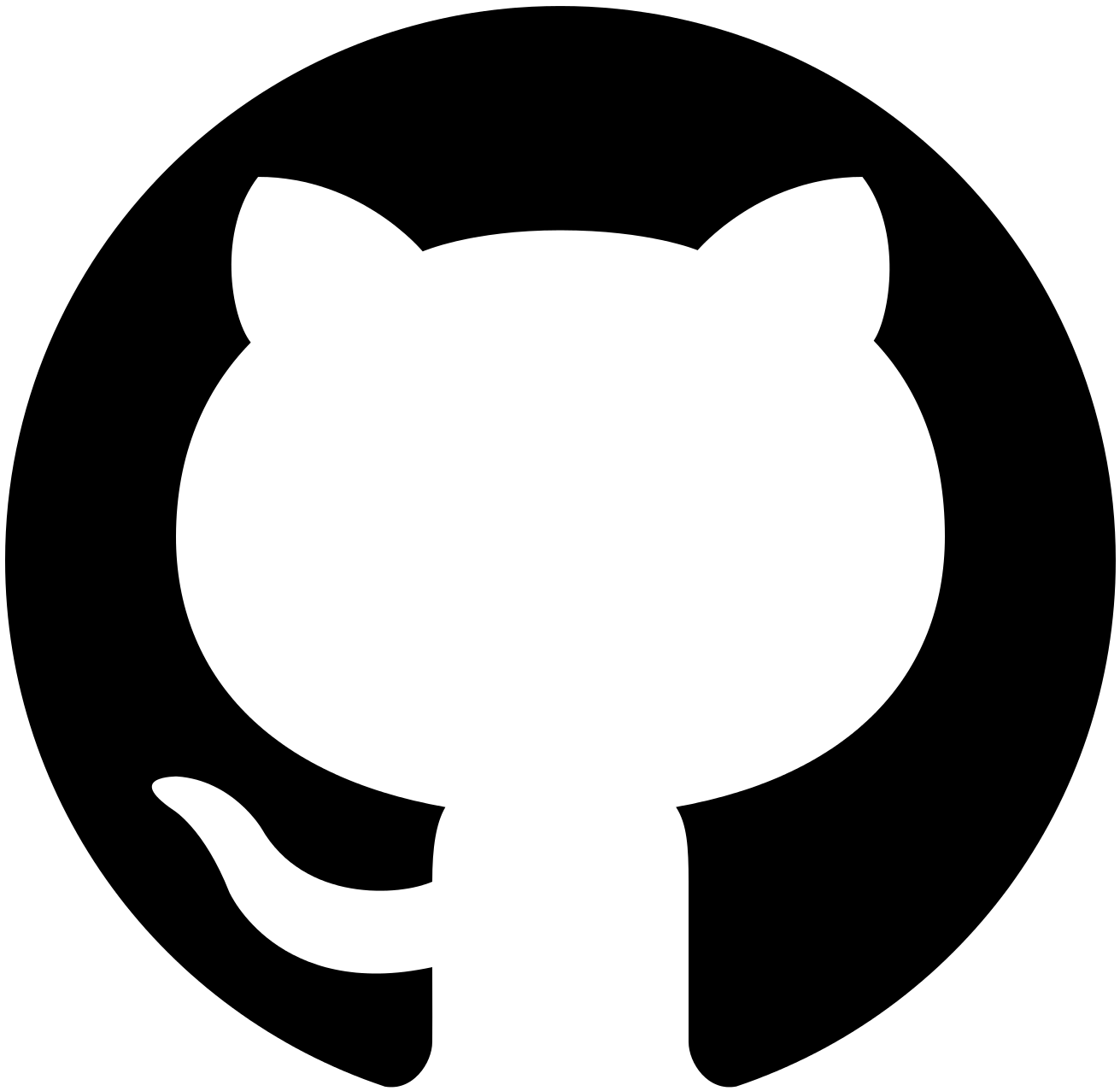}}~~GitHub Repo]{\href{https://github.com/DAGroup-PKU/HumanNet/tree/Open-HumanWAM}{\texttt{https://github.com/DAGroup-PKU/HumanNet/tree/Open-HumanWAM}}
\\[-1.7ex]}

\checkdata[
\raisebox{-0.3em}{\includegraphics[width=0.026\linewidth]{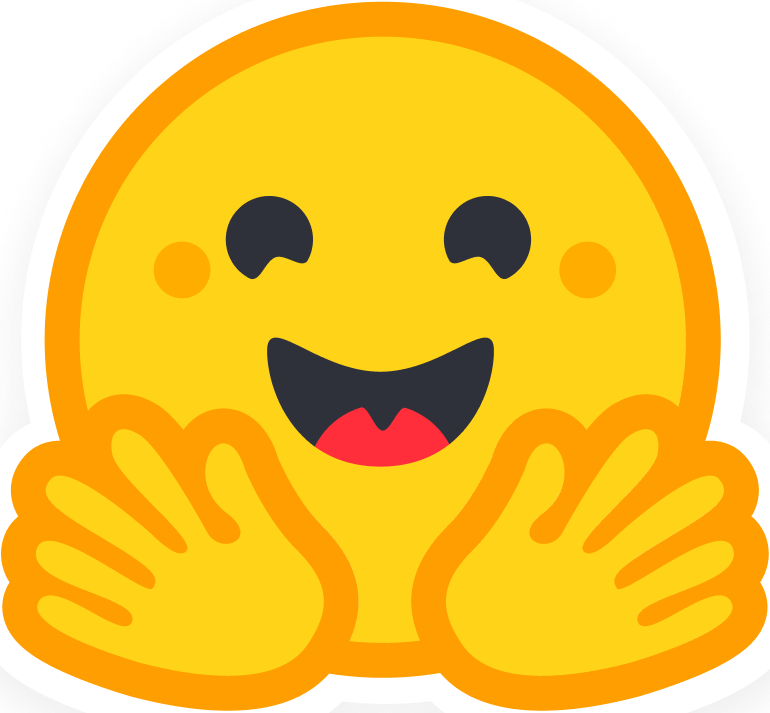}}~~HuggingFace Models]{\href{https://huggingface.co/collections/DAGroup-PKU/open-humanwam}{\texttt{https://huggingface.co/collections/DAGroup-PKU/open-humanwam}}
\\[-1.7ex]}

\checkdata[
\raisebox{-0.3em}{\includegraphics[width=0.026\linewidth]{figs/icons/hf.png}}~~HuggingFace Datasets]{\href{https://huggingface.co/datasets/DAGroup-PKU/Open-HumanWAM-50K}{\texttt{https://huggingface.co/datasets/DAGroup-PKU/Open-HumanWAM-50K}}
\\[-1.7ex]}

\begin{document}
\maketitle

\clearpage
\setcounter{tocdepth}{2}
\setcounter{secnumdepth}{3}
\tableofcontents
\clearpage

\section{Introduction}
\label{sec:intro}
\begin{figure}[h]
  \centering
  \includegraphics[width=\linewidth]{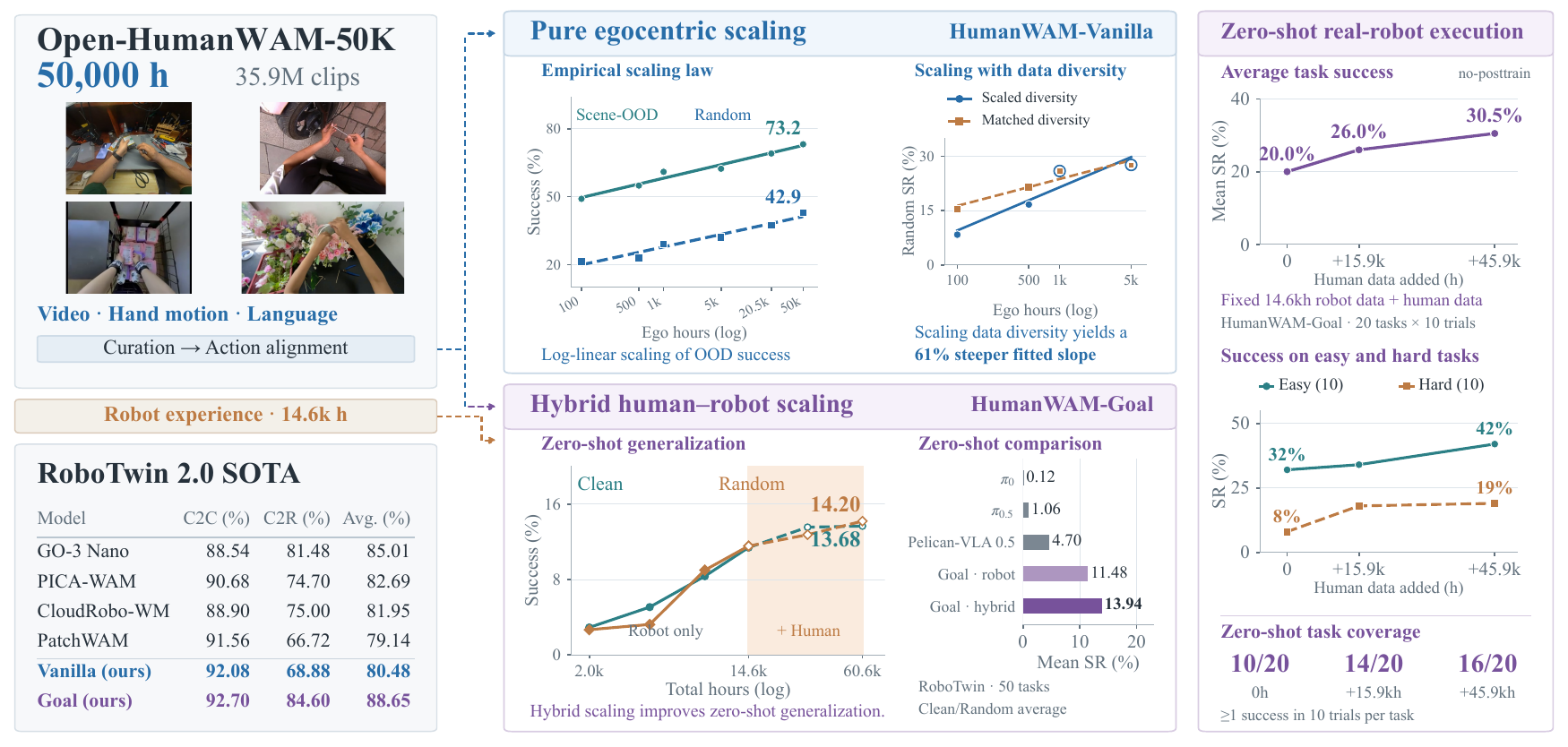}
  \caption{\textbf{Overview of Open-HumanWAM.} Using a curated 50,000-hour egocentric corpus, we show that scaling human data improves robot generalization after post-training, with greater task and scene diversity yielding steeper scaling curves. We then extend this pretraining approach to combine human and robot data, finding that hybrid scaling improves zero-shot success in simulation and across 20 real-world tasks.}
  \label{fig:teaser}
\end{figure}
Learning from human manipulation offers a scalable path toward general-purpose robot intelligence~\cite{egomimic,egoscale,humanscale}. Robot foundation models require diverse manipulation experience to acquire skills that transfer across tasks and environments, but collecting this experience through teleoperation is costly~\cite{droid,egomimic}. Egocentric videos paired with hand-motion and language annotations offer a rich record of human interactions with objects across diverse scenes and tasks~\cite{egodex,egoverse}. However, developing effective egocentric pretraining recipes that consistently improve downstream robot performance as data scale remains under investigation.

Previous work has demonstrated the value of human experience for robot learning. EgoScale~\cite{egoscale} shows that large-scale human motion pretraining improves dexterous robot manipulation, while Being-H0.8~\cite{beingh08} extends egocentric pretraining to joint visual and tactile world modeling. HumanScale~\cite{humanscale} further shows that, after robot post-training, validation action loss on held-out real-robot trajectories decreases log-linearly with egocentric pretraining hours. Dyna-2~\cite{dyna2} extends egocentric pretraining to one million hours and reports a scaling law in zero-shot robot action prediction, with corresponding improvements in robot task performance after post-training. These studies establish the empirical value of egocentric scaling, motivating further analysis of data diversity and the robot supervision required for downstream adaptation.

We present \textbf{Open-HumanWAM}, a fully open recipe for scaling egocentric pretraining for robot learning. Within a unified world--action model framework, we explore pretraining exclusively on 100 to 50{,}000 hours of human egocentric data, using a common robot post-training protocol and closed-loop simulation for controlled and reproducible comparisons of downstream robot task success. With curated data and a unified action representation, we observe an empirical log-linear scaling law relating pretraining hours to out-of-distribution success rates across two different attention configurations. Beyond duration, data diversity is a key scaling dimension, as expanding task and scene coverage alongside duration produces steeper scaling curves than increasing duration at fixed coverage. Larger-scale egocentric pretraining also improves the sample efficiency of robot adaptation, yielding higher success rates at matched demonstration budgets.

Following this recipe, we develop \textbf{HumanWAM-Vanilla}, a vanilla world--action model pretrained exclusively on 50{,}000 hours of egocentric human data. After robot post-training, HumanWAM-Vanilla achieves 92.08\% success on RoboTwin~2.0~\cite{robotwin2} Clean2Clean and 68.88\% on Clean2Random, together with a mean success rate of 14.17\% and a mean progress score of 19.00 on RoboDojo~\cite{robodojo}. The improvements extend to physical robots across multi-step, deformable-object, and long-horizon manipulation tasks. Under changes in tablecloth and lighting, mean success across four tasks increases from 2.5\% without egocentric pretraining to 67.5\% with HumanWAM-Vanilla, demonstrating stronger generalization.

To further push the model capability boundary, we extend our exploration to joint human-robot pretraining and propose \textbf{HumanWAM-Goal}, a world--action model with visual-goal conditioning. Human data provide broad manipulation experience, while robot data ground this experience in embodiment-specific perception and control~\cite{egoscale,aceego0}. Visual goals specify desired object configurations and spatial relationships, providing task context for zero-shot execution. 
With the specially designed unified action representation, HumanWAM-Goal architecture, and the hybrid training recipe, the co-trained model supports out-of-the-box execution without parameter updates during deployment.
We evaluate data scaling by expanding the robot corpus and the human data. The performance scaling trend persists in this hybrid data setting. On physical robots, adding 45.9k hours of human data to a fixed 14.6k-hour robot corpus increases mean zero-shot success across 20 tasks from 20.0\% to 30.5\%. On RoboTwin~2.0, HumanWAM-Goal achieves 13.94\% mean zero-shot success rate. After post-training, it reaches 92.70\% on Clean2Clean and 84.60\% on Clean2Random, yielding a state-of-the-art average success rate of 88.65\% on RoboTwin~2.0.

HumanWAM-Goal exhibits an emergent capacity for embodied in-context learning through visual goals. In real-robot experiments, the same policy adapts its execution to different desired outcomes specified by goal images without parameter updates. This capability is already present after joint pretraining and is retained when the model is post-trained for more complex, multi-step manipulation. Post-trained policies also follow goals depicted in egocentric human images, indicating that task-relevant spatial relationships can guide robot behavior across differences in embodiment and visual appearance.

To support reproducible scaling research in embodied AI, we open-source the Open-HumanWAM-50K corpus, data curation and pretraining pipelines, alongside the pretrained and post-trained HumanWAM checkpoints, inference code, and simulation evaluation suites. Open-HumanWAM-50K is a curated and aligned 50,000-hour egocentric pretraining corpus with hand-motion and language annotations, and our experiments demonstrate consistent gains in downstream robot generalization as pretraining on this corpus scales, with out-of-distribution success rates following an empirical log-linear scaling law.

The main contributions of this work are as follows:
\begin{enumerate}[label=\arabic*., leftmargin=1.8em, labelsep=0.5em, itemsep=5pt, topsep=4pt, parsep=0pt, before=\setlength{\emergencystretch}{2em}]
  \item \textbf{A systematic study of egocentric scaling for robot generalization.} We present Open-HumanWAM, a fully open recipe for scaling pure egocentric pretraining from 100 to 50k hours. We identify an empirical log-linear scaling law in downstream robot task success rates and systematically analyze the benefits of egocentric scaling for robot generalization and adaptation. Building on this recipe, we develop HumanWAM-Vanilla, which achieves success rates of 92.08\% on RoboTwin 2.0 Clean2Clean and 68.88\% on Clean2Random after robot post-training, alongside competitive performance on RoboDojo.

  \item \textbf{Emerging robot capabilities through hybrid human–robot data scaling.} With HumanWAM-Goal, we show that scaling joint human–robot pretraining improves zero-shot task execution in simulation and on physical robots. The model also exhibits embodied in-context learning, adapting its behavior to visual goals without parameter updates. After robot post-training, HumanWAM-Goal achieves a state-of-the-art average success rate of 88.65\% on RoboTwin 2.0, with 92.70\% on Clean2Clean and 84.60\% on Clean2Random.

  \item \textbf{A full-stack release for reproducible scaling research.} We provide the complete resources for reproducing our scaling studies and HumanWAM models. The release includes Open-HumanWAM-50K, a curated and aligned 50k-hour egocentric corpus with hand-motion and language annotations, together with training code, model checkpoints and evaluation protocols.
\end{enumerate}

\section{Building an Open Egocentric Pretraining Corpus}
\label{sec:data}
\begin{figure}[tb]
  \centering
  \includegraphics[width=\linewidth]{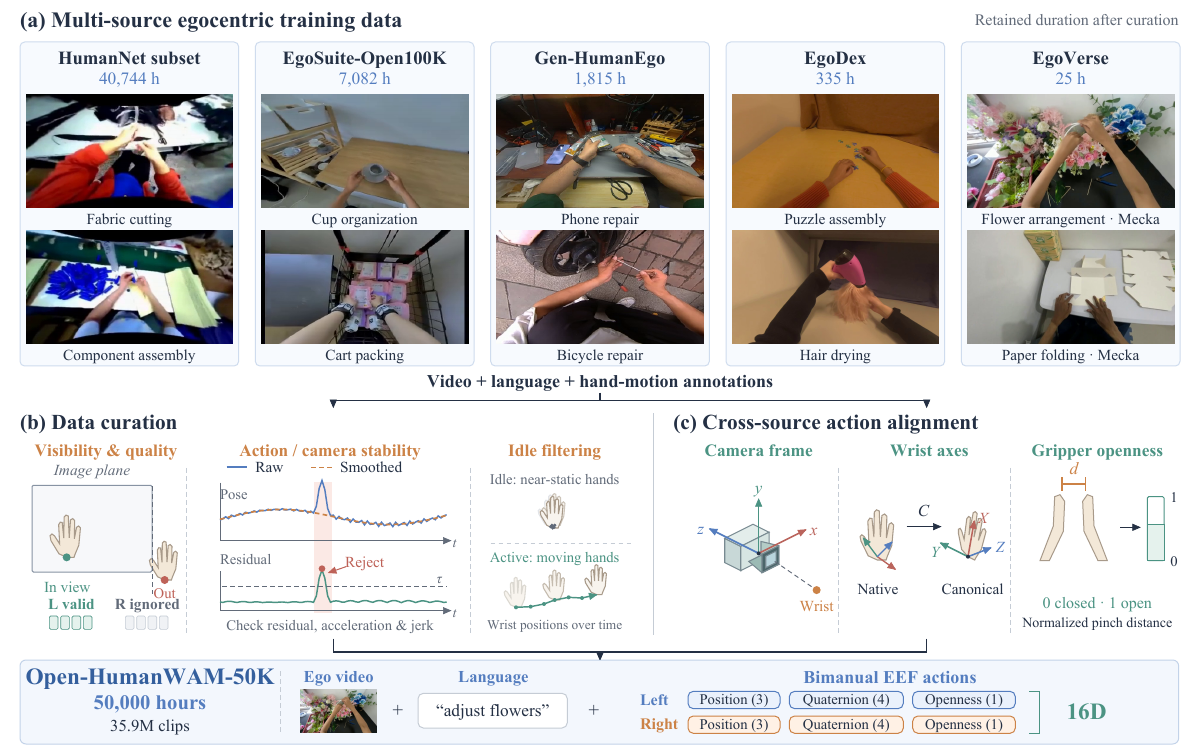}
  \caption{\textbf{Construction of Open-HumanWAM-50K.} (a) Five data sources and their retained durations. (b) Data curation based on hand visibility, motion stability, and manipulation activity. (c) Cross-source alignment of camera frames, wrist axes, and gripper openness.}
  \label{fig:data_overview}
\end{figure}

To support systematic scaling studies and large-scale egocentric pretraining, we build \textbf{Open-HumanWAM-50K}, an open corpus from multiple publicly available egocentric datasets (\cref{fig:data_overview}). HumanNet~\cite{deng2026humannet} is the primary source, contributing egocentric videos collected from the internet and paired with hand-motion and language annotations. We curate manipulation sequences based on hand visibility, motion stability, and manipulation activity. To prepare these heterogeneous sources for joint pretraining, we map action annotations into a unified bimanual end-effector representation, aligning camera coordinates, wrist axes, and gripper openness across sources to provide a common action interface for subsequent robot adaptation.

\subsection{Pretraining Data Sources}
\label{sec:sources}
To support systematic studies of egocentric scaling and large-scale model pretraining, we assemble a corpus of 50{,}000 hours of human manipulation video paired with language descriptions and hand-motion annotations. The corpus combines HumanNet, EgoSuite-Open100K, EgoDex, Gen-HumanEgo, and EgoVerse, which provide egocentric videos with accompanying hand-tracking annotations. \Cref{tab:sources} summarizes the data sources and their scales before and after curation. The corpus contains 31,396,799 physical episodes, which yield 35,934,235 continuous clips after curation, as individual episodes may be split into multiple retained segments.
The resulting 35.9 million clips cover a broad range of manipulation settings, including household, commercial service, and industrial environments. 

\begin{table}[t]
\caption{\textbf{Source datasets and curation statistics for Open-HumanWAM-50K}}
\label{tab:sources}
\centering
\small
\setlength{\tabcolsep}{3pt}
\renewcommand{\arraystretch}{1.16}
\begin{tabular*}{\linewidth}{@{\extracolsep{\fill}}lrrrrr@{}}
\toprule
\multirow{2}{*}{Dataset} & \multicolumn{2}{c}{Before curation} & \multicolumn{3}{c}{After curation} \\
\cmidrule(lr){2-3}\cmidrule(l){4-6}
 & Hours & Frames (M) & Hours & Frames (M) & Episodes \\
\midrule
HumanNet subset~\cite{deng2026humannet} & 90{,}275.25 & 9{,}749.73 & 40{,}743.58 & 4{,}400.31 & 30{,}833{,}247 \\
EgoSuite-Open100K~\cite{egosuite} & 11{,}193.99 & 1{,}208.95 & 7{,}081.65 & 764.82 & 348{,}724 \\
EgoDex~\cite{egodex} & 829.45 & 89.58 & 335.43 & 36.23 & 163{,}537 \\
Gen-HumanEgo~\cite{genhumanego} & 1{,}847.39 & 199.52 & 1{,}814.53 & 195.97 & 44{,}632 \\
EgoVerse~\cite{egoverse} & 143.33 & 15.48 & 24.81 & 2.68 & 6{,}659 \\
\midrule
Total & 104{,}289.41 & 11{,}263.26 & 50{,}000.00 & 5{,}400.00 & 31{,}396{,}799 \\
\bottomrule
\end{tabular*}
\end{table}

\paragraph{HumanNet.}
HumanNet~\cite{deng2026humannet} is a large-scale human-centric dataset combining one million hours of first- and third-person videos with language and motion annotations. We use an egocentric subset of internet videos covering manipulation, tool use, and multi-step activities across diverse objects and environments. These clips are accompanied by estimated 3D wrist poses, camera trajectories, and LLM-generated captions describing the observed activities and motions. After curation and corpus selection, we retain approximately 40{,}744 hours for pretraining.

\paragraph{EgoSuite-Open100K.}
EgoSuite-Open100K~\cite{egosuite} is a large-scale egocentric human activity dataset. It provides head-mounted video synchronized with head-camera poses and tracked hand poses. Each sample includes 21 joint transforms per hand, episode-level task and scene labels, and time-stamped subtask descriptions. The hand annotations support wrist-pose extraction and the estimation of hand openness from fingertip distances, while the task and scene labels enable explicit coverage measurements. We retain 6{,}358 hours from EgoStandard and 724 hours from EgoPro.

\paragraph{EgoVerse.}
EgoVerse~\cite{egoverse} is a continuously expanding collection of egocentric human demonstrations contributed by academic and industrial partners. Its initial release comprises 1{,}362 hours spanning 1{,}965 tasks, 240 scenes, and 2{,}087 demonstrators, combining standardized tasks collected across laboratories with open-ended activities recorded in diverse environments. The data include 3D hand and head poses, camera poses, and language annotations, with fine-grained subtask descriptions for industry-contributed recordings. We use only a small subset of the Mecka source, retaining 24.81 hours after curation for Open-HumanWAM-50K.

\paragraph{Gen-HumanEgo.}
Gen-HumanEgo~\cite{genhumanego} contains over 1{,}800 hours of human demonstrations across more than 10{,}000 tasks in domestic, commercial, industrial, and agricultural environments. The dataset provides 21 reconstructed 3D keypoints per hand, MANO~\cite{mano} parameters, hand meshes, and depth maps. Hierarchical video-, task-, and subtask-level language annotations describe activities at different granularities, with temporal boundaries for tasks and subtasks.

\paragraph{EgoDex.}
EgoDex~\cite{egodex} contains 829 hours of egocentric human demonstrations covering 194 tabletop manipulation tasks. Collected with Apple Vision Pro, it focuses on active manipulation of everyday objects, including dexterous activities such as tying shoelaces and folding laundry. The dataset pairs 30\,Hz RGB video with 3D head, upper-body, wrist, and finger poses, camera calibration, and natural-language task descriptions.

\subsection{Data Curation}
\label{sec:curation}
Effective large-scale pretraining depends on both corpus scale and supervision quality~\cite{yuan2026qwen,internw0delta}. We curate egocentric recordings to retain active manipulation with clear visual context and temporally consistent motion annotations. These checks remove unreliable annotations and prolonged inactivity while retaining usable manipulation segments for pretraining.

\paragraph{Hand visibility.}
We assess hand visibility from action annotations to retain manipulation sequences with effective visual context. We project annotated 3D EEF positions into the image using camera calibration to determine whether each end effector lies within the field of view. When only one end effector is in view, we retain the sequence and restrict action supervision to the corresponding hand, excluding the other hand from both the action loss and normalization statistics. Internet videos may also contain incomplete hand views, with only fingers or a small portion of a hand visible near image boundaries, or excessively close views that limit the visible interaction context. We identify and filter these cases using EEF depth, projected scale, and the location and size of detected hand regions.

\paragraph{Action and camera stability.}
Errors in estimated hand and camera poses can introduce abrupt jumps into manipulation trajectories. For EEF trajectories, we measure differences between original and smoothed positions, together with acceleration, jerk, and changes in position and orientation between adjacent frames. Camera trajectories are checked separately for abrupt pose changes and for excessive displacement or rotation relative to the pose at the start of each chunk. We remove frames that exceed any stability threshold and split sequences at those frames, retaining continuous segments for pretraining.

\paragraph{Idle filtering.}
Long periods of inactivity introduce redundant static samples to the pretraining corpus. We measure EEF positional variation within 32-frame sliding windows and classify a window as inactive when this variation is at or below the motion threshold. We trim inactive portions from both ends of each chunk and discard chunks with fewer than 70\% active windows.

\subsection{Unified Action Representation}
\label{sec:spec}
Joint pretraining on heterogeneous egocentric datasets requires consistent action supervision across sources. Although these datasets broaden the range of manipulation experience, their raw action data can encode similar physical motions differently, complicating the learning of shared visual--action relationships. We therefore align human trajectories and robot actions within a unified bimanual end-effector space. This alignment gives actions consistent geometric meaning across sources, allowing diverse demonstrations to contribute to a common action prediction objective.

\paragraph{Unified bimanual action space.}
To establish a common action interface between egocentric pretraining and robot adaptation, we map human and robot trajectories into a unified bimanual end-effector (EEF) space. Human wrist poses provide EEF positions and orientations, while gripper openness is derived from the available hand annotations. The resulting action vector is
\begin{equation}
  \mathbf{a}_t =
  [\mathbf{p}_t^L,\mathbf{q}_t^L,g_t^L,
   \mathbf{p}_t^R,\mathbf{q}_t^R,g_t^R]
  \in \mathbb{R}^{16},
\end{equation}
where $\mathbf{p}_t^h\in\mathbb{R}^3$ is the position in meters, $\mathbf{q}_t^h\in\mathbb{R}^4$ is a unit quaternion in $wxyz$ order, and $g_t^h\in[0,1]$ denotes gripper openness for hand $h\in\{L,R\}$. This representation captures bimanual motion and hand opening without requiring a shared joint configuration across human and robot embodiments. We restrict the action space to bimanual manipulation without introducing additional dimensions for locomotion, which is motivated by the difficulty of deriving meaningful supervision from largely task-irrelevant, irregular human body or head motion in egocentric data, compounded by incomplete annotations across sources.

\paragraph{Camera-frame alignment.}
A shared end-effector representation does not by itself resolve geometric differences between source-specific reference frames. To align action supervision with visual observations, we transform wrist trajectories into the camera frame associated with each video frame. For sources with world-frame wrist trajectories and camera extrinsics, let $T_{AB}$ map coordinates from frame $B$ to frame $A$. The wrist pose in the camera frame is
\begin{equation}
  T_{C_{\mathrm{cv}}E}^{h}(t)
  = T_{WC_{\mathrm{cv}}}(t)^{-1}T_{WE}^{h}(t).
\end{equation}
Wrist and camera poses are synchronized at each timestamp, so each action is expressed relative to the camera at the corresponding target time. We further standardize the camera axes to $+x$ right, $+y$ up, and $+z$ backward. For example, EgoVerse camera coordinates use $+y$ down and $+z$ forward. We convert these coordinates by left-multiplying both position vectors and rotation matrices by $S=\operatorname{diag}(1,-1,-1)$,
\begin{equation}
  \mathbf{p}_{CE}^{h}(t)
  = S\mathbf{p}_{C_{\mathrm{cv}}E}^{h}(t),
  \qquad
  R_{CE}^{h}(t)
  = SR_{C_{\mathrm{cv}}E}^{h}(t).
\end{equation}

\paragraph{Wrist-axis alignment.}
A common camera frame does not resolve differences in the local coordinate axes assigned to each wrist. We define a canonical wrist frame with $+X$ pointing from the wrist toward the middle-finger knuckle, $+Y$ along the thumb direction projected onto the plane orthogonal to $+X$, and $+Z=X\times Y$. For each source $s$ and hand $h$, a constant rotation $C_{s,h}\in\mathrm{SO}(3)$ aligns the native wrist axes with this convention,
\begin{equation}
  R_{\mathrm{aligned}}^{h}(t)
  = R_{\mathrm{camera}}^{h}(t)C_{s,h}.
\end{equation}
For sources with 3D hand keypoints and native wrist orientations, we estimate $C_{s,h}$ through orthogonal Procrustes alignment~\cite{procrustes} with anatomical frames derived from the keypoints. Gen-HumanEgo wrist orientations are constructed directly from these anatomical frames. For HumanNet, we determine the axis permutations and sign changes using projected wrist axes and monocular hand landmarks.

\paragraph{\textbf{Gripper alignment.}}
For sources with hand keypoints, we map the thumb--index fingertip distance \(d_t^h\) to normalized gripper openness,
\[
g=\operatorname{clip}\!\left(
\frac{d-d_{\mathrm{closed}}}
{d_{\mathrm{open}}-d_{\mathrm{closed}}},
0,1
\right).
\]
Here, \(d_{\mathrm{closed}}\) and \(d_{\mathrm{open}}\) denote reference distances for closed and open hand configurations, respectively. These distances are specified for each data source, with separate values for the two hands where applicable. Across sources, \(0\) denotes fully closed and \(1\) denotes fully open.

\section{Exploration of Pure Egocentric Data Scaling}
\label{sec:study}
Egocentric data offer a promising route to scaling robot foundation models through abundant and diverse human manipulation experience. Prior work has investigated how pretraining exclusively on human egocentric data affects performance on downstream real-robot tasks~\cite{egoscale,humanscale}. However, scaling outcomes vary across pretraining recipes, with limited or inconsistent downstream generalization gains reported in other settings~\cite{openwam,ego4wam}. We first establish empirical scaling laws for egocentric pretraining with our proposed recipe, then examine data diversity and the sample efficiency of robot adaptation, and finally conduct controlled analyses of the pretraining and adaptation recipe.
\begin{figure}[tb]
  \centering
  \includegraphics[width=\linewidth]{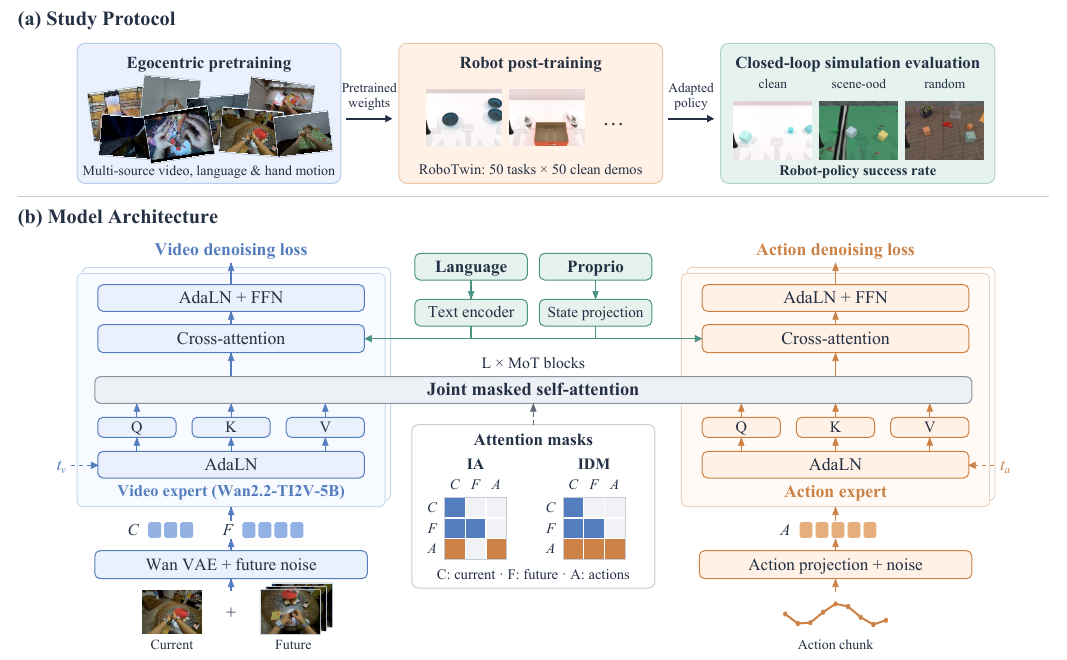}
  \caption{\textbf{Model architecture of HumanWAM-Vanilla and experimental protocol.} (a) Egocentric pretraining, robot post-training on clean demonstrations, and closed-loop evaluation under clean, scene-ood, and random conditions. (b) Conditioned on language and proprioception, the video and action experts interact through joint masked self-attention. The masks specify visibility among current-frame (C), future-video (F), and action (A) tokens under isolated attention (IA) and inverse-dynamics (IDM) attention.}
  \label{fig:study_model_architecture}
\end{figure}

We organize the scaling analysis around the following questions:
\begin{itemize}[leftmargin=1.2em, itemsep=1pt]
  \item \textbf{Q1.} What improves in downstream robot policies as egocentric pretraining scales from 100 to 50k hours? 
  \item \textbf{Q2.} How important is data diversity for effective egocentric scaling?
  \item \textbf{Q3.} How does scaling egocentric pretraining affect the sample efficiency of robot adaptation? 
  \item \textbf{Q4.} Does cross-source action alignment matter for egocentric pretraining? 
 \item \textbf{Q5.} Does predicting human motion during egocentric pretraining improve robot action learning?

\item \textbf{Q6.} Can egocentric pretraining surpass robot pretraining at matched data scales?

\item \textbf{Q7.} How do we extend single-view egocentric pretraining to multi-view robot policies?
\end{itemize}

\subsection{Pretraining and Evaluation Protocol}
\label{sec:backbone}
\label{sec:protocol}
To study scaling in pure egocentric pretraining, we adopt a unified world--action model backbone and a standardized training and evaluation pipeline (\cref{fig:study_model_architecture}). After pretraining on human egocentric data, we adapt each model to RoboTwin~2.0 using clean demonstrations and measure its task success rate in closed-loop evaluation. Standardized tasks and success criteria, controlled environmental variation, and shared evaluation seeds support reproducible comparisons across pretraining scales and training settings. Evaluations within the demonstration distribution and under environmental randomization measure in-distribution performance and out-of-distribution generalization, respectively. Below, we describe the model architecture and the pretraining, robot post-training, and evaluation procedures.

\subsubsection{Model.}
\label{sec:study_model}
Our experiments build on HumanWAM-Vanilla, a world--action model that jointly generates future video and actions from a single observation and a language instruction. The model extends Wan2.2-TI2V-5B~\cite{wan2025} with an action expert that has the same depth as the video backbone, forming a Mixture-of-Transformers architecture. Following prior work~\cite{yuan2026fastwamworldactionmodels,motubrain,imagewam}, we apply a widely used isolated attention mask in most experiments, allowing action tokens to attend to the first video frame but not to future video frames. During inference, the model can therefore omit future video generation and denoise only the action chunk, enabling fast policy execution. We use the unified 16-D bimanual end-effector action space defined in \cref{sec:spec} and compute delta poses in the camera coordinate frame relative to the first-frame observation.

\par\penalty0\noindent
\begin{minipage}[b]{\linewidth}
\subsubsection{Pretraining.}
\noindent\begin{minipage}[t]{0.48\linewidth}
\vspace{0pt}
We pretrain on different egocentric corpora with the same training recipe. We use the AdamW optimizer~\cite{adamw} with a peak learning rate of $10^{-4}$ and a weight decay of $0.01$. The learning rate follows a cosine schedule over 100k steps, with a linear warm-up during the first 5\% of the steps.
The training budget is adapted to the corpus size. In preliminary runs we observed that smaller corpora saturate quickly and that their generalization can even degrade with further training, so a fixed budget would either under-train the large corpora or over-train the small ones. We therefore adjust the global batch size according to the corpus size, as listed in \cref{tab:lineages}, and we post-train and evaluate checkpoints saved every 10k pretraining steps so that each corpus is compared at its own best.
\end{minipage}\hfill
\begin{minipage}[t]{0.48\linewidth}
\vspace{0pt}
\begingroup
\centering
\captionsetup{font=small,aboveskip=0pt,belowskip=5pt}
\captionof{table}{\textbf{Pretraining settings across data scales.}}
\label{tab:lineages}
\small
\setlength{\tabcolsep}{3pt}
\renewcommand{\arraystretch}{1.08}
\begin{tabular*}{\linewidth}{@{\extracolsep{\fill}}rrrr@{}}
\toprule
Hours & \shortstack[r]{Frames\\(M)} & \shortstack[r]{Global\\batch size} & \shortstack[r]{Epochs /\\10k steps} \\
\midrule
100 & 10.8 & 512 & 0.58 \\
500 & 54.0 & 1{,}024 & 0.23 \\
1{,}000 & 108.0 & 2{,}048 & 0.23 \\
5{,}000 & 540.0 & 2{,}048 & 0.05 \\
10{,}000 & 1{,}080 & 4{,}096 & 0.04 \\
20{,}500 & 2{,}214 & 4{,}096 & 0.02 \\
50{,}000 & 5{,}400 & 8{,}192 & 0.02 \\
\bottomrule
\end{tabular*}
\par\smallskip
\endgroup

\end{minipage}
\end{minipage}\par

\subsubsection{Post-training.}
We post-train the pretrained model on RoboTwin~2.0~\cite{robotwin2}, a bimanual tabletop benchmark with 50 tasks on the Aloha-AgileX embodiment. The model is post-trained on 50 clean demonstrations per task without data augmentation for 12k optimizer steps with a global batch size of 512, corresponding to approximately 10 epochs. For each pretraining run, we post-train checkpoints saved every 10k pretraining steps and evaluate the resulting policies in both clean and out-of-distribution settings, as described below, to track downstream performance and identify the best checkpoint for each corpus. We retain the single head-camera view from egocentric pretraining because introducing three camera views during post-training impairs the transfer of egocentric pretraining priors and head-camera images alone provide sufficient visual information to perform all RoboTwin tasks (see \cref{sec:camera_view_ablation}).
\begin{figure}[th]
  \centering
  \includegraphics[width=\linewidth]{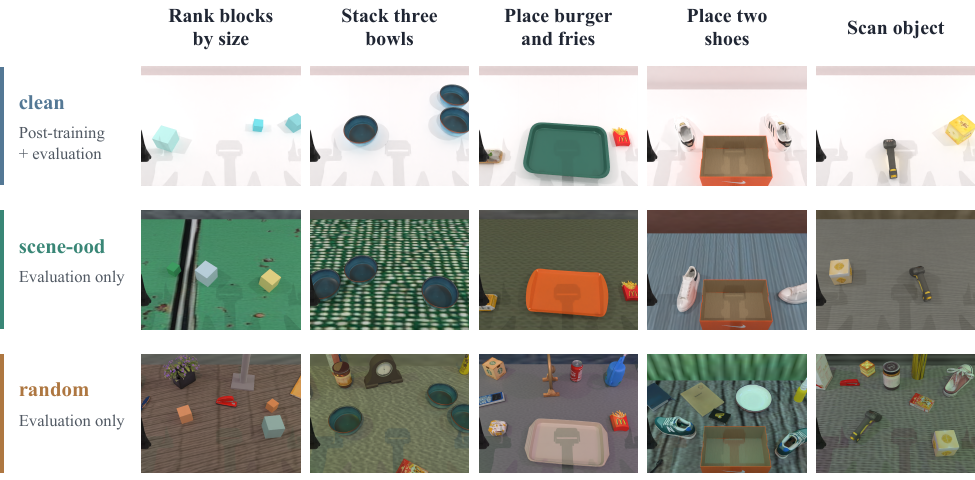}
  \caption{\textbf{RoboTwin evaluation setup for our egocentric pretraining exploration.} We post-train policies exclusively on clean demonstrations and evaluate them under clean, scene-ood, and random conditions.}
  \label{fig:robotwin_benchmark}
\end{figure}
\subsubsection{Evaluation.}
\label{sec:ckpt_rule}
To assess how egocentric pretraining affects in-distribution performance and out-of-distribution generalization, we evaluate post-trained checkpoints under the clean, scene-ood, and random settings. Each checkpoint is rolled out on all 50 tasks with 50 episodes per task in each setting. Representative tasks and evaluation conditions are illustrated in \cref{fig:robotwin_benchmark}.
\begin{itemize}[leftmargin=1.2em, itemsep=1pt]
  \item \textbf{clean}: the training distribution, with the plain background, fixed lighting, no clutter, and fixed table height used to collect the demonstrations.
  \item \textbf{scene-ood}: randomized backgrounds and tabletop textures, with all other conditions identical to clean.
  \item \textbf{random}: the full RoboTwin randomization, with randomized backgrounds, table clutter, randomized lighting, and table-height jitter of up to 3\,cm.
\end{itemize}
Clean measures how well the policy fits the demonstration distribution, whereas scene-ood and random assess gains in downstream generalization from egocentric pretraining. Specifically, scene-ood focuses on robustness to visual scene variation. We find that the post-trained checkpoint with the best clean performance does not necessarily generalize best to randomized scenes. To examine pretraining gains in in-distribution performance and out-of-distribution generalization separately, we therefore report the success rate of the best checkpoint independently for each evaluation setting. To support fair comparisons and accelerate evaluation, we save expert-validated evaluation seeds and reuse them across evaluation rounds.

\subsection{Scaling Egocentric Data from 100 to 50k Hours: What Improves?}
\label{sec:study1}
Whether and how scaling human data improves robot policy performance is a central question for robot foundation models. We investigate this question through a systematic study of egocentric pretraining from 100 to 50,000 hours. Under the same pretraining recipe, we find that scaling egocentric pretraining consistently improves downstream generalization across all three settings. We validate this trend under both isolated attention and inverse-dynamics (IDM) attention, which additionally allows action tokens to attend to the future video.

\begin{figure}[th]
  \centering
  \includegraphics[width=\linewidth]{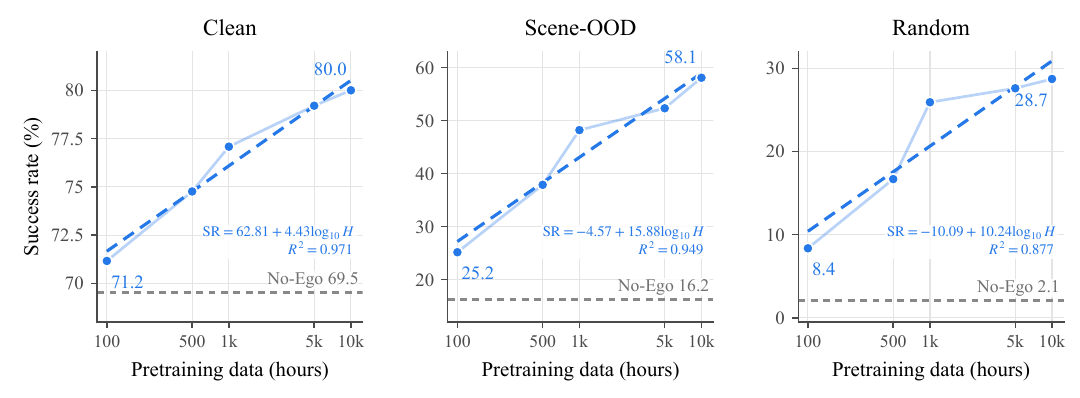}
\caption{\textbf{Scaling egocentric pretraining with isolated attention improves in-distribution performance and generalization.} Blue dashed lines show least-squares fits of $\mathrm{SR}=a+b\log_{10}H$, where $H$ denotes pretraining hours, with $R^2$ reported in each panel. Gray dashed lines denote the baseline without egocentric pretraining. Success rates increase across all three settings, with larger gains on scene-ood and random, indicating that scaling primarily strengthens generalization beyond the demonstration scenes.}
  \label{fig:scaling_hours_default}
\end{figure}

We first examine scaling with isolated attention. Models are pretrained on 100, 500, 1{,}000, 5{,}000, and 10{,}000 hours of egocentric data, then post-trained and evaluated under a common downstream protocol. \Cref{fig:scaling_hours_default} shows that scaling egocentric pretraining with isolated attention consistently improves downstream policy performance on RoboTwin. On clean, success increases from 71.2\% at 100 hours to 80.0\% at 10{,}000 hours, compared with 69.5\% without egocentric pretraining. These gains indicate improved in-distribution performance under the same post-training protocol. The benefits are larger under distribution shift. From 100 to 10{,}000 hours, scene-ood success rises from 25.2\% to 58.1\%, while random success increases from 8.4\% to 28.7\%. Without egocentric pretraining, the corresponding success rates are only 16.2\% and 2.1\%.

Downstream success follows an empirical log-linear scaling law, $\mathrm{SR}=a+b\log_{10}H$, where $H$ denotes egocentric pretraining hours. The high goodness of fit ($R^2=0.971$, $0.949$, and $0.877$ for clean, scene-ood, and random, respectively) indicates a consistent relationship between pretraining scale and downstream performance. Each tenfold increase in pretraining hours corresponds to fitted gains of 4.43\,pp on clean, 15.88\,pp on scene-ood, and 10.24\,pp on random. These trends suggest that further scaling egocentric pretraining will continue to improve downstream policy performance, with particularly pronounced gains in out-of-distribution generalization.

\textbf{Recent work motivates us to examine whether this scaling trend extends to architectures that condition actions on future visual information}. We therefore validate the scaling experiments with inverse-dynamics attention and absolute camera-frame poses, allowing action tokens to attend to both current and future video frames. During training, action denoising is conditioned on the ground truth future video with noise augmentation. During inference, the model first generates future video latents and then predicts actions. This configuration substantially improves downstream performance, with detailed ablations examining the contributions of these design choices in \cref{sec:model_ablation}.

\begin{figure}[th]
  \centering
  \includegraphics[width=\linewidth]{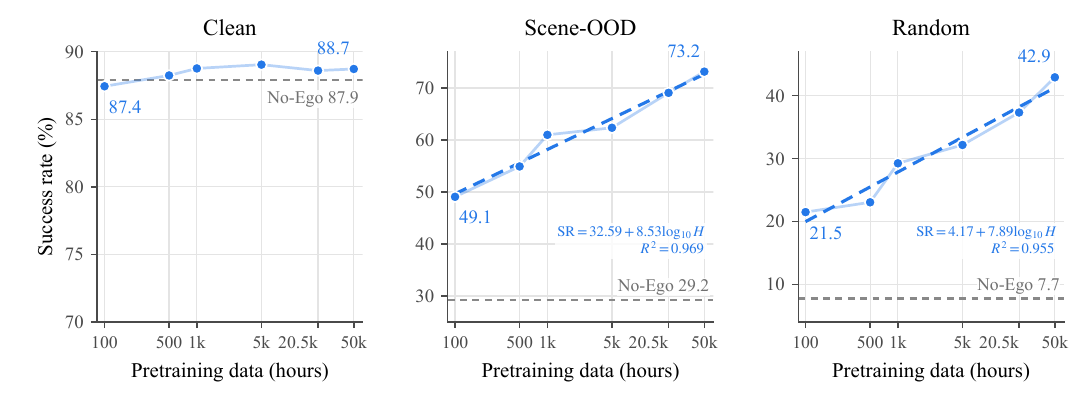}
  \caption{\textbf{Scaling egocentric pretraining with IDM attention strengthens downstream generalization.} Clean success remains high across scales, while scene-ood and random success improve consistently. Blue dashed lines show log-linear scaling fits for the two out-of-distribution settings, with $R^2$ reported in the corresponding panels. Gray dashed lines denote the baseline without egocentric pretraining.}
  \label{fig:scaling_hours_idm}
\end{figure}
As shown in \cref{fig:scaling_hours_idm}, clean success remains between 87.4\% and 89.0\% as egocentric pretraining scales from 100 to 50{,}000 hours. Even without egocentric pretraining, the model achieves 87.9\% on clean, indicating strong in-distribution performance under IDM attention with absolute camera-frame poses. Scaling therefore yields only limited additional gains on the training distribution. The generalization gains are substantially larger. From 100 to 50{,}000 hours, scene-ood success increases from 49.1\% to 73.2\%, and random success rises from 21.5\% to 42.9\%, compared with 29.2\% and 7.7\% without egocentric pretraining. Improvements persist beyond 5{,}000 hours, where the corresponding success rates are 62.4\% and 32.2\%. Both out-of-distribution settings closely follow the empirical log-linear scaling law observed with isolated attention, with higher $R^2$ values of 0.969 for scene-ood and 0.955 for random. This consistency across the two model configurations suggests that the generalization benefits of scaling egocentric pretraining extend beyond a specific attention design and action representation, even when in-distribution performance is already strong.

\subsection{Scaling Diversity, Not Just Hours}
\label{sec:study2}
Recent studies~\cite{egoscale,humanscale,dyna2} have explored the scaling behavior of egocentric pretraining, with data duration as the primary scaling variable. Data diversity, including variation in tasks and scenes, is another important dimension of scaling embodied pretraining~\cite{robotdatascaling,robotdiversity}. We investigate this dimension through controlled experiments on the EgoSuite-Open100K dataset with matched durations but varying task and scene coverage. Our results show that jointly scaling diversity and duration yields greater downstream gains than increasing duration alone.

\paragraph{Setup.}
We construct two series of pretraining corpora at 100, 500, 1{,}000, and 5{,}000 hours from the EgoSuite-Open100K dataset. In the \emph{scaled-diversity} series, task and scene coverage expands with data duration. In the \emph{matched-diversity} series, the 100-, 500-, and 1{,}000-hour subsets cover the same set of tasks and scenes as the 5{,}000-hour parent corpus, so increasing duration adds trajectories within fixed task and scene coverage. At 100 hours, the matched-diversity corpus spans 2{,}288 tasks and 2{,}953 scenes, compared with 62 tasks and 222 scenes in the scaled-diversity corpus. For each corpus, we pretrain the model, apply the same post-training protocol, and evaluate success rates under all three settings.

\begin{table}[htbp]
\caption{Task and scene coverage of the two EgoSuite-Open100K corpus families and their shared 5{,}000-hour parent corpus. Task and scene counts are based on the dataset captions.}
\label{tab:corpus_families}
\centering
\small
\setlength{\tabcolsep}{8pt}
\renewcommand{\arraystretch}{1.15}
\begin{tabular*}{0.8\linewidth}{@{\extracolsep{\fill}}lrrrr@{}}
\toprule
Family & Hours & Clips & Tasks & Scenes \\
\midrule
\multirow{3}{*}{Scaled-diversity} & 100 & 62{,}206 & 62 & 222 \\
 & 500 & 305{,}889 & 309 & 961 \\
 & 1{,}000 & 589{,}216 & 547 & 1{,}442 \\
\midrule
\multirow{3}{*}{Matched-diversity} & 100 & 81{,}076 & 2{,}288 & 2{,}953 \\
 & 500 & 345{,}053 & 2{,}288 & 2{,}953 \\
 & 1{,}000 & 653{,}567 & 2{,}288 & 2{,}953 \\
\midrule
Shared parent & 5{,}000 & 2{,}813{,}547 & 2{,}288 & 2{,}953 \\
\bottomrule
\end{tabular*}
\end{table}

\begin{figure}[th]
  \centering
  \includegraphics[width=\linewidth]{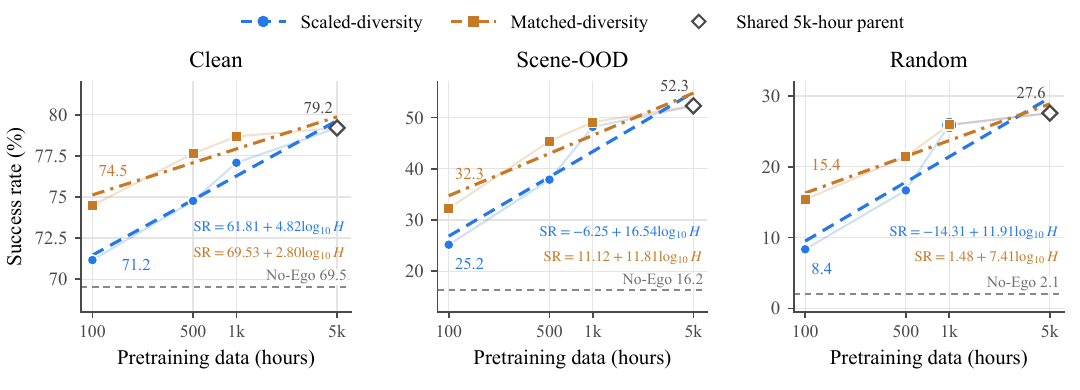}
  \caption{\textbf{Scaling egocentric pretraining with growing or fixed task and scene diversity.} Blue circles denote scaled-diversity corpora, and orange squares denote matched-diversity corpora. Dashed and dash-dotted lines show the respective least-squares fits of $\mathrm{SR}=a+b\log_{10}H$, with both equations shown in each panel. Scaled-diversity yields a steeper fitted slope in all three settings.}
  \label{fig:scale_coverage}
\end{figure}

\paragraph{Results.}
\Cref{fig:scale_coverage} shows that expanding diversity alongside duration produces steeper scaling slopes than increasing duration at fixed coverage. Relative to matched-diversity, scaled-diversity increases the fitted slopes by 2.02 on clean, 4.73 on scene-ood, and 4.50 on random, indicating greater performance gains as pretraining data scale. The larger slope increases on scene-ood and random suggest that diversity expansion particularly benefits downstream generalization. \textbf{These results suggest a practical scaling rule for egocentric pretraining: jointly scaling diversity and duration yields greater performance gains than scaling duration alone.}

\subsection{Scaling Egocentric Pretraining for Data-Efficient Robot Adaptation}
\label{sec:study5}
Generalization to new tasks and in-context learning have attracted growing interest in embodied pretraining~\cite{intelligence2025pi_,icrt,liu2026roboicl}. In this section, we seek to explore how egocentric pretraining scale affects the sample efficiency of downstream robot policy adaptation.
\paragraph{Setup.}
To assess how egocentric pretraining scale affects the sample efficiency of robot adaptation, we conduct post-training experiments with different initializations and demonstration budgets. We select the best-performing checkpoints from pretraining on 500 and 5{,}000 hours of egocentric data and compare them with a baseline without egocentric pretraining (No-Ego). For each initialization, separate post-training runs use 5, 10, 25, or 50 clean demonstrations per task.

\paragraph{Results.}
\begin{figure}[thb]
  \centering
  \includegraphics[width=\linewidth]{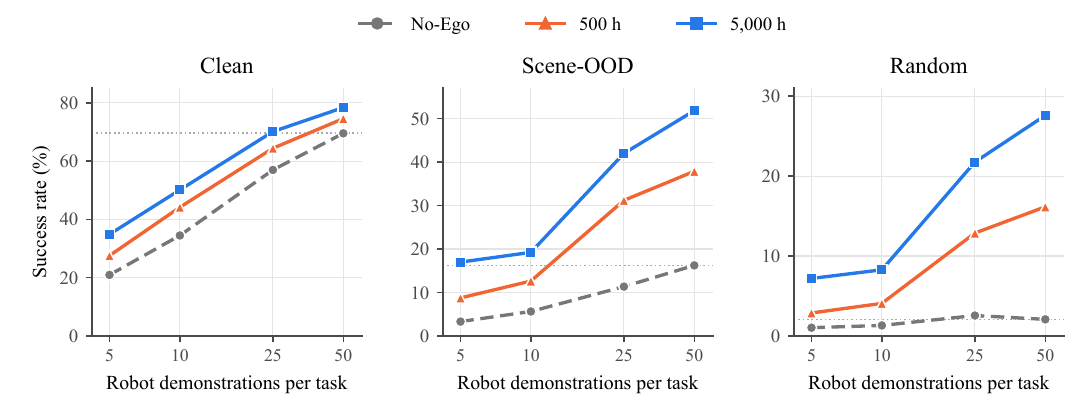}
  \caption{\textbf{Scaling egocentric pretraining improves the sample efficiency of robot adaptation.} Success rates as a function of the number of robot demonstrations per task for different egocentric pretraining scales. }
  \label{fig:demo_ratio}
\end{figure}

\Cref{fig:demo_ratio} shows a consistent scaling trend in post-training sample efficiency. Across all three settings, the 500-hour model outperforms No-Ego at every demonstration budget, with further gains from scaling to 5{,}000 hours. With 10 demonstrations per task, clean success increases from 34.5\% for No-Ego to 44.2\% after 500 hours of pretraining and 50.1\% after 5{,}000 hours, indicating progressively more effective adaptation from limited robot demonstrations. At the same budget, scaling from 500 to 5{,}000 hours also approximately doubles random success, from 4.1\% to 8.3\%. The additional pretraining therefore improves both in-distribution performance and generalization without requiring more robot demonstrations.

These gains also lower demonstration requirements for comparable performance. On scene-ood, the No-Ego success rate with 50 demonstrations per task is exceeded by the 500-hour model with 25 demonstrations and by the 5{,}000-hour model with only 5. Further scaling from 500 to 5{,}000 hours continues to reduce dependence on robot demonstrations. With 25 demonstrations, the 5{,}000-hour model outperforms the 500-hour model trained on 50 demonstrations in both scene-ood (42.0\% vs.\ 37.9\%) and random (21.8\% vs.\ 16.2\%). This improvement establishes a clear scaling trend in post-training sample efficiency, with stronger generalization achieved using half the robot demonstrations.

\subsection{Does Cross-Source Action Alignment Matter?}
\label{sec:study3}
Egocentric datasets obtain human action trajectories through different data collection and motion estimation pipelines, producing raw action data that lack a consistent physical interpretation across sources. Prior work incorporates geometric alignment to support learning from heterogeneous human and robot data~\cite{internw0delta,aceego0,yuan2026qwen}. We quantify the effect of camera-frame and wrist-axis alignment on pretraining loss and downstream policy performance under a fixed data mixture and training configuration.
\paragraph{Setup.}
To evaluate the effect of cross-source action alignment, we construct a 1{,}000-hour corpus comprising EgoSuite-Open100K (593.7\,h), HumanNet (250\,h), EgoDex (114.5\,h), and EgoVerse (41.8\,h). We compare two models pretrained on identical clips and captions with the same sampling strategy, training schedule, and normalization procedure. In the \emph{Aligned} setting, actions from each source are transformed into camera coordinates, and wrist axes are standardized to obtain a consistent action representation. In the \emph{Raw} setting, actions from each source are used directly in their original representation.
\begin{figure}[th]
  \centering
  \includegraphics[width=\linewidth]{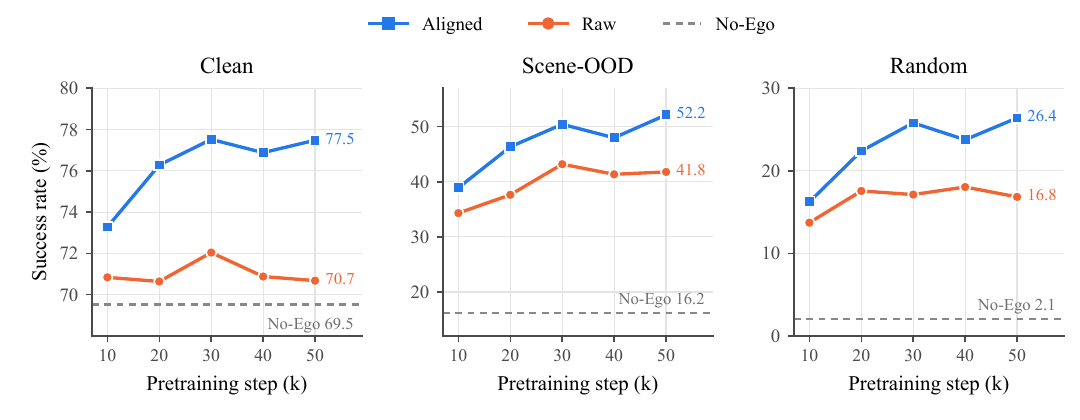}
  \caption{\textbf{Cross-source action alignment improves downstream policy performance.} Aligned and Raw models are pretrained on the same four-source 1{,}000-hour corpus using aligned and native action representations. Each point reports the best success rate across post-trained checkpoints for the corresponding pretraining run and evaluation setting. Gray dashed lines denote the baseline without egocentric pretraining.}
  \label{fig:alignment}
\end{figure}

\paragraph{Results.}
\Cref{fig:alignment} shows that Aligned outperforms Raw at every evaluated pretraining step across all three settings. Comparing the best checkpoints, clean success increases from 72.0\% to 77.5\% (+5.5\,pp). The gains are larger under distribution shift, with scene-ood improving from 43.2\% to 52.2\% and random from 18.0\% to 26.4\%. Both pretrained models outperform the No-Ego baseline, showing that multi-source pretraining remains beneficial without alignment. Under the same data mixture and training protocol, cross-source action alignment further improves in-distribution performance and, more substantially, out-of-distribution generalization.
\begin{figure}[th]
  \centering
  \includegraphics[width=\linewidth]{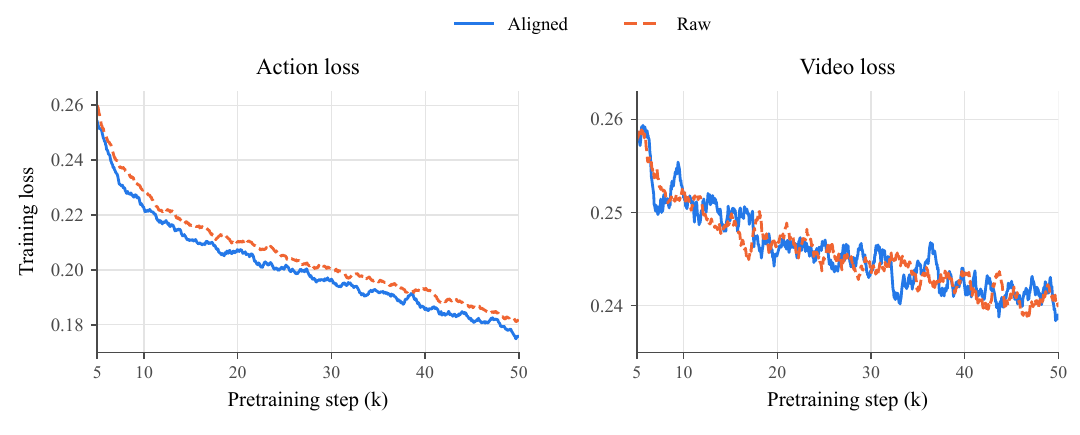}
  \caption{\textbf{Cross-source action alignment reduces pretraining action loss with little change in video loss.} Curves show training losses for Aligned and Raw models on the same four-source 1{,}000-hour corpus.}
  \label{fig:alignment_loss}
\end{figure}

The effect of action alignment is also evident during pretraining (\cref{fig:alignment_loss}). Aligned exhibits lower action loss, while video loss remains comparable between the two models. These trends suggest that a well-aligned action representation provides more consistent supervision across sources and facilitates joint learning.

\subsection{Action Supervision in Egocentric Pretraining}
\label{sec:action_supervision}
Egocentric human data capture rich physical interactions in first-person videos paired with hand trajectories~\cite{egodex,egoverse}. Whether to incorporate human motion prediction into world--action model pretraining remains an open question~\cite{ego4wam,internw0,motubrain,zimablue}. We show that jointly predicting human motion and future video substantially improves downstream policy generalization over video prediction alone on the same egocentric corpus.

\paragraph{Setup.}
We pretrain two models on the same 1{,}000-hour scaled-diversity corpus, with and without human motion prediction. For the video-only baseline, we set the action loss weight to zero and optimize video prediction alone. To assess the additional value of human motion targets, we train a second model with action loss enabled only for a fixed subset of original episodes covering approximately 50\% of the total video duration. All models receive full robot action supervision during post-training, following \cref{sec:protocol}. 

\begin{figure}[thb]
  \centering
  \includegraphics[width=\linewidth]{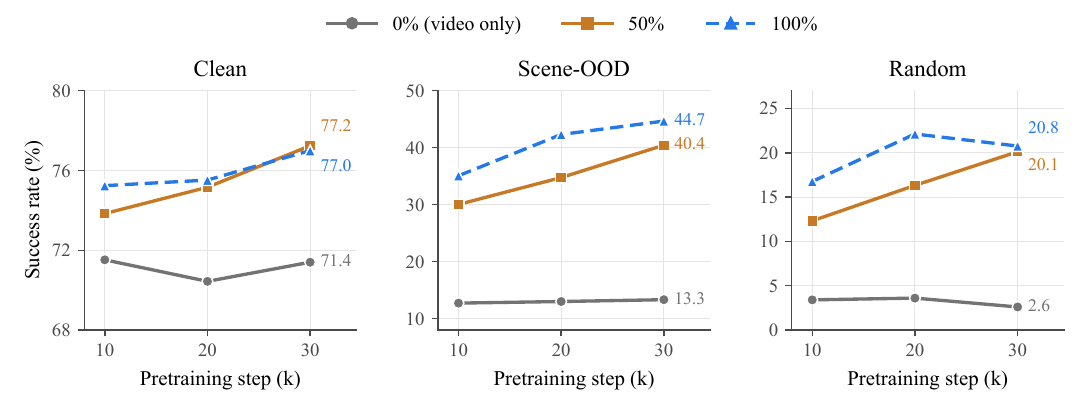}
  \caption{\textbf{Human action supervision during egocentric pretraining improves downstream generalization.} Models are pretrained on the same 1{,}000-hour video corpus with 0\%, 50\%, or 100\% human action supervision. }
  \label{fig:action_supervision}
\end{figure}

\paragraph{Results.}
\Cref{fig:action_supervision} shows that learning human motion alongside video prediction consistently improves downstream robot policy performance across all three settings. At 20k pretraining steps, using motion targets for approximately 50\% of the video duration increases clean success from 70.4\% to 75.2\% compared with video-only pretraining. The gains are larger under distribution shift, with scene-ood improving from 13.0\% to 34.8\% and random from 3.6\% to 16.3\%. The two objectives also exhibit distinct trends as pretraining progresses. From 10k to 30k steps, video-only performance remains nearly flat, with scene-ood and random success staying near 13\% and 3\%, respectively. Over the same interval, the model that jointly predicts human motion and future video improves steadily from 30.0\% to 40.4\% on scene-ood and from 12.3\% to 20.1\% on random. The widening gap indicates that human motion targets improve both performance at matched pretraining steps and the gains in downstream generalization from continued pretraining.

\subsection{Can Egocentric Pretraining Surpass Robot Pretraining?}
\label{sec:robot_vs_ego}
Egocentric human data offer broader interaction coverage, greater scene diversity, and higher information density at lower collection cost than teleoperated robot data~\cite{humanscale}. To assess their relative value for robot policy learning, HumanScale pretrained models on equal-duration egocentric or real-robot corpora, then post-trained each model on the same real-robot demonstrations. This comparison favored egocentric pretraining, which yielded lower validation action loss on held-out real-robot trajectories, particularly on tasks excluded from post-training. We extend this matched-scale comparison to closed-loop task success on RoboTwin to examine how the relative performance of the two pretraining sources varies with data scale.

\paragraph{Setup.}
We compare egocentric and real-robot pretraining at 100, 500, 1{,}000, and 5{,}000 hours. The egocentric series uses the scaled-diversity corpora in \cref{fig:scaling_hours_default}, while the real-robot series draws on AgiBot Beta~\cite{agibotworld}, ABC-130k~\cite{abc130k}, and our in-house Cobot Magic data as shown in \cref{tab:robot_pretraining_corpora}. Both series use only head-camera observations during pretraining, given the effects of additional wrist-camera views discussed in \cref{sec:camera_view_ablation}. We curate the robot corpora from language-annotated trajectories with verified head-camera extrinsics, excluding data with base or head-camera motion.

\begin{table}[t]
  \centering
  \caption{\textbf{Composition of the real-robot pretraining corpora.}}
  \label{tab:robot_pretraining_corpora}
  \small
  \setlength{\tabcolsep}{5pt}
  \begin{tabular}{rrrrr}
    \toprule
    Scale (h) & AgiBot Beta (h) & XDOF/ABC-130k (h) & Cobot (h) & Total frames \\
    \midrule
    100       & 32.19      & 67.80       & ---   & 10{,}798{,}713 \\
    500       & 176.52     & 323.46      & ---   & 53{,}997{,}982 \\
    1{,}000   & 361.70     & 638.16      & ---   & 107{,}985{,}621 \\
    5{,}000   & 1{,}944.80 & 2{,}983.46  & 71.52 & 539{,}977{,}159 \\
    \bottomrule
  \end{tabular}
\end{table}

\begin{figure}[thb]
  \centering
  \includegraphics[width=\linewidth]{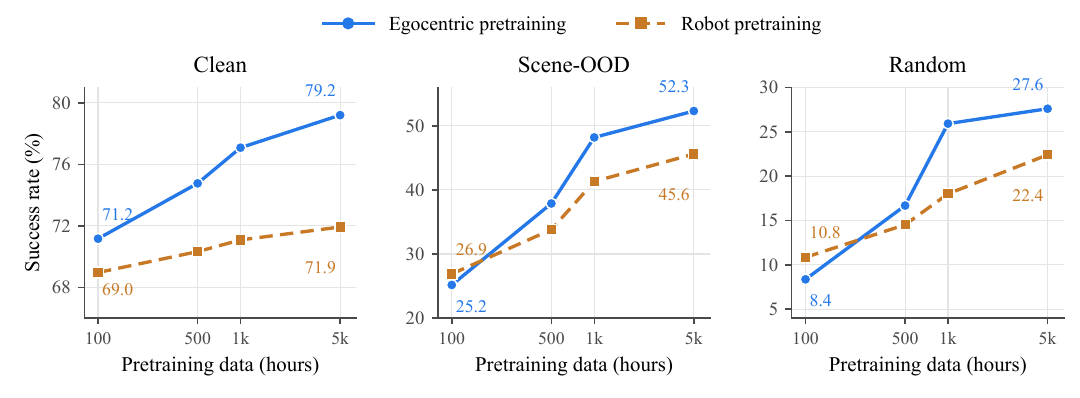}
  \caption{\textbf{Egocentric pretraining outperforms robot pretraining from 500 to 5{,}000 hours.} Downstream RoboTwin success rates after pretraining on 100, 500, 1{,}000, and 5{,}000 hours of egocentric human or robot data. The egocentric series follows the scaled-diversity configuration in \cref{fig:scaling_hours_default}. Each point reports the best post-training success rate among the evaluated pretraining checkpoints for that scale and setting.}
  \label{fig:robot_vs_ego_scaling}
\end{figure}

\paragraph{Results.}
\Cref{fig:robot_vs_ego_scaling} shows that egocentric pretraining outperforms robot pretraining in all three settings from 500 to 5{,}000 hours. At 100 hours, robot pretraining retains a small advantage on scene-ood and random, but this ordering reverses at larger scales. On clean, the egocentric advantage grows from 2.2\,pp at 100 hours to 7.3\,pp at 5{,}000 hours. At 5{,}000 hours, egocentric pretraining also achieves higher out-of-distribution success, reaching 52.3\% versus 45.6\% on scene-ood and 27.6\% versus 22.4\% on random. Its advantage therefore spans both in-distribution performance and robustness under distribution shift. The 1{,}000-hour egocentric model also outperforms the 5{,}000-hour robot model in all three settings, including scene-ood (48.2\% versus 45.6\%) and random (25.9\% versus 22.4\%). These comparisons extend the validation-loss evidence from HumanScale to closed-loop task success on RoboTwin, where egocentric pretraining yields stronger downstream policies despite the embodiment gap.

\subsection{From Egocentric Pretraining to Multi-View Robot Policies}
\label{sec:camera_view_ablation}
Most egocentric human data are collected with a single head-mounted camera~\cite{egodex,egoverse}. Although head-camera observations alone can support most robot manipulation tasks, wrist cameras provide complementary close-range views for precise gripper--object interactions and improve visibility when the head view is occluded~\cite{droid,aloha}. We find that introducing wrist-camera inputs directly during robot post-training can substantially damage the generalization benefits of egocentric pretraining. EgoPro~\cite{egosuite} provides approximately 1,000 hours of egocentric human video with synchronized head- and wrist-camera views. We use this dataset to investigate how to extend egocentric pretraining to multi-view robot policies while preserving its generalization benefits.
\begin{figure}[th]
  \centering
  \includegraphics[width=\linewidth]{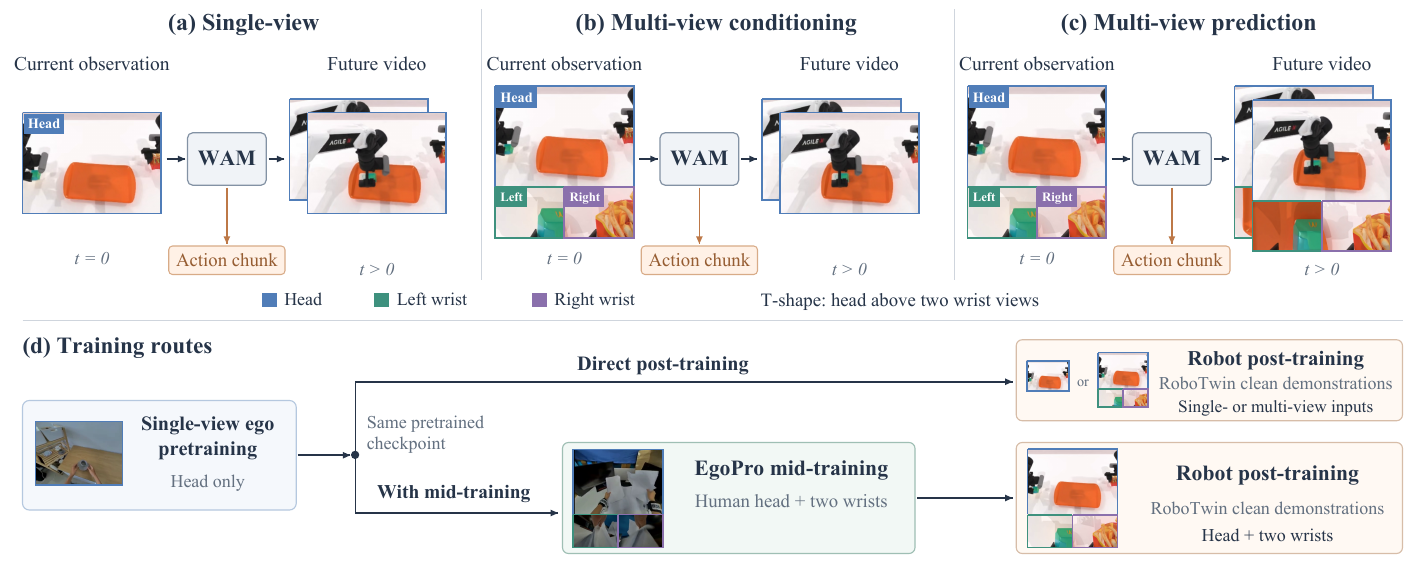}
  \caption{\textbf{From egocentric pretraining to multi-view robot policies.} Three view configurations and training routes with or without EgoPro mid-training.}
  \label{fig:multiview_adaptation}
\end{figure}

\paragraph{Multi-view configurations.}
We compare three configurations for visual conditioning and video prediction, illustrated in \cref{fig:multiview_adaptation}. \emph{Single-view} uses the current head-camera observation and predicts future video from the single view. \emph{Multi-view conditioning} adds current observations from both wrist cameras while restricting future video prediction to the head camera. \emph{Multi-view prediction} uses the same three-camera inputs and predicts future video for all three views. We evaluate these variants separately under isolated attention with relative EEF actions and IDM attention with absolute camera-frame actions. Within each model configuration, all three variants are post-trained on RoboTwin from the same 20k-step egocentric pretraining checkpoint.

\begin{table}[ht]
  \centering
  \caption{\textbf{From egocentric pretraining to multi-view robot policies.} RoboTwin success rates (\%) from the best available complete post-training evaluation for each setting.}
  \label{tab:multiview_configurations}
  \small
  \setlength{\tabcolsep}{5pt}
  \renewcommand{\arraystretch}{1.15}
  \begin{tabular*}{\linewidth}{@{\extracolsep{\fill}}lllrrr@{}}
    \toprule
    Model configuration & View configuration & Mid-training & Clean & Scene-OOD & Random \\
    \midrule
    \multirow[t]{3}{*}{\begin{tabular}[t]{@{}l@{}}Isolated attention\\(relative)\end{tabular}} & Single-view & None & 75.68 & 46.08 & 22.48 \\
                       & Multi-view conditioning & None & 74.60 & 34.08 & 12.92 \\
                       & Multi-view prediction   & None & 75.04 & 33.00 & 14.84 \\
    \midrule
    \multirow[t]{5}{*}{\begin{tabular}[t]{@{}l@{}}IDM attention\\(absolute)\end{tabular}} & Single-view & None & 88.48 & 62.00 & 28.56 \\
                        & \multirow[t]{2}{*}{Multi-view conditioning} & None & 90.56 & 59.52 & 29.08 \\
                        &                         & EgoPro & 90.28 & 61.20 & 30.68 \\
                        & \multirow[t]{2}{*}{Multi-view prediction} & None & 89.84 & 56.28 & 26.16 \\
                        &                         & EgoPro & 88.72 & 58.80 & 27.36 \\
    \bottomrule
  \end{tabular*}
\end{table}

\Cref{tab:multiview_configurations} shows that introducing wrist views during robot post-training under isolated attention degrades the out-of-distribution generalization gained from egocentric pretraining, while clean success decreases only slightly. Relative to \emph{Single-view}, scene-ood success falls from 46.1\% to 33.0--34.1\% across the two multi-view variants, while random success falls from 22.5\% to 12.9--14.8\%. Under IDM attention, \emph{Multi-view conditioning} improves clean success from 88.5\% to 90.6\% and maintains comparable random success, while scene-ood success decreases from 62.0\% to 59.5\%. With the input views unchanged, \emph{Multi-view prediction} lowers success in all three settings relative to \emph{Multi-view conditioning}, with scene-ood decreasing from 59.5\% to 56.3\% and random from 29.1\% to 26.2\%.

\paragraph{Multi-view mid-training.}
Under IDM attention, both multi-view configurations achieve lower scene-ood success than \emph{Single-view} after direct robot post-training. We therefore investigate whether mid-training on EgoPro, which provides human egocentric videos with wrist-camera views, can bridge single-view egocentric pretraining and multi-view robot post-training while preserving the generalization gained from human data. For both \emph{Multi-view conditioning} and \emph{Multi-view prediction}, we compare this intermediate stage followed by robot post-training with direct post-training from the same pretrained checkpoint, keeping each view configuration fixed across mid-training and robot post-training.

\Cref{tab:multiview_configurations} shows that mid-training improves out-of-distribution performance for both view configurations. With \emph{Multi-view conditioning}, scene-ood success increases from 59.5\% to 61.2\% and random success from 29.1\% to 30.7\%. \emph{Multi-view prediction} also improves, from 56.3\% to 58.8\% on scene-ood and from 26.2\% to 27.4\% on random. Clean success changes little with conditioning and decreases with prediction, so the gains are concentrated in out-of-distribution generalization. After mid-training, \emph{Multi-view conditioning} approaches \emph{Single-view} on scene-ood (61.2\% versus 62.0\%) and exceeds it on random (30.7\% versus 28.6\%). It also outperforms \emph{Multi-view prediction} in all three settings, supporting our choice of \emph{Multi-view conditioning} mid-training before robot post-training.

\section{Pure Egocentric Data Foundation Model Training}
\label{sec:model}
Building on the exploration in Section 3, we develop HumanWAM-Vanilla, a world–action model pretrained exclusively on 50,000 hours of curated egocentric human data with aligned action trajectories. HumanWAM-Vanilla achieves competitive performance on the bimanual simulation benchmarks RoboTwin~2.0 and RoboDojo following robot post-training, with strengths in environmental robustness, precise control, and multi-step manipulation. Real-robot experiments further show that egocentric pretraining improves task success over a baseline without egocentric pretraining, with particularly large gains under visual distribution shifts.

\subsection{Model and Training Setup}
\label{sec:model_pretrain}

\paragraph{Model configuration.}
HumanWAM-Vanilla follows the world--action architecture described in \cref{sec:study_model} and uses inverse-dynamics attention, allowing action tokens to attend to both current and future video frames. During training, action denoising is conditioned on ground-truth future video latents with noise augmentation. For policy inference, the model first generates future video latents and then predicts an action chunk conditioned on these latents.

\paragraph{Action representation and normalization.}
We use the unified 16-D bimanual end-effector action space defined in \cref{sec:spec}. Each end effector is represented by its three-dimensional position, quaternion orientation, and gripper openness. EEF poses are expressed as absolute poses in the camera frame at the corresponding target time, with camera and wrist axes aligned across sources. Actions and proprioceptive states are standardized using dimension-wise z-score normalization after action alignment. Pretraining statistics are estimated jointly across the curated corpus, while mid-training and robot post-training use statistics estimated from their respective training datasets. Invalid or unsupervised hand dimensions are excluded from both normalization statistics and the action loss.

\paragraph{Egocentric pretraining.}
We pretrain the model exclusively on Open-HumanWAM-50K curated and aligned as described in \cref{sec:data}, comprising human videos, language annotations, and action trajectories. The action chunk length is 32, and video frames are sampled every four action steps, yielding nine frames per sample. We optimize the model with AdamW~\cite{adamw}, setting the global batch size to 8{,}192 and the weight decay to \(0.01\). The learning-rate schedule spans 100{,}000 optimizer steps, with linear warm-up to \(10^{-4}\) over the first 5{,}000 steps followed by cosine decay to \(10^{-6}\).

\paragraph{Mid-training.}
After egocentric pretraining with a single head-camera view, mid-training adapts the model to multi-view inputs for robot policies. We continue training on EgoPro~\cite{egosuite}, a human egocentric dataset with synchronized video from one head-mounted camera and two wrist-mounted cameras. Under \emph{Multi-view conditioning} (\cref{sec:camera_view_ablation}), current observations from all three cameras condition action prediction, while future video prediction remains restricted to the head camera.

\begin{table}[htbp]
  \centering
  \caption{\textbf{Post-training settings for RoboTwin~2.0, RoboDojo, and real-robot experiments.}}
  \label{tab:model_posttraining}
  \small
  \setlength{\tabcolsep}{5pt}
  \renewcommand{\arraystretch}{1.15}
  \begin{tabular*}{\linewidth}{@{\extracolsep{\fill}}llrr@{}}
    \toprule
    Setting & Robot demonstrations & \shortstack[r]{Optimizer\\steps} & \shortstack[r]{Global\\batch size} \\
    \midrule
    RoboTwin~2.0 & 2500 & 24{,}000 & 512 \\
    RoboDojo & 3500 & 50{,}000 & 512 \\
    Real robot & 50/task & 10{,}000 & 256 \\
    \bottomrule
  \end{tabular*}
\end{table}

\paragraph{Robot post-training.}
Separate policies are post-trained for RoboTwin~2.0, RoboDojo, and real-robot experiments using the corresponding robot demonstrations. \Cref{tab:model_posttraining} summarizes the training data and optimization settings. RoboTwin--Clean2Clean and RoboTwin--Clean2Random use the same post-training data, comprising 50 clean demonstrations per task. Robot EEF poses follow the absolute camera-frame representation. Post-training uses AdamW with a weight decay of $0.01$ and a cosine learning-rate schedule preceded by linear warm-up over the first 5\% of optimization steps.

\subsection{Simulation Experiments}
\label{sec:model_eval}
Since our egocentric pretraining primarily covers bimanual tabletop manipulation, we evaluate HumanWAM-Vanilla on RoboTwin~2.0 and RoboDojo. Comparisons with existing vision--language--action models and world--action models assess robustness to environmental variation and capabilities such as precision, memory, and long-horizon execution.
\paragraph{RoboTwin 2.0.}
\begin{table*}[t]
  \centering
  \caption{\textbf{Evaluation results on RoboTwin 2.0.} Success rates (\%) are reported after clean-only post-training. Average is the mean of Clean2Clean and Clean2Random. The best result in each column is highlighted in bold.}
  \label{tab:model_robotwin_results}
  \small
  \setlength{\tabcolsep}{7pt}
  \renewcommand{\arraystretch}{1.12}
  \begin{tabular*}{\textwidth}{@{}l@{\hspace{1.5em}}l@{\extracolsep{\fill}}rrr@{}}
    \toprule
    Type & Method & Clean2Clean $\uparrow$ & Clean2Random $\uparrow$ & Average $\uparrow$ \\
    \midrule
    \multirow{7}{*}{\color{black!65}\sffamily VLA} & Qwen-RobotManip-Context~\cite{yuan2026qwen} & 84.7 & 69.4 & 77.1 \\
     & GigaBrain-0.7~\cite{gigabrain07} & 66.80 & 67.92 & 67.36 \\
     & $\pi_{0.5}$~\cite{intelligence2025pi_} & 70.70 & 46.00 & 58.35 \\
     & Spatial Forcing~\cite{spatialforcing} & 77.20 & 26.74 & 51.97 \\
     & X-VLA~\cite{xvla} & 68.00 & 20.90 & 44.45 \\
     & Abot-M0~\cite{abotm0} & 57.40 & 30.40 & 43.90 \\
     & GR00T-N1.7~\cite{gr00tn17} & 43.6 & 20.7 & 32.2 \\
    \midrule
    \multirow{12}{*}{\color{black!65}\sffamily WAM} & GO-3 Nano Preview & 88.54 & \textbf{81.48} & \textbf{85.01} \\
     & PICA-WAM~\cite{picawam} & 90.68 & 74.70 & 82.69 \\
     & CloudRobo-WM & 88.90 & 75.00 & 81.95 \\
     & MWA-Phoenix~\cite{phoenixmwa} & 86.54 & 72.42 & 79.48 \\
     & PatchWAM~\cite{patchwam} & 91.56 & 66.72 & 79.14 \\
     & ME-Dex-1.0~\cite{medex} & 89.58 & 68.12 & 78.85 \\
     & EWAM~\cite{ewam} & 82.16 & 72.08 & 77.12 \\
     & ViGAR~\cite{gong2026rethinking} & 82.00 & 67.02 & 74.51 \\
     & UniWAM~\cite{uniwam} & 75.12 & 67.56 & 71.34 \\
     & OpenWAM-$\alpha$~\cite{openwam} & 89.4 & 48.7 & 69.0 \\
     & 4D-WAM~\cite{wam4d} & 81.50 & 41.80 & 61.65 \\
    \cmidrule(l){2-5}
     & \textbf{HumanWAM-Vanilla (Ours)} & \textbf{92.08} & 68.88 & 80.48 \\
    \bottomrule
  \end{tabular*}
  \par\vspace{4pt}
  \begin{minipage}{\textwidth}
    \footnotesize
    Baseline results are taken from the \href{https://robotwin-platform.github.io/leaderboard}{official RoboTwin~2.0 leaderboard}~\cite{robotwin_leaderboard} (Co-train track, without observed depth at inference), except for GR00T-N1.7, Qwen-RobotManip-Context, and OpenWAM-$\alpha$, whose published results are sourced from InternW0-$\Delta$~\cite{internw0delta}. HumanWAM-Vanilla results are obtained from our own evaluations.
  \end{minipage}
\end{table*}

RoboTwin~2.0~\cite{robotwin2} is a benchmark built on SAPIEN~\cite{sapien} comprising 50 bimanual tabletop manipulation tasks on the Aloha-AgileX embodiment, including pick-and-place, stacking, object handovers, and articulated-object manipulation. We jointly post-train a single policy on clean demonstrations from all tasks and evaluate it in two settings. \emph{Clean2Clean} evaluates performance within the demonstration environment distribution, whereas \emph{Clean2Random} randomizes background and tabletop textures, lighting, clutter, and table height.

At inference, the policy receives RGB observations from the head camera and both wrist cameras, together with the language instruction and proprioceptive state. It predicts 32-step action chunks and executes the first 24 steps before replanning from updated observations. We evaluate the same checkpoint in both settings, with 100 episodes per task, totaling 5{,}000 episodes per setting. We report the mean success rate across the 50 tasks in each setting and the average of these two means.

\Cref{tab:model_robotwin_results} shows that HumanWAM-Vanilla achieves the highest Clean2Clean success rate among the compared methods, reaching 92.08\% after post-training on 50 clean robot demonstrations per task. The same policy achieves 68.88\% on Clean2Random, yielding an average success rate of 80.48\% that exceeds all VLA baselines and most WAM baselines in the table. The remaining gap to the leading WAMs lies in out-of-distribution performance, with GO-3 Nano Preview reaching 81.48\% on Clean2Random. This gap may partly reflect architectural differences, as HumanWAM-Vanilla uses a basic world--action architecture without additional specialized modules. These results support pretraining exclusively on human egocentric data as an effective foundation for bimanual robot manipulation, with accurate task execution and competitive generalization to visual conditions absent from the robot demonstrations.

\paragraph{RoboDojo.}
RoboDojo~\cite{robodojo} is an Isaac Sim-based benchmark with 42 bimanual manipulation tasks covering generalization, precision, memory, long-horizon execution, and open-vocabulary instruction following. We compare HumanWAM-Vanilla against representative VLA and WAM baselines.

We post-train on the official set of 3{,}500 robot demonstrations, comprising 100 demonstrations for each of 34 training tasks and 100 auxiliary domain-randomized demonstrations. The Open tasks are excluded from post-training. Generalization is evaluated separately under standard (Gen-Std) and randomized (Gen-Rand) conditions. We evaluate the same checkpoint across all six groups and report success rate, which measures complete task success, and progress score (0--100), which credits partial completion. 

\begin{table}[th]
  \centering
  \caption{\textbf{RoboDojo comparison with VLA and WAM baselines.} Each entry reports success rate (\%) / progress score. Mean is recomputed for every method as an equally weighted average over the six groups.}
  \label{tab:model_robodojo_results}
  \small
  \setlength{\tabcolsep}{3pt}
  \renewcommand{\arraystretch}{1.16}
  \begin{adjustbox}{max width=\linewidth}
  \begin{tabular}{@{}ll*{7}{r}@{}}
    \toprule
    Type & Method & Gen-Std & Gen-Rand & Precision & \shortstack{Long-\\Horizon} & Memory & Open & Mean \\
    \midrule
    \multirow{8}{*}{\color{black!65}\sffamily VLA} & StarVLA-$\alpha$~\cite{starvlaalpha} & 5.00\,/\,7.54 & 0.00\,/\,0.33 & 4.33\,/\,9.90 & 6.50\,/\,14.15 & 2.44\,/\,3.34 & 0.58\,/\,0.68 & 3.14\,/\,5.99 \\
     & X-VLA~\cite{xvla} & 12.00\,/\,17.90 & 1.00\,/\,3.04 & 12.00\,/\,18.32 & 9.75\,/\,16.53 & 3.56\,/\,4.76 & 0.50\,/\,0.55 & 6.47\,/\,10.18 \\
     & $\pi_{0.5}$~\cite{intelligence2025pi_} & 15.00\,/\,20.93 & 1.00\,/\,5.82 & 5.50\,/\,12.40 & 14.67\,/\,23.54 & 4.56\,/\,5.78 & 1.67\,/\,1.98 & 7.07\,/\,11.74 \\
     & Spatial Forcing~\cite{spatialforcing} & 15.00\,/\,21.25 & 4.00\,/\,6.98 & 10.58\,/\,17.33 & 14.58\,/\,23.26 & 4.11\,/\,5.43 & 1.58\,/\,1.78 & 8.31\,/\,12.67 \\
     & Hy-Embodied-0.5-VLA~\cite{zhang2026hy} & 17.00\,/\,21.98 & 0.00\,/\,1.57 & 8.00\,/\,13.81 & 14.92\,/\,25.74 & 12.11\,/\,13.37 & 0.58\,/\,0.65 & 8.77\,/\,12.85 \\
     & Xiaomi-Robotics-1~\cite{xiaomirobotics1} & 28.00\,/\,35.65 & 6.00\,/\,11.44 & 18.83\,/\,26.69 & 23.67\,/\,38.39 & 6.56\,/\,7.81 & 3.58\,/\,3.94 & 14.44\,/\,20.65 \\
     & Galaxea G0.5~\cite{galaxeag05} & 20.00\,/\,26.74 & 6.00\,/\,11.16 & 20.42\,/\,28.25 & 32.25\,/\,44.12 & 7.33\,/\,8.61 & 1.58\,/\,1.73 & 14.60\,/\,20.10 \\
     & DM0.5~\cite{dm05} & 18.00\,/\,23.49 & 4.00\,/\,8.06 & 16.75\,/\,24.82 & 19.50\,/\,33.70 & 47.44\,/\,47.74 & 2.08\,/\,2.43 & 17.96\,/\,23.37 \\
    \midrule
    \multirow{6}{*}{\color{black!65}\sffamily WAM} & Fast-WAM~\cite{yuan2026fastwamworldactionmodels} & 2.00\,/\,4.33 & 0.00\,/\,0.34 & 0.00\,/\,1.96 & 5.17\,/\,9.14 & 3.44\,/\,3.55 & 0.42\,/\,0.42 & 1.84\,/\,3.29 \\
     & AHA-WAM~\cite{ahawam} & 6.00\,/\,10.32 & 0.00\,/\,1.26 & 2.42\,/\,5.86 & 2.67\,/\,8.61 & 2.78\,/\,2.97 & 0.83\,/\,0.88 & 2.45\,/\,4.98 \\
     & GigaWorld-Policy~\cite{gigaworldpolicy} & 6.00\,/\,10.28 & 0.00\,/\,0.41 & 1.83\,/\,6.15 & 8.92\,/\,15.51 & 2.22\,/\,3.46 & 0.50\,/\,0.54 & 3.25\,/\,6.06 \\
     & X-WAM~\cite{xwam} & 5.00\,/\,11.24 & 1.00\,/\,3.54 & 1.83\,/\,6.72 & 9.08\,/\,17.47 & 4.67\,/\,6.32 & 0.25\,/\,0.57 & 3.64\,/\,7.64 \\
     & OpenWAM-$\alpha$~\cite{openwam} & 25.56\,/\,33.16 & 4.11\,/\,8.26 & 9.25\,/\,18.45 & 25.33\,/\,34.93 & 9.11\,/\,10.41 & 1.08\,/\,1.41 & 12.41\,/\,17.77 \\
    \cmidrule(l){2-9}
     & {\bfseries HumanWAM-Vanilla} & {\bfseries 22.33\,/\,29.10} & {\bfseries 1.67\,/\,3.25} & {\bfseries 20.50\,/\,28.83} & {\bfseries 33.00\,/\,44.00} & {\bfseries 6.00\,/\,7.00} & {\bfseries 1.50\,/\,1.80} & {\bfseries 14.17\,/\,19.00} \\
    \bottomrule
  \end{tabular}
  \end{adjustbox}
\end{table}

\Cref{tab:model_robodojo_results} shows that HumanWAM-Vanilla achieves a mean success rate of 14.17\% and a mean progress score of 19.00. It exceeds the WAM baselines in our comparison on both aggregate metrics and approaches the overall performance of Galaxea G0.5. Its advantage over OpenWAM-\(\alpha\) is most pronounced on Precision and Long-Horizon, where success reaches 20.50\% and 33.00\%, compared with 9.25\% and 25.33\%, respectively. These results show that pretraining exclusively on human egocentric data, followed by adaptation with 100 robot demonstrations per training task, can support competitive robot manipulation, particularly in tasks requiring precise control and multi-step execution.

Despite these gains, HumanWAM-Vanilla remains below the strongest recent models and agent-based systems on RoboDojo. Beyond manipulation skills, some RoboDojo tasks require memory of earlier observations or arithmetic reasoning, such as addition and subtraction. However, HumanWAM-Vanilla is built on a basic world--action model architecture similar to those of Fast-WAM~\cite{yuan2026fastwamworldactionmodels} and OpenWAM-\(\alpha\)~\cite{openwam}, without dedicated memory or reasoning modules. We attribute a substantial part of the performance gap to these architectural limitations. We retain this basic architecture as a common foundation for studying egocentric pretraining and deriving general scaling principles.

\subsection{Real-Robot Experiments}
\label{sec:model_real}

\begin{figure}[tb]
  \centering
  \includegraphics[width=\linewidth,height=0.62\textheight,keepaspectratio]{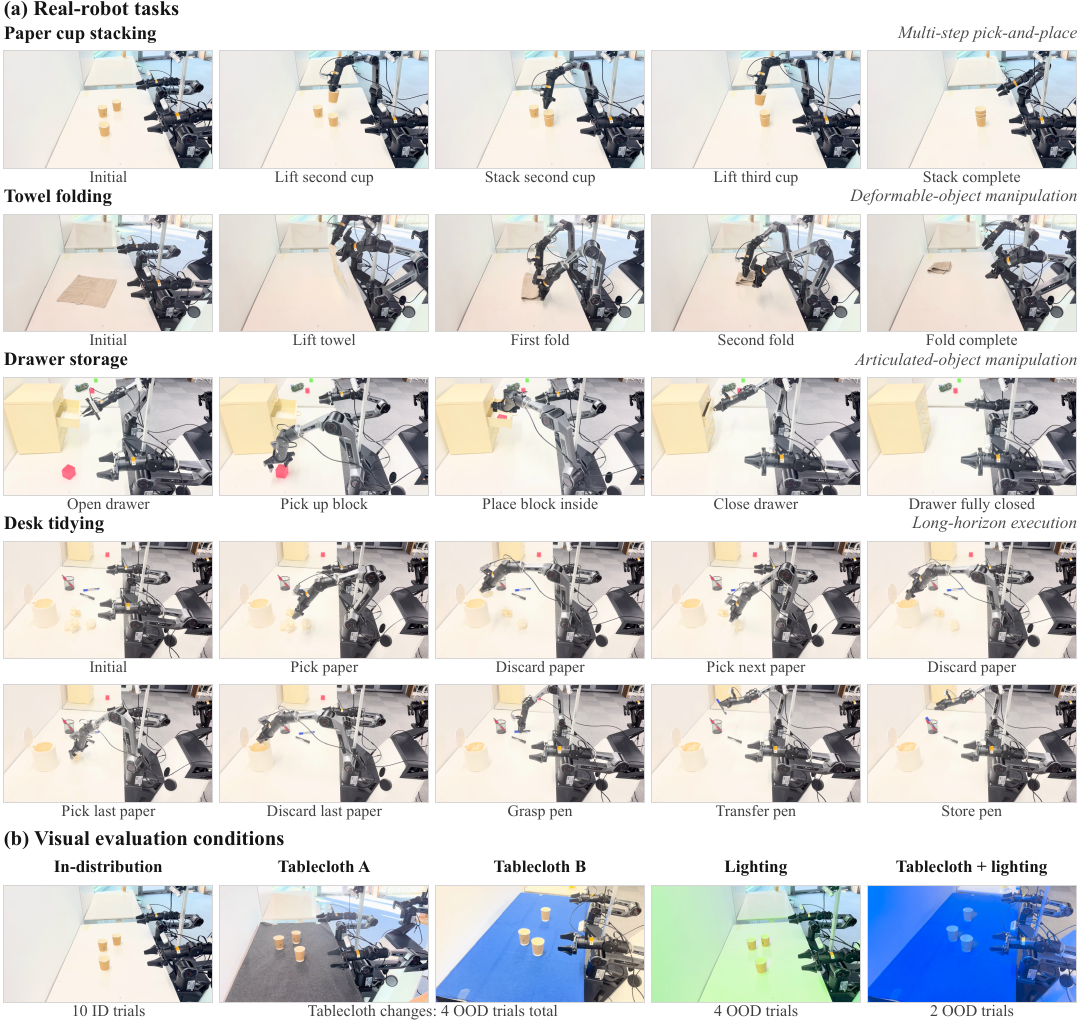}
  \caption{\textbf{Real-robot tasks and visual evaluation conditions.} }
  \label{fig:model_real_robot_tasks}
\end{figure}

\paragraph{Setup.}
To assess whether the generalization benefits observed in simulation extend to physical robots, we evaluate HumanWAM-Vanilla on paper cup stacking, towel folding, drawer storage, and desk tidying with a Cobot Magic bimanual robot, as shown in \cref{fig:model_real_robot_tasks}. These tasks cover multi-step pick-and-place, deformable-object manipulation, articulated-object manipulation, and long-horizon execution. We compare HumanWAM-Vanilla with a No-Ego baseline that omits egocentric pretraining. Both models are post-trained on 50 demonstrations per task for 10{,}000 optimizer steps with a global batch size of 256.

We evaluate robustness to visual distribution shift by changing the tablecloth and lighting, individually and in combination, relative to the demonstration environment. For each task and model, we conduct 10 in-distribution trials under the visual conditions of the training demonstrations and 10 out-of-distribution trials, allocating four to each individual change and two to their combination. We report per-task success rates and their mean across the four tasks separately for each setting.

\begin{table}[tb]
  \centering
  \caption{\textbf{Real-robot task success with and without egocentric pretraining.} Success rates are reported in percent over 10 trials per task, model, and evaluation setting. Mean averages the four task success rates.}
  \label{tab:model_real_robot_results}
  \small
  \setlength{\tabcolsep}{6pt}
  \renewcommand{\arraystretch}{1.15}
  \begin{tabular*}{\linewidth}{@{\extracolsep{\fill}}lrrrr@{}}
    \toprule
    & \multicolumn{2}{c}{In-distribution} & \multicolumn{2}{c}{Out-of-distribution} \\
    \cmidrule(lr){2-3}\cmidrule(l){4-5}
    Task & No-Ego & HumanWAM-Vanilla & No-Ego & HumanWAM-Vanilla \\
    \midrule
    Paper cup stacking & 70.0 & \textbf{90.0} & 0.0 & \textbf{80.0} \\
    Towel folding & 50.0 & \textbf{80.0} & 10.0 & \textbf{60.0} \\
    Drawer storage & 60.0 & \textbf{90.0} & 0.0 & \textbf{70.0} \\
    Desk tidying & 50.0 & \textbf{90.0} & 0.0 & \textbf{60.0} \\
    \midrule
    Mean & 57.5 & \textbf{87.5} & 2.5 & \textbf{67.5} \\
    \bottomrule
  \end{tabular*}
\end{table}

\paragraph{Results.}
HumanWAM-Vanilla achieves higher success rates than No-Ego on all four tasks in both settings (\cref{tab:model_real_robot_results}). Mean in-distribution success increases from 57.5\% to 87.5\%. These gains span deformable-object manipulation and long-horizon execution, with success increasing from 50\% to 80\% on towel folding and from 50\% to 90\% on desk tidying. Success on drawer storage increases from 60\% to 90\%.

The gap widens under visual distribution shift. HumanWAM-Vanilla achieves 60--80\% success across the four tasks, whereas No-Ego succeeds in only one of the 40 out-of-distribution trials. Mean out-of-distribution success increases from 2.5\% to 67.5\%, a gain of 65.0\,pp compared with 30.0\,pp in distribution. These results show that the generalization benefits of egocentric pretraining extend to physical robots, with improved task execution under visual conditions absent from the robot demonstrations.

\subsection{Ablation Studies}
\label{sec:model_ablation}
To improve robot policy performance after post-training, we refine the model and training configuration used in the egocentric pretraining exploration. HumanWAM-Vanilla adopts IDM attention and absolute camera-frame actions in place of the default isolated attention and relative actions described in \cref{sec:study_model}. We examine these choices, together with action-expert initialization, through ablations on the four-source, 1{,}000-hour corpus described in \cref{sec:study3}. All models are pretrained for 20{,}000 steps and post-trained for 12{,}000 steps on 50 clean demonstrations per RoboTwin task.

\paragraph{Attention strategies.}
World--action models differ in how video and action tokens exchange information. We compare four attention strategies, each applied consistently throughout egocentric pretraining and robot post-training. \emph{Isolated attention} restricts visual conditioning for action prediction to the current frame. \emph{Mutual attention} allows video and action tokens to attend to each other during joint denoising. \emph{IDM attention} conditions action denoising on ground-truth future video latents during training and generated future latents at inference. \emph{Optional-IDM}, following the variant released by Fast-WAM~\cite{fastwam_code}, trains a single model to predict actions with or without future video conditioning. Our implementation uses ground-truth future-video conditioning with probability 0.5 and current-frame-only conditioning otherwise.

\begin{table}[htbp]
  \centering
\caption{\textbf{Ablation of attention strategies and action representations.}}
  \label{tab:model_attention_ablation}
  \small
  \setlength{\tabcolsep}{5pt}
  \renewcommand{\arraystretch}{1.15}
  \begin{tabular*}{\linewidth}{@{\extracolsep{\fill}}lllrrr@{}}
    \toprule
    Attention & Action representation & Inference & Clean & Scene-OOD & Random \\
    \midrule
    IA & Relative & Action-only & 76.28 & 46.36 & 22.40 \\
    Mutual & Relative & Joint denoising & 81.28 & 57.96 & 32.92 \\
    IDM & Relative & Video $\rightarrow$ action & 83.56 & 58.08 & 27.56 \\
    Mutual & Absolute camera-frame & Joint denoising & 84.88 & 61.12 & 30.68 \\
    \textbf{IDM} & \textbf{Absolute camera-frame} & \textbf{Video $\rightarrow$ action }& \textbf{88.48 }& \textbf{62.00} & \textbf{28.56} \\
    \midrule
    Optional-IDM & Relative & First-frame only & 73.44 & 48.44 & 23.40 \\
    Optional-IDM & Relative & Video $\rightarrow$ action & 81.96 & 57.04 & 28.04 \\
    \bottomrule
  \end{tabular*}
\end{table}

\Cref{tab:model_attention_ablation} shows that mutual and IDM attention improve clean and scene-ood success over isolated attention with relative actions held fixed. Scene-ood success rises from 46.36\% to 57.96\% and 58.08\%, respectively. IDM attention also improves random success from 22.40\% to 27.56\%. For Optional-IDM, enabling future video generation at inference raises clean success from 73.44\% to 81.96\% and scene-ood success from 48.44\% to 57.04\% using the same checkpoint. These results indicate that generating future video before predicting actions improves both in-domain fitting and generalization under distribution shift.

\paragraph{Action parameterization.}
Embodied pretraining methods adopt different action representations. Relative EEF motions are commonly used to reduce dependence on global coordinate frames and facilitate cross-embodiment transfer~\cite{umi,egoscale}. We compare relative and absolute camera-frame actions, retaining each representation throughout egocentric pretraining and robot post-training. Relative actions express target EEF poses with respect to the EEF pose at the current observation, whereas absolute actions express them directly in the camera frame at each target time.

\Cref{tab:model_attention_ablation} shows that absolute camera-frame actions improve clean and scene-ood performance over relative actions under both mutual and IDM attention. Under IDM attention, clean success increases from 83.56\% to 88.48\% and scene-ood success from 58.08\% to 62.00\%, with a smaller improvement on random. With mutual attention, clean success increases from 81.28\% to 84.88\% and scene-ood success from 57.96\% to 61.12\%.

\paragraph{Action-expert initialization.}
We further examine whether to initialize the action expert randomly or from linearly interpolated pretrained Wan2.2 weights before egocentric pretraining. Both variants start from the same pretrained video backbone and are pretrained for 20{,}000 steps on the same four-source, 1{,}000-hour corpus. They use IDM attention and absolute camera-frame actions throughout pretraining and post-training, with identical photometric augmentation during the 12{,}000-step robot post-training stage. We observed that initializing the action expert from video-model weights does not improve downstream success in this comparison (\cref{tab:model_action_init_ablation}). The two variants achieve comparable clean and scene-ood success, while random initialization increases random success from 51.52\% to 54.12\%.

\begin{table}[htbp]
  \centering
\caption{\textbf{Ablation of action-expert initialization.}}
  \label{tab:model_action_init_ablation}
  \small
  \setlength{\tabcolsep}{5pt}
  \renewcommand{\arraystretch}{1.15}
  \begin{tabular*}{\linewidth}{@{\extracolsep{\fill}}lrrr@{}}
    \toprule
    Action-expert initialization & Clean & Scene-OOD & Random \\
    \midrule
    Interpolated Wan2.2 weights & 89.56 & 83.20 & 51.52 \\
    Random initialization & 90.28 & 83.92 & 54.12 \\
    \bottomrule
  \end{tabular*}
\end{table}

\section{Pushing the Model Capability Boundary with Hybrid Data Scaling}
\label{sec:human_robot_scaling}

Scaling pure egocentric pretraining to 50k hours instills strong manipulation priors, substantially improving out-of-distribution generalization after task-specific post-training. However, purely human pretraining leaves an inherent embodiment gap: without exposure to physical robot morphology and control dynamics, the model cannot be deployed directly and remains confined to a finetuning paradigm that requires demonstrations for every target task. Conversely, while large-scale robot datasets~\cite{lingbotvla} provide direct, executable control supervision, scaling teleoperated trajectories alone quickly plateaus due to collection cost and limited scene diversity. To overcome these single-modality bottlenecks, we investigate \textbf{joint co-training on human and robot data} within an extended unified action space. Through systematic scaling experiments across progressive human and robot data budgets, we characterize how hybrid data scaling expands the boundary of model generalization and execution capability.

A central question is whether scaling hybrid training experience reduces reliance on task-specific adaptation. We therefore evaluate our co-trained models across two complementary protocols. In the \textit{zero-shot protocol}, the pretrained checkpoint is deployed directly in simulation and on physical robots without target-task demonstrations or parameter updates, testing whether larger hybrid budgets expand the repertoire of tasks executable out of the box. In the \textit{post-training protocol}, the same checkpoints are finetuned on target-task demonstrations, assessing whether hybrid pretraining continues to raise the performance ceiling once in-domain data are available, particularly under out-of-distribution shifts.

Deploying a generalist policy without task-specific finetuning also requires an unambiguous interface to specify unfamiliar tasks. Because natural language instructions often leave object configurations, spatial relations, and terminal states underspecified, we augment our world-action model with explicit goal-image conditioning~\cite{gong2026rethinking}. The resulting model, \textbf{HumanWAM-Goal}, jointly predicts future visual states and executable actions from current observations and a visual goal. Beyond disambiguating task intent for zero-shot execution, this interface enables embodied \textbf{in-context learning}~\cite{liu2026roboicl}, steering a frozen policy toward distinct manipulation outcomes purely by varying the visual goal at inference time. 

To realize this framework, we first introduce our robot pretraining corpus (\cref{ssec:robot_data}), extend the unified action representation (\cref{ssec:extended_action}), and detail the HumanWAM-Goal architecture (\cref{ssec:goal_wam}) alongside its co-training recipe (\cref{ssec:goalwam_training_recipe}). We then formalize our evaluation protocols (\cref{ssec:eval_protocol}), examine zero-shot and post-trained scaling trends (\cref{ssec:goalwam_results}), and finally demonstrate emergent zero-shot execution, visual in-context learning, and long-horizon manipulation on physical robots in \cref{sec:real_world_icl}.

\subsection{Robot Pretraining Data}
\label{ssec:robot_data}
\begin{figure}[!htbp]
  \centering
  \includegraphics[width=0.8\linewidth]{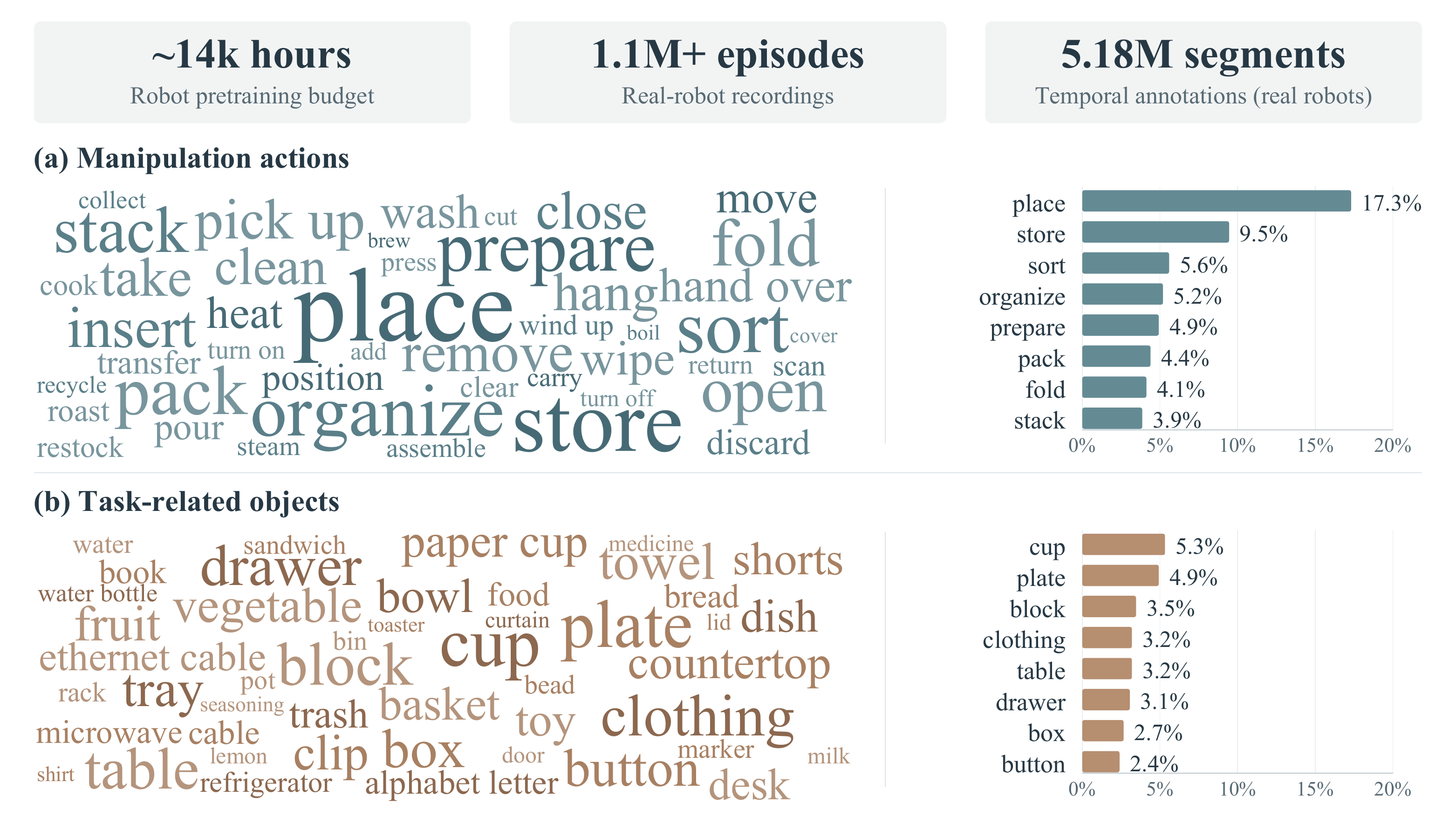}
  \caption{\textbf{Robot pretraining data overview.}
  The top row reports the overall robot pretraining budget.
  Word clouds summarize selected (a) action and (b) object terms, while the adjacent bars show label-level percentages for the eight most frequent terms in each group.
  Percentages use all 751 normalized task-collection labels as the denominator.}
  \label{fig:robot_pretraining_data}
\end{figure}

To complement egocentric human experience with embodiment-specific control supervision, we assemble approximately 14.6k hours of robot pretraining data spanning multiple embodiments.
The real-robot portion contains over 1.1 million episodes across eight recorded robot domains, providing diverse control experience for cross-embodiment pretraining.
\Cref{fig:robot_pretraining_data} summarizes the overall training scale and the action and object vocabulary of the real-robot recordings.

\paragraph{Task diversity.}
The corpus spans activities such as organizing books and tools, folding clothing, preparing food, operating appliances, and inserting cables.
To characterize this diversity, we extract action and object terms from the task-collection labels.
Removing administrative identifiers and standardizing surface forms reduces 864 original labels to 751 normalized labels.
Among the extracted action terms, \textit{place}, \textit{fold}, and \textit{insert} occur in 130, 31, and 20 normalized labels, respectively.
The object vocabulary includes tableware, garments, containers, tools, and appliances; for example, \textit{cup}, \textit{plate}, and \textit{clothing} occur in 40, 37, and 24 labels.
These statistics characterize the range of activities and interaction objects described by the task labels, complementing the aggregate data scale.

\subsection{Hybrid Action Representation}
\label{ssec:extended_action}

To accommodate multi-embodiment robot control and mobile manipulation alongside human videos, we extend the 16-D bimanual action space of \cref{sec:spec} into a 49-D hybrid representation:
\begin{equation}
    \mathbf{a}^{\mathrm{ext}}_t =
    \big[
        \underbrace{\mathbf{j}_t}_{\text{joint: }23},\;
        \underbrace{\mathbf{b}_t}_{\text{chassis: }3},\;
        \underbrace{\mathbf{a}_t,\,\mathbf{p}^{H}_t,\,\mathbf{q}^{H}_t}_{\text{EEF and head: }23}
    \big]
    \in \mathbb{R}^{49}.
\end{equation}
Here, $\mathbf{j}_t \in \mathbb{R}^{23}$ encodes robot joint states in a canonical layout spanning the arms, grippers, torso, and head, with unused degrees of freedom zero-padded and masked; $\mathbf{b}_t=(x_t,y_t,\psi_t) \in \mathbb{R}^3$ represents planar base translation and yaw; and $\mathbf{a}_t\in\mathbb{R}^{16}$ is the unified bimanual EEF action. The 7-D head pose $(\mathbf{p}^{H}_t,\mathbf{q}^{H}_t)$ records the head-camera extrinsics used to anchor bimanual coordinates; within the network interface, these slots serve strictly as geometric metadata and are masked from both state conditioning and action prediction losses.

\paragraph{Metric gripper aperture.}
Whereas \cref{sec:spec} normalizes gripper openness per dataset, zero-shot deployment requires a consistent physical scale across robots and humans. We therefore redefine the gripper dimensions in $\mathbf{a}_t$ as the absolute physical aperture in meters ($0$ denoting fully closed), normalized by a universal reference width of $0.12$\,m. For human videos, we use the metric 3D distance between the estimated thumb and index fingertips directly as the physical aperture, aligning human hand opening and robot gripper width on a unified metric scale.

\paragraph{Reference frames.}
While absolute camera-frame EEF poses excel when post-training adapts normalization statistics to a single target embodiment (\cref{sec:model_ablation}), heterogeneous robot platforms exhibit substantial variations in camera mounting geometry, arm reach, and workspace offsets that impede direct zero-shot transfer. To eliminate these embodiment-specific spatial biases during multi-domain co-training and zero-shot deployment, we parameterize EEF motions as chunk-relative displacements anchored to the current observation. Specifically, given an observation at time $t$, EEF translation targets at each chunk step $\tau$ are expressed as displacements from the current EEF position projected into the anchor head-camera frame at time $t$, $\Delta\mathbf{p}(\tau\mid t)=\mathbf{R}^{H}(t)^{\top}(\mathbf{p}(\tau)-\mathbf{p}(t))$, ensuring a stationary, embodiment-agnostic reference frame across the entire chunk even under camera motion. Similarly, EEF orientation targets are expressed as relative quaternions with respect to the wrist orientation at time $t$, and chassis targets $\mathbf{b}_\tau$ are expressed relative to the current base pose.

\paragraph{Hybrid supervision and joint dropout.}
Human and robot trajectories share common input and output projections for the 16-D EEF and 3-D chassis subspaces, whereas $\mathbf{j}_t$ uses embodiment-specific projections. Per-source validity masks exclude unavailable dimensions (such as joints and chassis in human videos) from the action loss. During robot training, we additionally apply stochastic dropout to the joint branch $\mathbf{j}_t$, preventing the model from over-relying on embodiment-specific joint proprioception and ensuring that the shared 16-D EEF pathway remains self-sufficient. Because $\mathbf{a}_t$ parameterizes camera-frame metric displacements and physical gripper widths in an embodiment-agnostic coordinate frame, its global pretraining normalization statistics transfer directly to unseen robots at test time without target-domain recalibration.

\begin{figure}[ht]
  \centering
  \includegraphics[width=\linewidth]{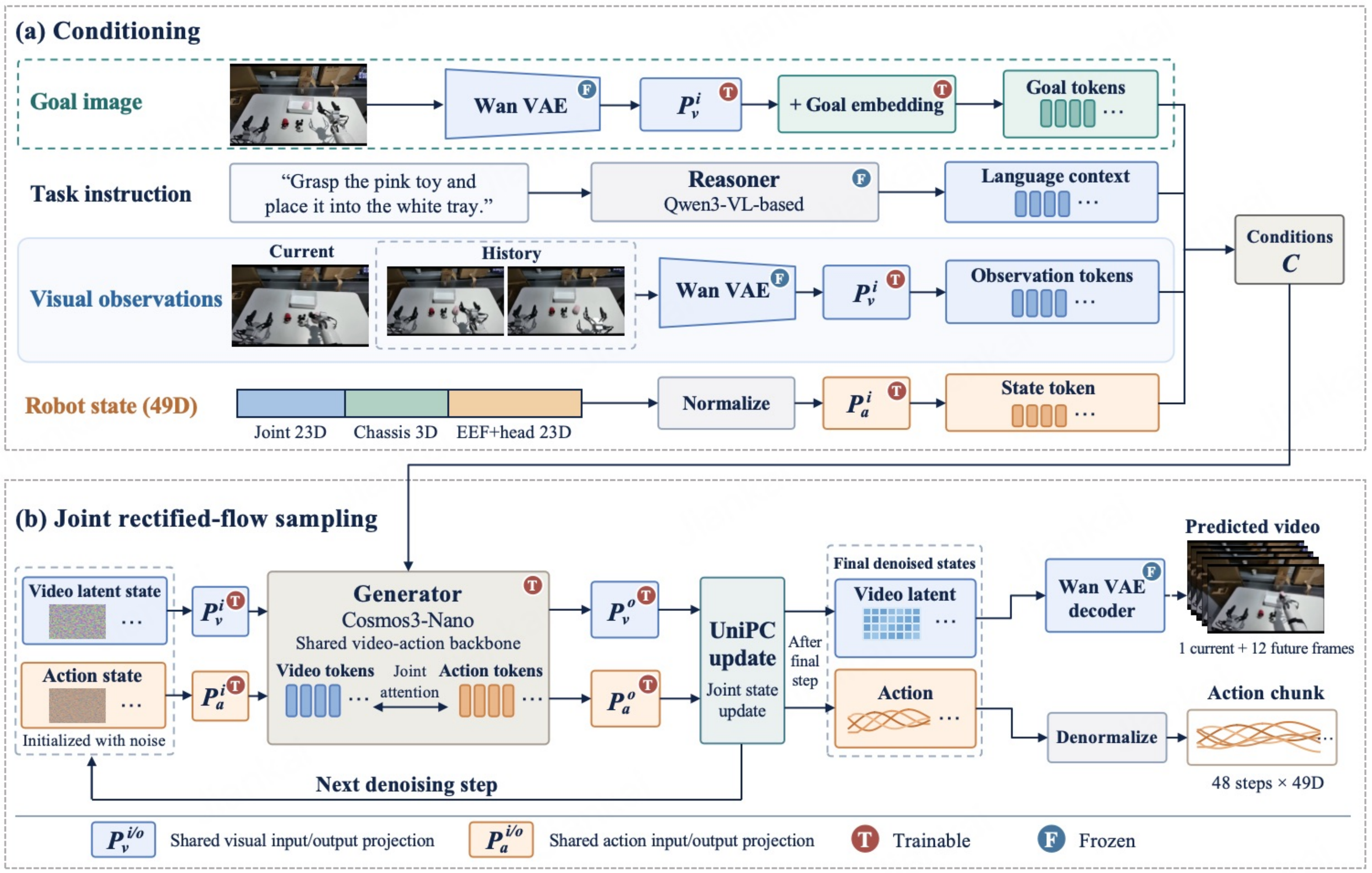}
  \caption{\textbf{HumanWAM-Goal architecture.}
  (a) \textbf{Conditioning.} An explicit goal image augments language instructions, current and optional historical observations, and the 49-D robot state.
  (b) \textbf{Joint rectified-flow sampling.} With conditioning tokens held fixed, the shared video-action Generator iteratively predicts video and action velocities via UniPC updates.}
  \label{fig:goalwam_architecture}
\end{figure}

\subsection{HumanWAM-Goal Architecture}
\label{ssec:goal_wam}

To serve as a general-purpose action executor for multimodal agents~\cite{si2026fewer,liu2026roboicl}, a foundation policy must translate explicit target outcomes into continuous physical control without task-specific parameter updates. We therefore introduce \textbf{HumanWAM-Goal} (\cref{fig:goalwam_architecture}), which augments world-action modeling with explicit visual-goal conditioning~\cite{gong2026rethinking}. Given a language instruction $\ell$, current observation $\mathbf{o}_t$, proprioceptive state $\mathbf{s}_t$, optional observation history $\mathcal{H}_t$, and a goal image $\mathbf{g}$ specifying a desired intermediate or terminal scene configuration, the model estimates
\begin{equation}
    p_{\theta}\!\left(\mathbf{O}^{+}_t, \mathbf{A}_t \mid \mathbf{c}_t\right),
    \qquad
    \mathbf{c}_t =
    \left(\ell,\mathbf{o}_t,\mathbf{s}_t,\mathcal{H}_t,\mathbf{g}\right),
\end{equation}
where $\mathbf{O}^{+}_t$ denotes $12$ future video frames and $\mathbf{A}_t \in \mathbb{R}^{48 \times 49}$ is a $48$-step action chunk in the hybrid action space of \cref{ssec:extended_action}. We build HumanWAM-Goal upon the pretrained Cosmos3-Nano-16B backbone~\cite{nvidia2026cosmos3}, which pairs a frozen Qwen3-VL-based \textit{Reasoner} for language encoding with a shared video-action \textit{Generator}.

\paragraph{Visual goal and observation conditioning.}
To enable unambiguous task specification without modifying the backbone topology, all visual inputs are encoded into spatial latents via a frozen Wan VAE and mapped into the Generator through a shared visual projection $P^{i}_v$. While the current observation $\mathbf{o}_t$ serves as the clean anchor frame of the rollout, the goal image $\mathbf{g}$ is encoded separately, augmented with a learnable goal embedding, and supplied as clean conditioning tokens. This design preserves the fine-grained spatial structure of the desired outcome while distinguishing goal targets from current observations. When historical frames $\mathcal{H}_t$ are included, they enter as additional clean visual tokens with temporal rotary position embeddings (RoPE) reflecting their elapsed time relative to $\mathbf{o}_t$.

\paragraph{Hybrid state-action interface and inference.}
In language-only world-action modeling (\cref{sec:model_ablation}), where terminal states are visually underspecified, two-stage inverse-dynamics (IDM) inference---generating complete future video latents prior to action denoising---is essential to ground control in predicted visual dynamics, albeit at double the sequential sampling cost. In HumanWAM-Goal, because the clean goal tokens $\mathbf{g}$ directly anchor the target spatial configuration, we replace two-stage cascading with single-stage \textit{joint rectified-flow sampling} within the shared-parameter Generator (\cref{fig:goalwam_architecture}). Specifically, video and action tokens share Transformer weights and interact bidirectionally at every denoising step through joint self-attention via modality-specific projections $P^{i/o}_v$ and $P^{i/o}_a$. We interface the 49-D hybrid representation (\cref{ssec:extended_action}) by prepending the normalized current state $\mathbf{s}_t$ as a single clean conditioning row ahead of the $48$ noisy future action rows, restricting the rectified-flow loss strictly to future actions. At inference, holding all conditioning tokens $\mathbf{c}_t$ fixed, the Generator synchronously denoises future video latents and the 48-step action chunk via iterative UniPC updates, preserving world-to-action coupling while halving sampling latency and optionally bypassing the Wan VAE decoder for real-time control.

\subsection{Training Recipe}\label{ssec:goalwam_training_recipe}

\paragraph{Training data.}
We investigate human--robot data scaling through two complementary series. In the \textit{robot-only series}, we progressively expand the robot training corpus from 2,005 to 14,618 hours (2,005 / 4,261 / 8,475 / 14,618 hours), drawn from eight real-world robot domains (\cref{ssec:robot_data}). In the \textit{hybrid series}, we hold the robot corpus fixed at 14,618 hours and incorporate either 15,901 or 45,938 hours of egocentric human data drawn from the curated corpus in \cref{sec:sources}, yielding total training budgets of approximately 30.5k and 60.6k hours, respectively. This design isolates the benefits of scaling robot demonstrations from the additional contribution of human experience at a fixed robot-data budget. Both modalities are represented in the unified action space described in \cref{ssec:extended_action}.

\paragraph{Goal images.}
During co-training, goal images are sampled online from future observations within the same trajectory using an identical strategy for human and robot data. For each training example, we sample a clean head-camera frame from four goal-conditioning options: a future observation 2--8\,s ahead (40\%), the episode's terminal observation (20\%), the final observation of an upcoming annotated subtask (20\%), or an empty goal (20\%). Goal timestamps are constrained to lie at or beyond the end of the supervised action chunk. Each goal image is encoded with a learnable goal embedding and serves exclusively as a conditioning signal without an associated reconstruction loss. When the goal is omitted, its tokens are removed entirely, enabling a single model to support both goal-conditioned and language-only execution.

\paragraph{Optimization.}
All configurations start from the same pretrained Cosmos3-Nano-16B initialization. We optimize the Generator, timestep embeddings, visual and action projections, action modality embedding, and goal embedding, while keeping the Reasoner and VAE frozen. Training uses FusedAdam with a base learning rate of $4\times10^{-5}$, $(\beta_1,\beta_2)=(0.9,0.99)$, weight decay of $0.05$, gradient-norm clipping at $1.0$, and a global batch size of 1,024. The learning rate increases linearly over the first 1,000 steps and subsequently decays linearly over a 500k-step schedule. We apply a $5\times$ learning-rate multiplier to the action projections, action modality embedding, and goal embedding. All evaluations use exponential moving average (EMA) weights.

\paragraph{RoboTwin post-training.}
For simulation evaluation, we initialize from the co-trained checkpoint and adapt the model using 50 clean demonstrations from the ALOHA-AgileX robot per RoboTwin task, totaling 2,500 trajectories. \Cref{tab:rt_eval_settings} summarizes the adaptation protocols used in our post-training scaling and leaderboard evaluations (\cref{sssec:rt_posttrain}). Action normalization statistics are computed jointly from the clean RoboTwin datasets and pretraining data.

For goal specification during post-training, we use a mixture of ground-truth terminal frames and synthesized goal images. Specifically, 20\% of the training episodes use the ground-truth terminal observation as the goal, while the remaining 80\% use goal images synthesized by GPT Image 2.5~\cite{openai2026image25}. Each goal image is generated from the initial scene observation and task instruction using a fixed task-specific prompt (Appendix~\ref{app:goal_prompt}). To ensure fidelity to the intended task, every generated image undergoes an automated review using ChatGPT Astra High~\cite{openai2026astrahigh}, which verifies the target object and destination, designated manipulation arm, object count and appearance, and background and viewpoint consistency. Images that fail any check are regenerated and reviewed until accepted; this procedure enforces visual and instructional consistency rather than physical feasibility. At evaluation time, each scene is assigned a single generated goal image shared across all evaluated checkpoints to eliminate goal variation as a source of performance differences.

\subsection{Evaluation Settings}
\label{ssec:eval_protocol}

\paragraph{Benchmark.}
To ensure reproducibility and fair comparison with existing robot policies, we adopt RoboTwin 2.0~\cite{robotwin2}, a standardized simulation benchmark for bimanual manipulation. Built on the SAPIEN physics engine, RoboTwin 2.0 provides 50 manipulation tasks with predefined success criteria and controlled environmental variations. We evaluate all models using the ALOHA-AgileX embodiment, which is never exposed during pretraining. During each episode, the policy receives visual observations, robot states, and task instructions, executing actions in closed loop. Task success is determined by the benchmark's automated, task-specific completion checks.

RoboTwin 2.0 defines two evaluation conditions: \textit{Clean}, which uses the default scene configuration, and \textit{Random}, which introduces variations in backgrounds, lighting, object clutter, and table height. Together, they assess task execution under standard conditions and robustness to environmental shifts across the same 50 tasks and identical success criteria.

\paragraph{Evaluation protocols.}
We consider two complementary protocols to disentangle the capabilities acquired during co-training from those gained through target-task adaptation (\cref{tab:rt_eval_settings}). In the \textit{zero-shot} protocol, co-trained checkpoints are deployed directly on RoboTwin 2.0 without any target-task demonstrations or parameter updates, testing whether scaling hybrid experience enables unseen manipulation tasks to be executed out of the box. In the \textit{post-training} protocol, each co-trained checkpoint is adapted using 50 Clean demonstrations per task to evaluate how pretraining scale affects downstream in-domain fit and out-of-distribution generalization. For computational efficiency, zero-shot evaluations use 50 episodes per task under each condition (Clean and Random), whereas post-training scaling and leaderboard evaluations follow the official protocol of 100 episodes per task and condition (10,000 episodes in total).

\begin{table}[!htbp]
\centering
\caption{\textbf{RoboTwin 2.0 evaluation protocols.}}
\label{tab:rt_eval_settings}
\small
\begin{tabular}{@{}lp{0.39\linewidth}cl@{}}
\toprule
Setting & Target-task adaptation & Episodes & Results \\
\midrule
Zero-shot & None & 50 & \Cref{tab:robotwin_zero_shot} \\
Post-training & 50 clean demos/task & 100 & \Cref{fig:robotwin_data_scaling} \\
Leaderboard & Best configuration + augmentation & 100 & \Cref{tab:robotwin_leaderboard} \\
\bottomrule
\end{tabular}
\end{table}

\paragraph{Metrics.}
We report task success rates averaged equally across all 50 tasks for Clean and Random, along with their arithmetic mean. To characterize the breadth of zero-shot execution beyond aggregate success rates, we also report \textit{task coverage}, defined as the number of tasks achieving at least 2\% mean success across Clean and Random (corresponding to at least two successful rollouts out of 100 episodes). Alongside individual checkpoint coverage, we track \textit{cumulative coverage} across the scaling series, defined as the union of feasible tasks across all checkpoints up to a given data budget.

\subsection{Comparison with State-of-the-Art Embodied Foundation Models}
\label{ssec:goalwam_results}

We evaluate HumanWAM-Goal on RoboTwin 2.0 across both protocols defined in \cref{ssec:eval_protocol}. We first examine zero-shot execution against representative embodied foundation models and across hybrid data scales (\cref{sssec:zs_perf}), and then assess post-trained performance and official leaderboard comparisons (\cref{sssec:rt_posttrain}).

\subsubsection{Zero-Shot Performance}
\label{sssec:zs_perf}

\begin{table*}[!htbp]
  \centering
  \caption{\textbf{Zero-shot evaluation on RoboTwin 2.0 (ALOHA-AgileX).}
  Pretrained checkpoints are evaluated directly without RoboTwin-specific adaptation using 50 episodes per task under both Clean and Random conditions. Success rates are reported in percent. Status codes \texttt{ERR0}--\texttt{ERR2} denote deployment barriers that prevent zero-shot evaluation.}
  \label{tab:robotwin_zero_shot}
  \small
  \setlength{\tabcolsep}{5pt}
  \renewcommand{\arraystretch}{1.2}

  \begin{tabularx}{\textwidth}{@{}
    >{\raggedright\arraybackslash}X
    *{3}{>{\centering\arraybackslash}p{0.18\textwidth}}@{}}
    \toprule
    \textbf{Model} & \textbf{Clean} & \textbf{Random} & \textbf{Average} \\
    \midrule
    LingBot-VLA 2.0
      & \texttt{ERR0} & \texttt{ERR0} & \texttt{ERR0} \\
    LingBot-VA
      & \texttt{ERR0} & \texttt{ERR0} & \texttt{ERR0} \\
    OpenWAM~\cite{openwam}
      & \texttt{ERR0} & \texttt{ERR0} & \texttt{ERR0} \\
    \addlinespace[3pt]
    DM0.5
      & \texttt{ERR1} & \texttt{ERR1} & \texttt{ERR1} \\
    GR00T N1.7
      & \texttt{ERR1} & \texttt{ERR1} & \texttt{ERR1} \\
    DreamZero-AgiBot~\cite{ye2026worldactionmodelszeroshot}
      & \texttt{ERR1} & \texttt{ERR1} & \texttt{ERR1} \\
    \addlinespace[3pt]
    LingBot-VA 2.0
      & \texttt{ERR2} & \texttt{ERR2} & \texttt{ERR2} \\
    Fast-WAM~\cite{yuan2026fastwamworldactionmodels}
      & \texttt{ERR2} & \texttt{ERR2} & \texttt{ERR2} \\
    MV-WAM
      & \texttt{ERR2} & \texttt{ERR2} & \texttt{ERR2} \\
    \midrule
    InternVLA-A1.5~\cite{ma2026internvla}
      & 0.00 & 0.00 & 0.00 \\
    InternW0-$\Delta$~\cite{internw0delta}
      & 0.00 & 0.00 & 0.00 \\
    Zero-WAM~\cite{zhou2026zero}
      & 0.00 & 0.00 & 0.00 \\
    Hy-VLA-UMI~\cite{zhang2026hy}
      & 0.00 & 0.00 & 0.00 \\
    \midrule
    $\pi_0$~\cite{pi0}
      & 0.16 & 0.08 & 0.12 \\
    $\pi_{0.5}$~\cite{intelligence2025pi_}
      & 0.88 & 1.24 & 1.06 \\
    Pelican-VLA 0.5~\cite{ding2026pelican}
      & 5.36 & 4.04 & 4.70 \\
    \midrule
    HumanWAM-Goal (Robot Only)
      & 11.40 & 11.56 & 11.48 \\
    \textbf{HumanWAM-Goal (Human-Robot Hybrid)}
      & \textbf{13.68} & \textbf{14.20} & \textbf{13.94} \\
    \bottomrule
  \end{tabularx}

  \par\vspace{5pt}
  \begin{minipage}{\textwidth}
    \footnotesize
    \raggedright
    \textit{Notes.}
    Average is the arithmetic mean of Clean and Random. A value of $0.00$ indicates that rollouts were executed but yielded no successful episodes.
    \par\smallskip
    \texttt{ERR0}: Evaluation could not be completed because the released checkpoints lacked the required action normalization statistics.
    \par\smallskip
    \texttt{ERR1}: Evaluation was not conducted because a compatible control mapping to the target embodiment was unavailable.
    \par\smallskip
    \texttt{ERR2}: No publicly available pretrained checkpoint suitable for this evaluation protocol was available.
  \end{minipage}
\end{table*}

\paragraph{Zero-shot comparison with existing foundation models.}
\Cref{tab:robotwin_zero_shot} compares HumanWAM-Goal with representative open-source VLA and WAM baselines under the zero-shot protocol. Most existing models are designed exclusively for a pretrain-then-finetune workflow and cannot be deployed out of the box due to unreleased action normalization statistics (\texttt{ERR0}), incompatible embodiment-specific control spaces (\texttt{ERR1}), or the absence of public pretrained weights without fine-tuning on RoboTwin (\texttt{ERR2}). Among the models that can be executed directly, several exhibit static or non-task-directed motions ($0.00\%$ success), while $\pi_0$, $\pi_{0.5}$, and Pelican-VLA~0.5 achieve average success rates of $0.12\%$, $1.06\%$, and $4.70\%$, respectively. By contrast, because HumanWAM-Goal aligns camera-frame EEF displacements and metric gripper apertures on a unified physical scale across pretraining sources while masking embodiment-specific joints (\cref{ssec:extended_action}), its global EEF normalization statistics represent embodiment-agnostic physical motions and apply directly to unseen robots via standard inverse kinematics without target-domain recalibration. The robot-only HumanWAM-Goal checkpoint reaches $11.48\%$ average success, and human--robot hybrid co-training further elevates performance to $13.94\%$. These results indicate that combining a unified physical action interface with explicit visual-goal conditioning enables transferable zero-shot execution directly from pretraining.

\begin{figure}[!htbp]
  \centering
  \begin{subfigure}[t]{0.48\linewidth}
    \centering
    \includegraphics[width=\linewidth]{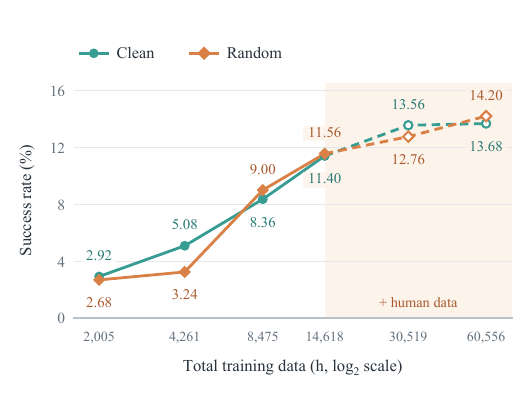}
    \caption{Success rate.}
    \label{fig:wam_success}
  \end{subfigure}
  \hfill
  \begin{subfigure}[t]{0.48\linewidth}
    \centering
    \includegraphics[width=\linewidth]{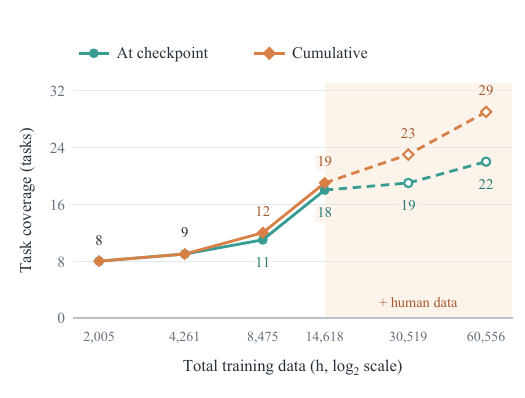}
    \caption{Task coverage.}
    \label{fig:wam_coverage}
  \end{subfigure}
  \caption{\textbf{Zero-shot performance under hybrid data scaling.}
  Task success rates and task coverage ($\ge 2\%$ mean success) across six HumanWAM-Goal pretraining configurations on RoboTwin 2.0 (50 Clean and 50 Random episodes per task).}
  \label{fig:human_robot_scaling}
\end{figure}

\paragraph{Scaling robot and hybrid data.}
\Cref{fig:human_robot_scaling} tracks zero-shot success rates and task coverage across six independently trained configurations. As real-robot training data increase from 2,005 to 14,618 hours ($7.3\times$), average success rises steadily from $2.80\%$ to $11.48\%$, while single-checkpoint task coverage expands from 8 to 18 of the 50 tasks. Both metrics improve at every intermediate budget, indicating that scaling robot experience broadens the set of solvable tasks rather than merely improving reliability on a fixed subset. Holding robot data fixed at 14,618 hours, incorporating 15,901 hours of egocentric human video further increases average success to $13.16\%$ (19 covered tasks), and scaling to 45,938 hours of human video reaches $13.94\%$ (22 covered tasks, and 29 tasks in cumulative coverage), outperforming the robot-only baseline by $2.46$~pp. Notably, expanding the human corpus from 15,901 to 45,938 hours improves Random success by $1.44$~pp compared with $0.12$~pp on Clean, suggesting that human interaction diversity primarily strengthens zero-shot robustness under visual and environmental variation.

\begin{figure}[!th]
  \centering
  \includegraphics[width=\linewidth]{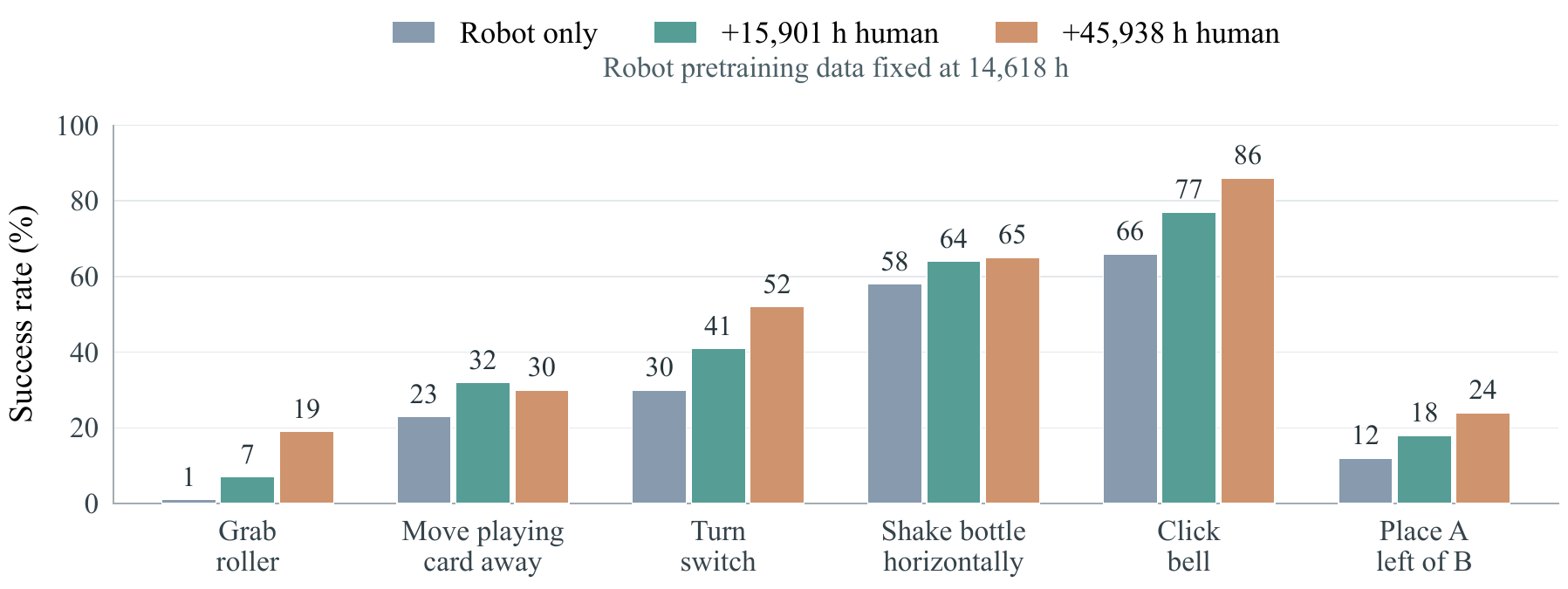}
  \caption{\textbf{Task-level zero-shot scaling with human-robot hybrid training.}
  Mean zero-shot success rates (averaged over Clean and Random) across six representative RoboTwin 2.0 tasks with robot data fixed at 14,618 hours and increasing egocentric human data.}
  \label{fig:robotwin_task_cotraining}
\end{figure}

\paragraph{Task-level scaling analysis.}
\Cref{fig:robotwin_task_cotraining} illustrates how adding human data affects individual manipulation skills across six representative tasks spanning grasping, contact, and spatial placement. We observe two distinct patterns of improvement. First, for tasks where robot-only pretraining struggles or rarely succeeds---such as \textit{Grab roller} ($1\% \rightarrow 19\%$), \textit{Move playing card away} ($23\% \rightarrow 30\%$), \textit{Turn switch} ($30\% \rightarrow 52\%$), and \textit{Place A left of B} ($12\% \rightarrow 24\%$)---hybrid co-training unlocks or substantially strengthens task feasibility. Second, for skills where the robot-only model already establishes a solid baseline, such as \textit{Shake bottle horizontally} ($58\% \rightarrow 65\%$) and \textit{Click bell} ($66\% \rightarrow 86\%$), human experience further refines execution reliability. While the marginal gain between the two human-data scales varies by task, hybrid co-training consistently improves upon robot-only pretraining across both emerging and established skills.

\subsubsection{Post-Trained Performance}
\label{sssec:rt_posttrain}

We next evaluate whether the advantages of hybrid co-training persist when target-task demonstrations are available for adaptation.

\begin{figure}[!htbp]
  \centering
  \includegraphics[width=\linewidth]{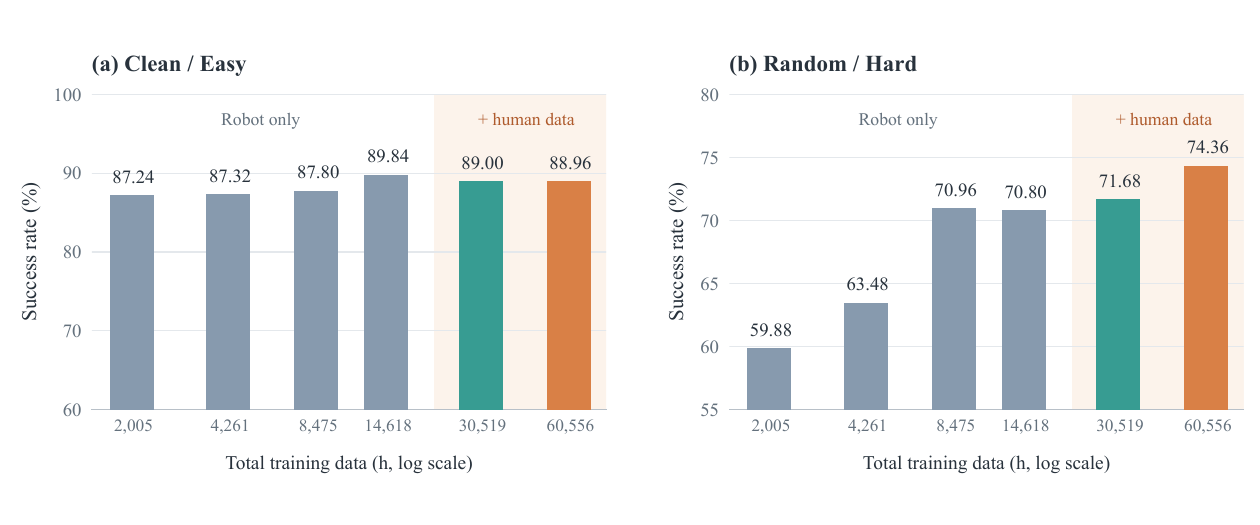}
  \caption{\textbf{Post-trained performance across pretraining scales on RoboTwin 2.0.}}
  \label{fig:robotwin_data_scaling}
\end{figure}

\paragraph{Post-training performance across pretraining scales.}
\Cref{fig:robotwin_data_scaling} reports downstream success rates after adapting each pretrained checkpoint on 50 Clean demonstrations per task. Scaling robot-only pretraining from 2,005 to 14,618 hours increases Clean success from $87.24\%$ to $89.84\%$ and Random success from $59.88\%$ to $70.80\%$. However, Random performance plateaus beyond 8,475 robot hours ($70.96\%$ at 8,475 hours vs.\ $70.80\%$ at 14,618 hours), reflecting the limited scene diversity of teleoperated robot trajectories. By contrast, augmenting the 14,618-hour robot corpus with 15,901 and 45,938 hours of egocentric human data breaks this plateau, raising Random success to $71.68\%$ and $74.36\%$ ($+3.56$~pp over robot-only), respectively, while Clean success rate remains saturated near $89\%$. Consistent with observations in Qwen-RobotManip~\cite{yuan2026qwen}, in-distribution evaluation (Clean) is largely saturated by target-task demonstrations, whereas out-of-distribution evaluation (Random) reveals the transferable visuomotor priors gained from large-scale human--robot co-training.

\begin{table*}[t]
  \centering
  \caption{\textbf{Evaluation results on RoboTwin 2.0.}
  Success rates (\%) are reported after clean-only post-training.
  Average is the mean of Clean2Clean and Clean2Random.
  The best result in each column is highlighted in bold.}
  \label{tab:robotwin_leaderboard}
  \small
  \setlength{\tabcolsep}{7pt}
  \renewcommand{\arraystretch}{1.12}
  \begin{tabular*}{\textwidth}
    {@{}l@{\hspace{1.5em}}l@{\extracolsep{\fill}}rrr@{}}
    \toprule
    Type & Method
         & Clean2Clean $\uparrow$
         & Clean2Random $\uparrow$
         & Average $\uparrow$ \\
    \midrule
    \multirow{7}{*}{\color{black!65}\sffamily VLA}
      & Qwen-RobotManip-Context~\cite{yuan2026qwen} & 84.7 & 69.4 & 77.1 \\
      & GigaBrain-0.7~\cite{gigabrain07}           & 66.80 & 67.92 & 67.36 \\
      & $\pi_{0.5}$~\cite{intelligence2025pi_}             & 70.70 & 46.00 & 58.35 \\
      & Spatial Forcing~\cite{spatialforcing}         & 77.20 & 26.74 & 51.97 \\
      & X-VLA~\cite{xvla}                   & 68.00 & 20.90 & 44.45 \\
      & Abot-M0~\cite{abotm0}                 & 57.40 & 30.40 & 43.90 \\
      & GR00T-N1.7~\cite{gr00tn17}              & 43.6 & 20.7 & 32.2 \\
    \midrule
    \multirow{13}{*}{\color{black!65}\sffamily WAM}
      & GO-3 Nano Preview       & 88.54 & 81.48 & 85.01 \\
      & PICA-WAM~\cite{picawam}                & 90.68 & 74.70 & 82.69 \\
      & CloudRobo-WM            & 88.90 & 75.00 & 81.95 \\
      & MWA-Phoenix~\cite{phoenixmwa}             & 86.54 & 72.42 & 79.48 \\
      & PatchWAM~\cite{patchwam}                & 91.56 & 66.72 & 79.14 \\
      & ME-Dex-1.0~\cite{medex}              & 89.58 & 68.12 & 78.85 \\
      & EWAM~\cite{ewam}                    & 82.16 & 72.08 & 77.12 \\
      & ViGAR~\cite{gong2026rethinking}                   & 82.00 & 67.02 & 74.51 \\
      & UniWAM~\cite{uniwam}                  & 75.12 & 67.56 & 71.34 \\
      & OpenWAM-$\alpha$~\cite{openwam}        & 89.4 & 48.7 & 69.0 \\
      & 4D-WAM~\cite{wam4d}                  & 81.50 & 41.80 & 61.65 \\
    \cmidrule(l){2-5}
      & HumanWAM-Vanilla (Ours)
                                & 92.08 & 68.88 & 80.48 \\
      & \textbf{HumanWAM-Goal (Ours)}
                                & \textbf{92.70}
                                & \textbf{84.60}
                                & \textbf{88.65} \\
    \bottomrule
  \end{tabular*}
  \par\vspace{4pt}
  \begin{minipage}{\textwidth}
    \footnotesize
    Baseline results are taken from the
    \href{https://robotwin-platform.github.io/leaderboard}
    {official RoboTwin~2.0 leaderboard}~\cite{robotwin_leaderboard}
    (Co-train track, without observed depth at inference),
    except for GR00T-N1.7, Qwen-RobotManip-Context,
    and OpenWAM-$\alpha$, whose published results are
    sourced from InternW0-$\Delta$~\cite{internw0delta}.
    HumanWAM-Vanilla and HumanWAM-Goal results are
    obtained from our own evaluations.
  \end{minipage}
\end{table*}

\paragraph{Leaderboard comparison with refined post-training.}
To evaluate the full potential of our co-trained initialization, we combine our best hybrid checkpoint with photometric data augmentation and refinements to the post-training recipe. \Cref{tab:robotwin_leaderboard} shows several representative entries on the RoboTwin 2.0 leaderboard~\cite{robotwin2}. Under the official 10,000-episode evaluation protocol, HumanWAM-Goal achieves $92.70\%$ on Clean and $84.60\%$ on Random ($88.65\%$ Average), outperforming the previous leading entry (GO-3 Nano Preview) by $4.16$~pp on Clean, $3.12$~pp on Random, and $3.64$~pp on average. Because this configuration incorporates post-training augmentations beyond the controlled scaling recipe, we report it separately from the scaling curves in \cref{fig:robotwin_data_scaling}.
\section{Emerging Capabilities in Real-World Robot Behavior}
\label{sec:real_world_icl}

While recent multimodal agents exhibit strong high-level reasoning and task planning \cite{si2026fewer,liu2026roboicl}, translating semantic intent into physical work requires a general-purpose action executor that can reliably control real robots. Building on the simulation scaling trends in \cref{sec:human_robot_scaling}—which show that joint human--robot co-training with visual-goal conditioning instills transferable visuomotor priors—we transition to physical hardware to evaluate whether HumanWAM-Goal can bridge this gap. Central to our deployment is explicit goal-image conditioning, which decouples high-level task specification from low-level sensorimotor control. Rather than relying solely on language instructions that often leave spatial arrangements and terminal states underspecified, an upstream planner or user supplies a visual goal as inference-time context. This unified interface allows the model to serve both as an ``out-of-the-box'' executor for unseen tasks and as a scalable foundation for downstream post-training.

Following our experimental setup (\cref{ssec:real_world_setup}), we evaluate the real-world behaviors of HumanWAM-Goal across three core capabilities. First, we investigate \textbf{zero-shot real-world manipulation} (\cref{ssec:real_world_zero_shot}) by deploying co-trained checkpoints directly across a 20-task benchmark, quantifying how hybrid human--robot scaling expands the set of skills executable without target-task demonstrations or parameter updates. Second, we examine \textbf{embodied in-context learning} (\cref{ssec:goal_based_icl}), testing whether a policy with strictly frozen weights—both before and after post-training—can dynamically alter its manipulation trajectory in response to varying robot- or human-depicted visual goals. Finally, we demonstrate \textbf{long-horizon execution} (\cref{ssec:real_world_long_horizon}) on post-trained policies, showcasing temporal continuity and subgoal chaining across extended, multi-stage tabletop and mobile-manipulation workflows.  

\subsection{Real-Robot Experimental Setup}
\label{ssec:real_world_setup}

\paragraph{Hardware platform and deployment protocol.}
We conduct physical-robot evaluations on \textbf{Astribot S1}, a wheeled mobile manipulator with an articulated torso. Across all experiments, policies receive RGB observations from a head-mounted camera and two wrist cameras alongside proprioceptive robot states, predicting a 48-step ($1.6\,\text{s}$) action chunk in the unified 49-D hybrid action space (\cref{ssec:extended_action}). Desired task outcomes are specified at inference time via the goal-image interface (\cref{ssec:goal_wam}), using target scene configurations captured from held-out robot demonstrations or egocentric human demonstrations (\cref{ssec:goal_based_icl}). To ensure low-latency closed-loop control on physical hardware, we deploy our models using PhyAI~\cite{phyai2026}, which leverages tensor parallelism and highly optimized kernels to achieve a $2.2\times$ inference speedup over a PyTorch baseline with CUDA Graphs.

\paragraph{Zero-shot task suite and evaluation metrics.}
Our zero-shot benchmark comprises 20 manipulation tasks stratified into 10 easy and 10 hard tasks. The suite spans five core manipulation families: contact and state changes (4 tasks), grasping and transport (5), articulated opening and closing (3), spatial placement and alignment (6), and stacking and insertion (2). We conduct 10 trials per task, totaling 200 rollouts per evaluated checkpoint.

We report both binary \textbf{task success rate} and a fine-grained \textbf{progress score} that captures partial task completion. Specifically, each task $k$ is decomposed into $M_k$ ordered, non-overlapping stages with fixed completion criteria judged manually from visual observations and end-effector poses. The initial stage always assesses target identification and approach—decoupling perception and reaching errors from subsequent manipulation failures—while the final stage encodes the terminal success predicate. Each stage $s \in \{1, \dots, M_k\}$ is assigned a predefined weight $w_{k,s} \in \{1,2,3,4\}$ such that $\sum_{s=1}^{M_k} w_{k,s}=10$. The progress score for trial $r$ is defined as
\begin{equation}
    P_{k,r} = \sum_{s=1}^{M_k} w_{k,s} b_{k,r,s}, \qquad P_{k,r} \in [0,10],
\end{equation}
where $b_{k,r,s} \in \{0, 1\}$ indicates whether stage $s$ is successfully completed. Both binary success and progress scores are averaged over the 10 trials per task and then macro-averaged equally across all 20 tasks. Appendix~\ref{app:real_world_tasks} lists the tasks, stage scores, and success outcomes.

\paragraph{Post-training protocol.}
For intricate, multi-stage, or long-horizon tasks that exceed zero-shot reliability, we adapt the co-trained checkpoint via task-specific post-training, resetting the optimizer, learning-rate schedule, and data sampler. We retain the co-training normalization statistics and recalculate only channels with substantially shifted distributions (e.g., the articulated torso of Astribot S1). During post-training, goal images are sampled online from the final frame of one of the next two annotated subtasks, a future frame within $4\,\text{s}$, or an empty goal. To improve robustness and enforce reliance on visual goals and shared EEF representations, we apply $20\%$ dropout to the joint-state input branch, $10\%$ dropout to language instructions, and weight the action loss $10\times$ relative to the video prediction loss. We select the EMA checkpoint that minimizes end-effector average displacement error (ADE) on held-out validation episodes.

\subsection{Real-World Zero-Shot Performance}
\label{ssec:real_world_zero_shot}

We evaluate the zero-shot transfer of HumanWAM-Goal on physical hardware by comparing the robot-only baseline ($14,618\,\text{h}$) against hybrid checkpoints co-trained with an additional $15,901$ or $45,938$ hours of egocentric human data (yielding total budgets of $30,519\,\text{h}$ and $60,556\,\text{h}$, respectively). All checkpoints are deployed directly out of the box without target-task demonstrations or parameter updates.

\begin{figure}[!htbp]
    \centering
    \includegraphics[width=0.85\linewidth]{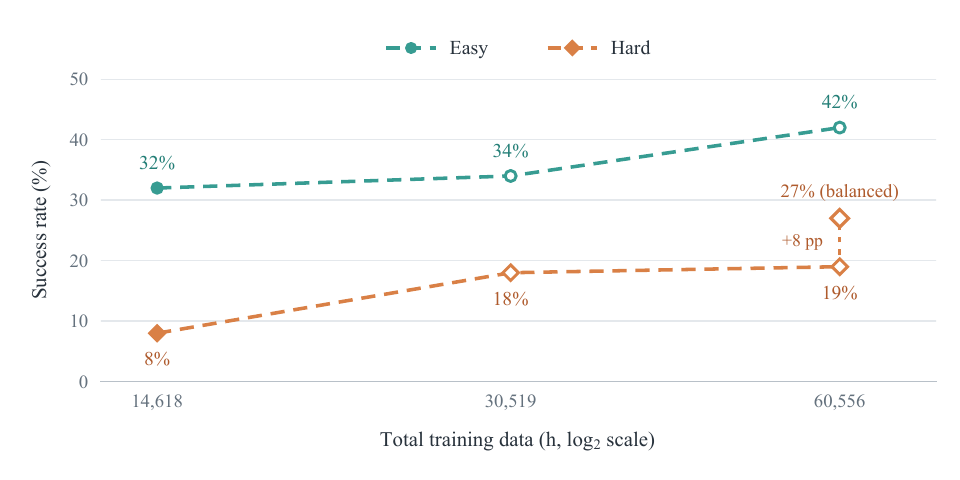}
    \caption{\textbf{Zero-shot real-robot performance stratified by task difficulty.}
    Success rates across easy and hard tasks as egocentric human data are added to a fixed $14,618\,\text{h}$ robot corpus. The vertical dashed line at $60,556\,\text{h}$ highlights the additional gain on hard tasks from robot-balanced batch sampling ($+8\,\text{pp}$).}
    \label{fig:real_robot_zero_shot_scaling}
\end{figure}

\paragraph{Scaling trends across task difficulty.}
\Cref{fig:real_robot_zero_shot_scaling} illustrates how scaling hybrid data affects real-world zero-shot execution across different task difficulties. As egocentric human data expand from $0$ to $15.9\text{k}$ and $45.9\text{k}$ hours, success on easy tasks rises steadily from $32.0\%$ to $34.0\%$ and $42.0\%$ ($+10.0\,\text{pp}$ overall), while success on hard tasks more than doubles from $8.0\%$ to $18.0\%$ and $19.0\%$ ($+11.0\,\text{pp}$). These consistent gains confirm that visuomotor priors learned from human manipulation transfer directly to physical robot execution.

Notably, while easy tasks benefit most when scaling from $30.5\text{k}$ to $60.6\text{k}$ total hours, gains on hard tasks begin to plateau under standard data-proportional sampling, where human clips constitute roughly $76\%$ of the training mixture. To test whether this plateau stems from diluted embodiment-specific supervision, we evaluate a \textbf{robot-balanced} configuration at the $60.6\text{k}$-hour scale that enforces a $1{:}1$ ratio of robot and human samples in every training batch. Holding the data budget fixed, robot-balanced sampling boosts hard-task success from $19.0\%$ to $27.0\%$ ($+8.0\,\text{pp}$). This result indicates that while massive human data provide broad interaction priors, preserving sufficient batch exposure to robot trajectories is crucial for executing more demanding multi-step or precision-critical skills.

\paragraph{Task-level analysis and coverage expansion.}
\begin{figure}[!htbp]
    \centering
    \includegraphics[width=0.9\linewidth]{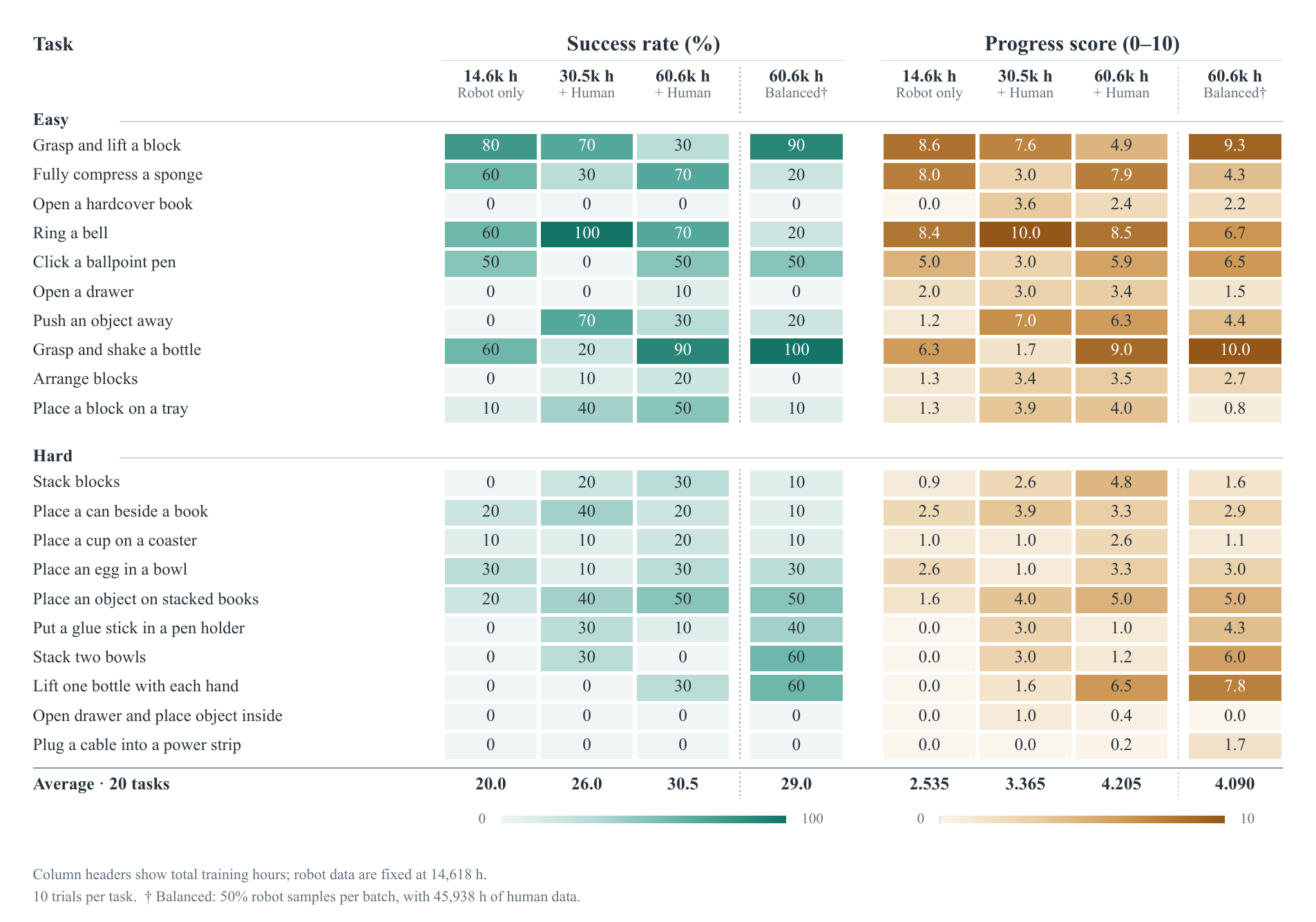}
    \caption{\textbf{Per-task zero-shot evaluation on physical robots.}
    Task-level success rates (left) and mean progress scores (right) across all 20 benchmark tasks for the robot-only baseline and three human--robot hybrid configurations.}
    \label{fig:real_robot_zero_shot}
\end{figure}

\Cref{fig:real_robot_zero_shot} breaks down zero-shot performance across all 20 benchmark tasks. Under the standard scaling series, overall mean success across the 20 tasks increases monotonically from $20.0\%$ (robot-only) to $26.0\%$ ($+15.9\text{k}\,\text{h}$ human) and $30.5\%$ ($+45.9\text{k}\,\text{h}$ human). Partial task completion improves in tandem, with the macro-averaged progress score rising from $2.535$ to $3.365$ and $4.205$ out of $10$. Equally important, hybrid scaling expands real-world \textbf{task coverage}: the number of tasks achieving at least one successful zero-shot rollout increases from $10/20$ with robot-only pretraining to $14/20$ and $16/20$ as human experience is incorporated. Echoing our simulation findings (\cref{sssec:zs_perf}), human data not only refine reliability on already feasible skills—such as \textit{Grasp and shake a bottle} ($60\% \rightarrow 90\%$) and \textit{Place a block on a tray} ($10\% \rightarrow 50\%$)—but also unlock entirely new capabilities where the robot-only baseline never succeeds ($0\%$), including \textit{Stack blocks} ($30\%$), \textit{Arrange blocks} ($20\%$), and \textit{Lift one bottle with each hand} ($30\%$).

Examining the per-task breakdown for the $60.6\text{k}$-hour robot-balanced model further clarifies its advantage on hard manipulation tasks. Compared with unweighted sampling at the same scale, balancing the human--robot ratio substantially improves tasks that require precise spatial alignment or bimanual coordination—such as \textit{Put a glue stick in a pen holder} ($10\% \rightarrow 40\%$), \textit{Stack two bowls} ($0\% \rightarrow 60\%$), and \textit{Lift one bottle with each hand} ($30\% \rightarrow 60\%$). By preventing the large human corpus from overwhelming embodiment-specific control dynamics, balanced co-training effectively combines human-scale visual-semantic diversity with high-precision robotic execution.

\begin{figure}[!htbp]
    \centering
    \includegraphics[width=0.9\linewidth]{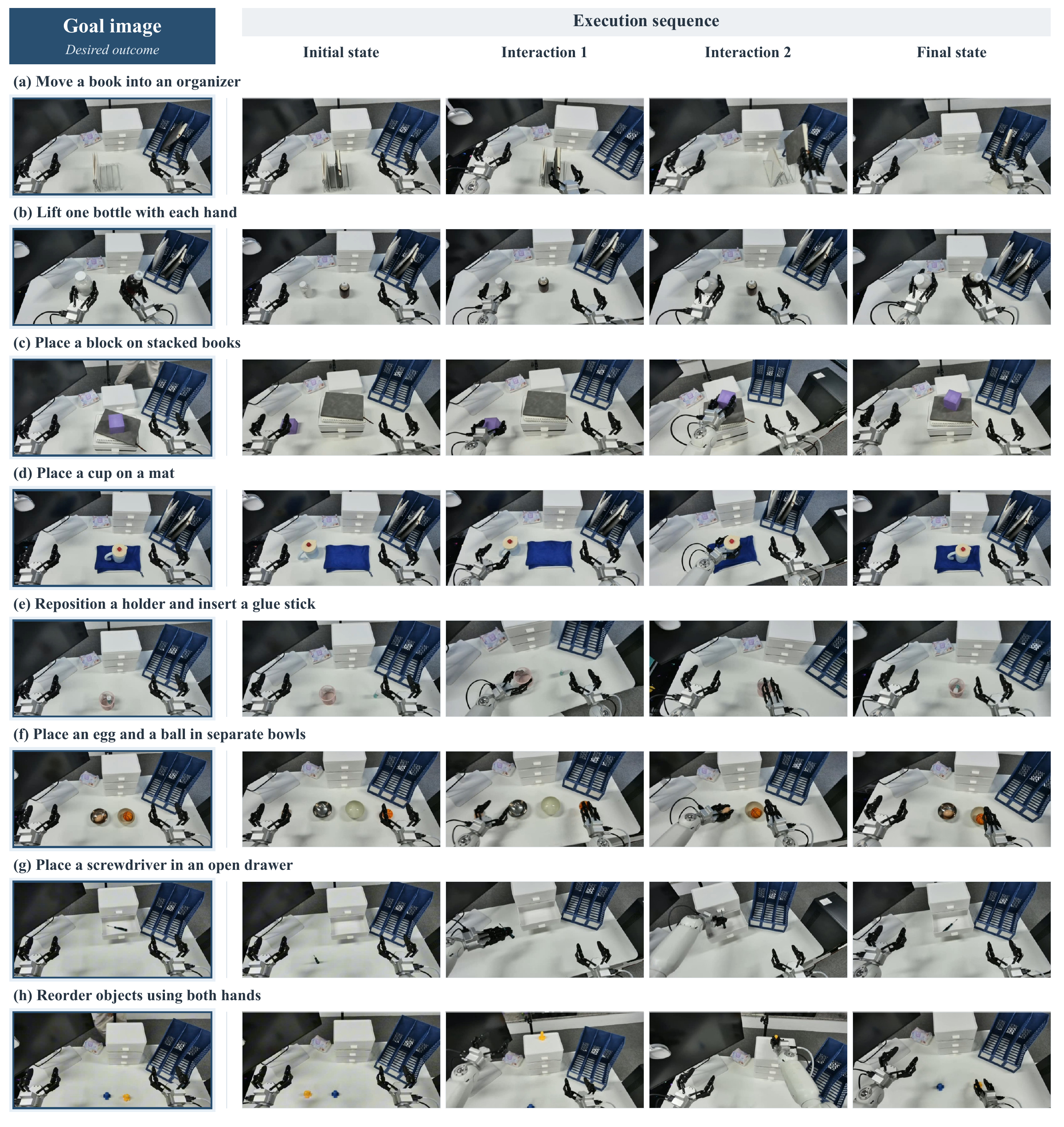}
    \caption{\textbf{Qualitative real-robot zero-shot rollouts.}
    Each row pairs the inference-time goal image (left) with four head-camera keyframes from the resulting zero-shot execution (right), showing the initial scene, two intermediate interaction stages, and the final state. The examples span object transport and placement, coordinated bimanual manipulation, object reordering, and multi-step tool/container interactions.}
    \label{fig:real_robot_zero_shot_qualitative}
\end{figure}

\paragraph{Qualitative analysis of zero-shot rollouts.}
\Cref{fig:real_robot_zero_shot_qualitative} visualizes representative zero-shot rollouts across eight real-world tasks. Beyond matching the terminal object arrangement depicted in the goal image, the policy exhibits purposeful intermediate behaviors that are not explicitly specified in the static visual goal—such as reorienting a pen holder prior to inserting a glue stick (row~e) and coordinating both arms to swap or distribute objects without collision (rows~b, f, h). These qualitative trajectories complement our quantitative metrics, demonstrating that HumanWAM-Goal translates static goal context into coherent, multi-step physical interactions out of the box.

\subsection{In-Context Learning with Visual Goals}
\label{ssec:goal_based_icl}

Whereas \cref{ssec:real_world_zero_shot} evaluates whether pretrained capabilities transfer directly to unseen tasks, we now examine a complementary question: can a single policy with \textbf{strictly frozen parameters} dynamically alter its execution trajectory when the desired outcome is modified purely through inference-time visual context? We evaluate this form of embodied in-context learning (ICL) by conditioning the same checkpoint on alternative goal images from identical initial scenes—varying target locations, category-to-container assignments, or stacking orders. We first verify that goal-responsive behavior emerges directly from co-training, then show that it scales to complex, multi-decision workflows after task-specific post-training, and finally demonstrate cross-embodiment ICL using egocentric human goal images.

\begin{figure}[!htbp]
    \centering
    \includegraphics[width=\linewidth]{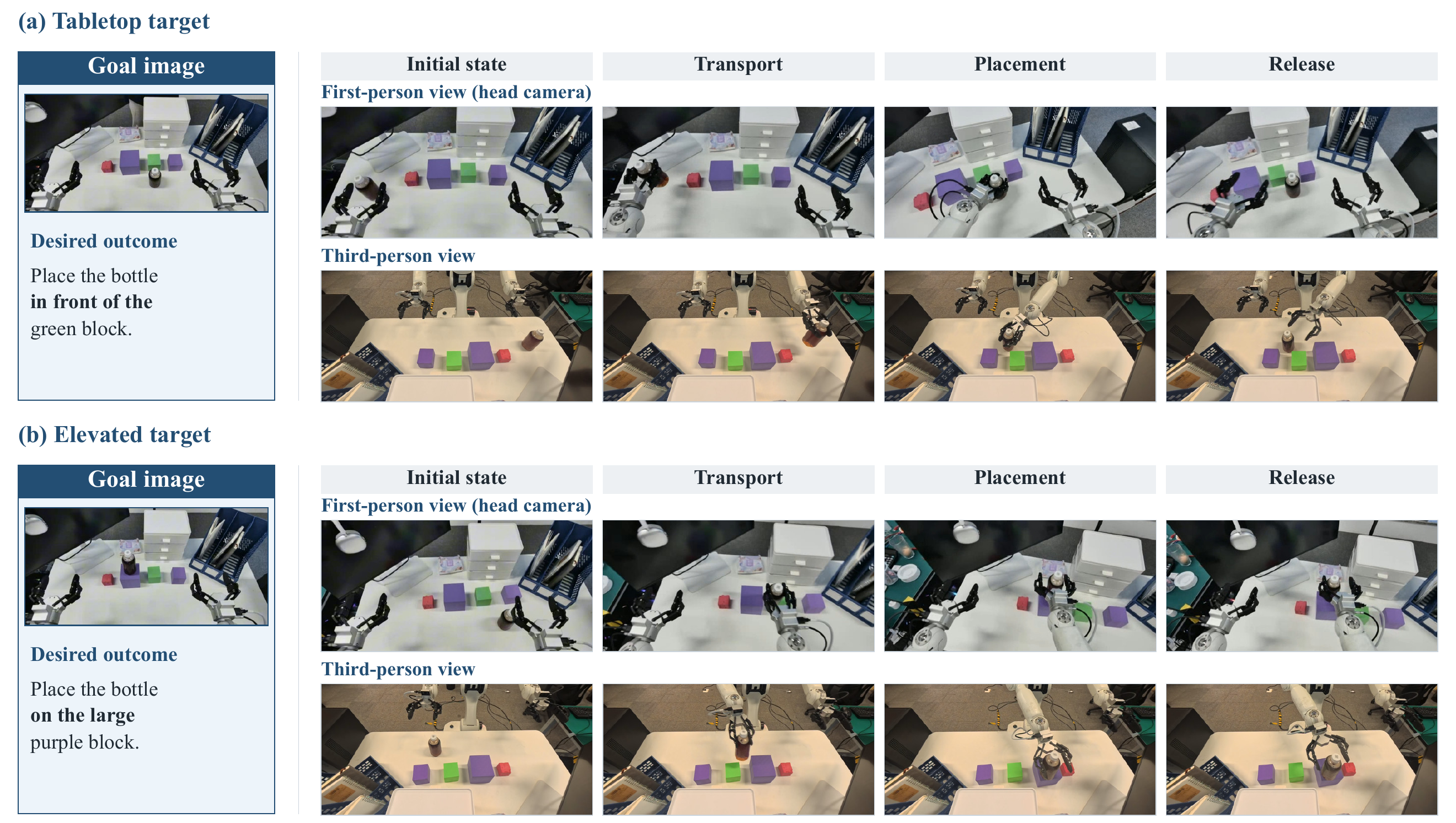}
    \caption{\textbf{Visual goal conditioning in pretrained HumanWAM-Goal.}
        Each panel pairs the inference-time goal image (left) with synchronized first-person and third-person keyframes (right). Without target-task post-training or parameter updates, changing the visual goal directs the policy to place the bottle (a) on the tabletop in front of the green block or (b) atop the large purple block.}
    \label{fig:pretrained_goal_positioning}
\end{figure}

\paragraph{Pretrained HumanWAM-Goal in-context learning behavior.}
We first probe visual goal conditioning in the co-trained checkpoint without any target-task post-training. \Cref{fig:pretrained_goal_positioning} compares two rollouts initialized from the same tabletop scene containing colored blocks, conditioned on visual goals that specify distinct destinations for a bottle. Keeping all policy weights fixed, switching the goal image steers the robot to place the bottle either on the tabletop in front of the green block (panel~a) or atop the elevated purple block (panel~b). While fine-grained release stability on the elevated block remains imperfect without task-specific adaptation, these rollouts confirm that spatial goal grounding is already present out of the box after hybrid pretraining.

\paragraph{In-context learning behavior after post-training.}
We next evaluate whether goal-conditioned steerability is preserved—and extended to multi-step, compositional decisions—after adapting the co-trained checkpoint on task-specific demonstrations. Crucially, once post-training concludes, policy parameters remain frozen across all evaluations; behavioral changes arise solely from the supplied goal image.

\begin{figure}[!htbp]
    \centering
    \includegraphics[width=\linewidth]{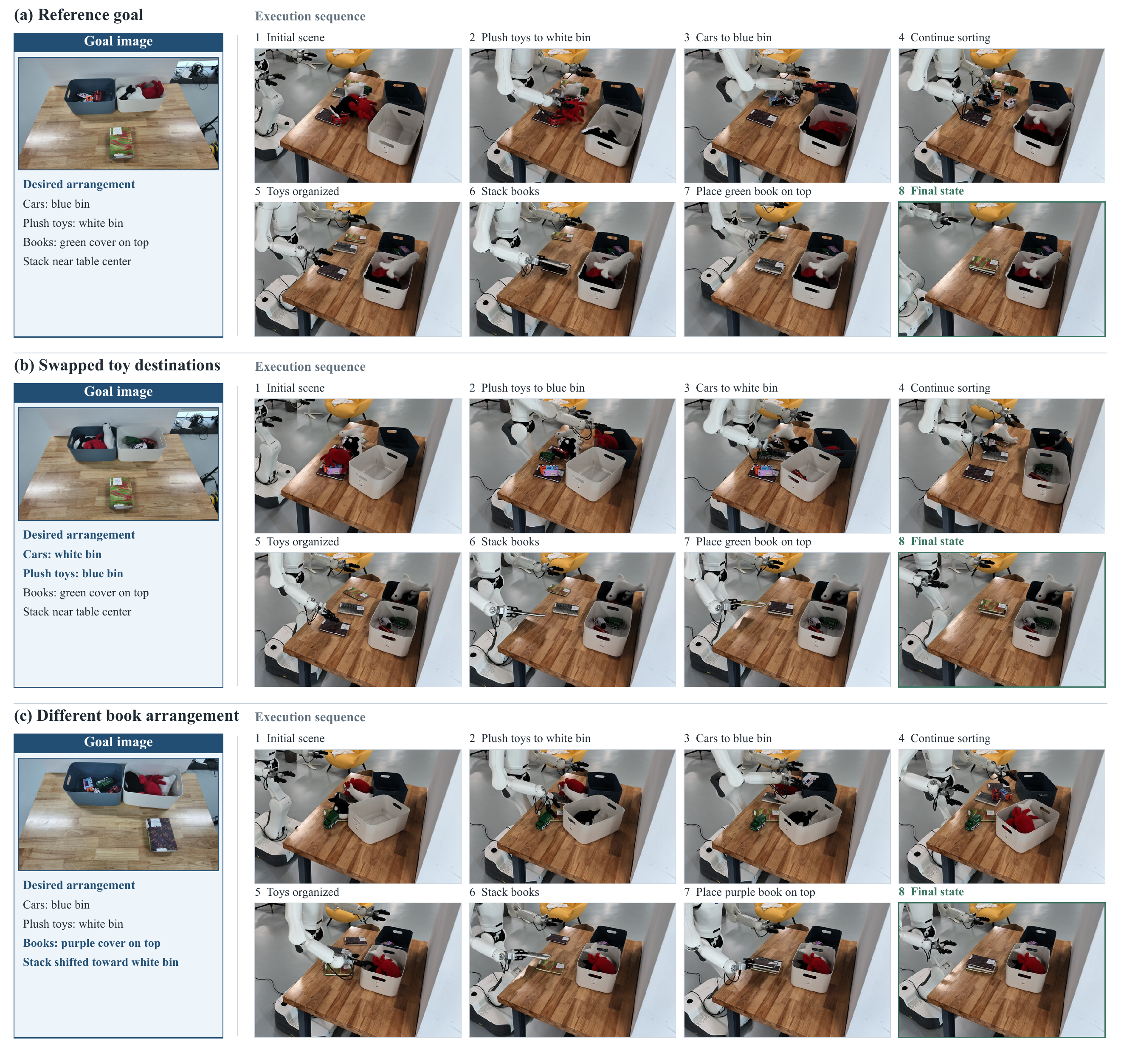}
    \caption{\textbf{Goal-based in-context learning after post-training for tabletop sorting and stacking.}
    Each panel pairs a supplied goal image (left, blue border) with eight chronological third-person frames (right, read left-to-right across the top then bottom row; green border marks the final state). The goals specify (a) sorting toy cars into the blue bin, plush toys into the white bin, and stacking books near the table center with the green book on top; (b) swapping the toy-to-bin assignments; and (c) retaining the original toy assignments while placing the purple book on top of a stack shifted toward the white bin.}
    \label{fig:real_robot_goal_icl}
\end{figure}

\Cref{fig:real_robot_goal_icl} evaluates a post-trained policy on a multi-object tabletop organization task under three distinct visual goals. Rather than memorizing a single canonical sorting routine, the frozen policy parses multiple compositional constraints from the goal image within each rollout: it reverses category-to-bin assignments when the destinations of toy cars and plush toys are swapped (panel~a vs.\ panel~b), and it modifies both the stacking order (green vs.\ purple book on top) and the spatial placement of the book stack when prompted by panel~c. Visual goal conditioning thus scales beyond single-target localization to coordinate multiple object-level decisions over extended manipulation horizons.

\begin{figure}[p]
    \centering
    \includegraphics[width=\linewidth,height=0.83\textheight,keepaspectratio]{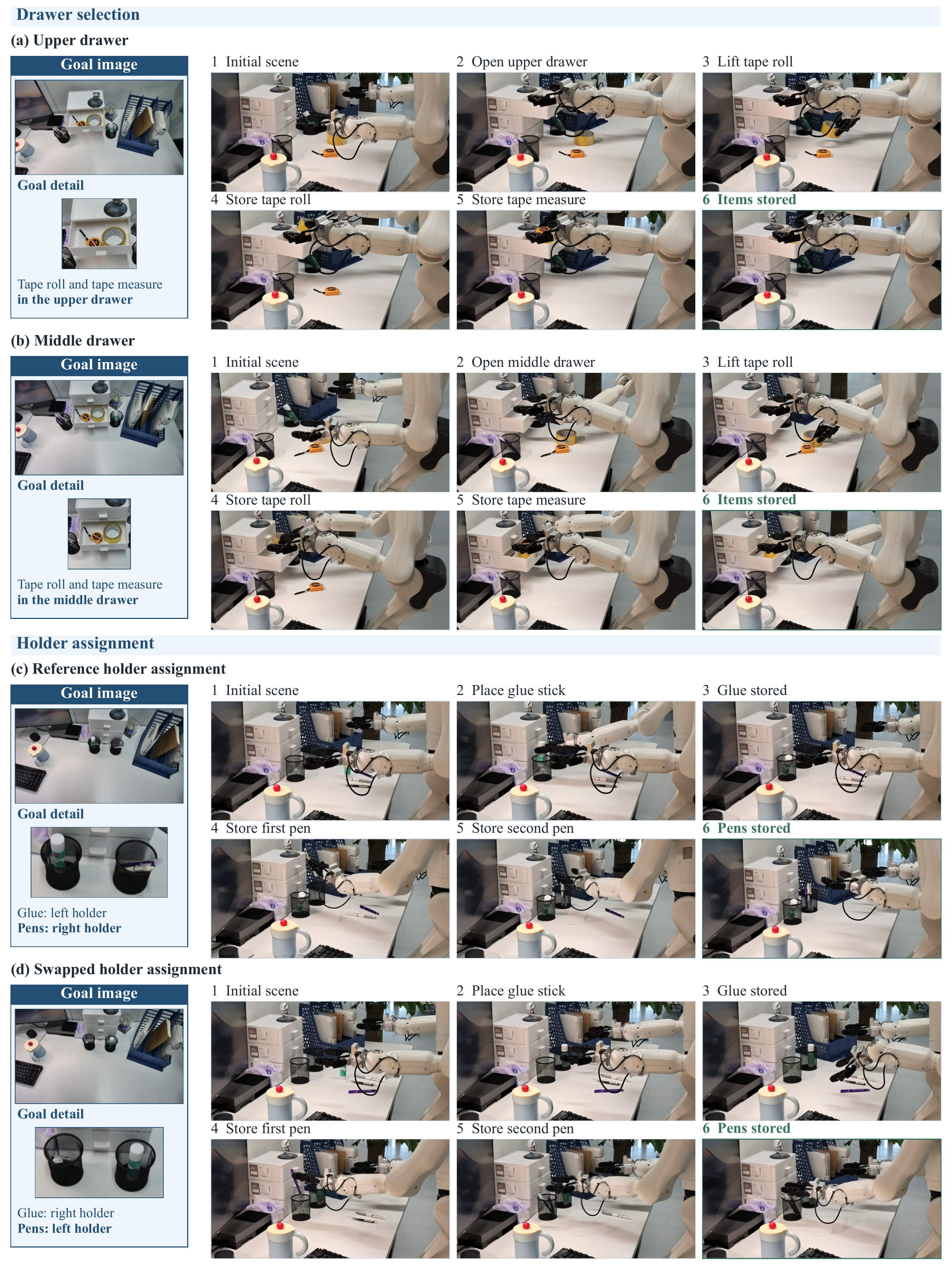}
    \caption{\textbf{Goal-based in-context learning after post-training for office desk organization.}
    Each panel pairs a goal image and enlarged detail view (left) with six ordered third-person execution frames (right). Panels (a, b) redirect the storage of a tape roll and tape measure from the upper drawer to the middle drawer. Panels (c, d) swap the destinations of a glue stick and two pens between the left and right holders (defined in the goal-image frame). Endpoints show item placement prior to drawer closure in (a, b) and after the final pen release in (c, d).}
    \label{fig:real_robot_desk_icl}
\end{figure}

\Cref{fig:real_robot_desk_icl} further tests paired goal variations in an office workspace involving articulated storage and multi-item sorting. In panels~(a, b), the visual goal dictates which drawer of a cabinet holds a tape roll and tape measure; to satisfy the goal, the policy selectively pulls open either the upper or the middle drawer before inserting both items. In panels~(c, d), swapping the visual arrangement of a glue stick and two pens between adjacent holders prompts the robot to invert its placement sequence accordingly. These comparisons demonstrate that post-trained policies remain tightly conditioned on visual context, even when realizing the target state requires prerequisite interactions with articulated mechanisms.

\paragraph{Cross-embodiment in-context learning from human goal images.}
Because HumanWAM-Goal is co-trained on heterogeneous human and robot corpora within a unified visuomotor framework, we hypothesize that it learns an embodiment-agnostic representation of scene structure and manipulation outcomes. To test this hypothesis directly, we supply the post-trained robot policy with \textbf{egocentric human goal images}—in which human hands, rather than robot grippers, are visible alongside the desired object configuration—and evaluate whether the robot can ground and execute the specified arrangement without parameter updates.

\begin{figure}[p]
    \centering
    \includegraphics[width=\linewidth,height=0.83\textheight,keepaspectratio]{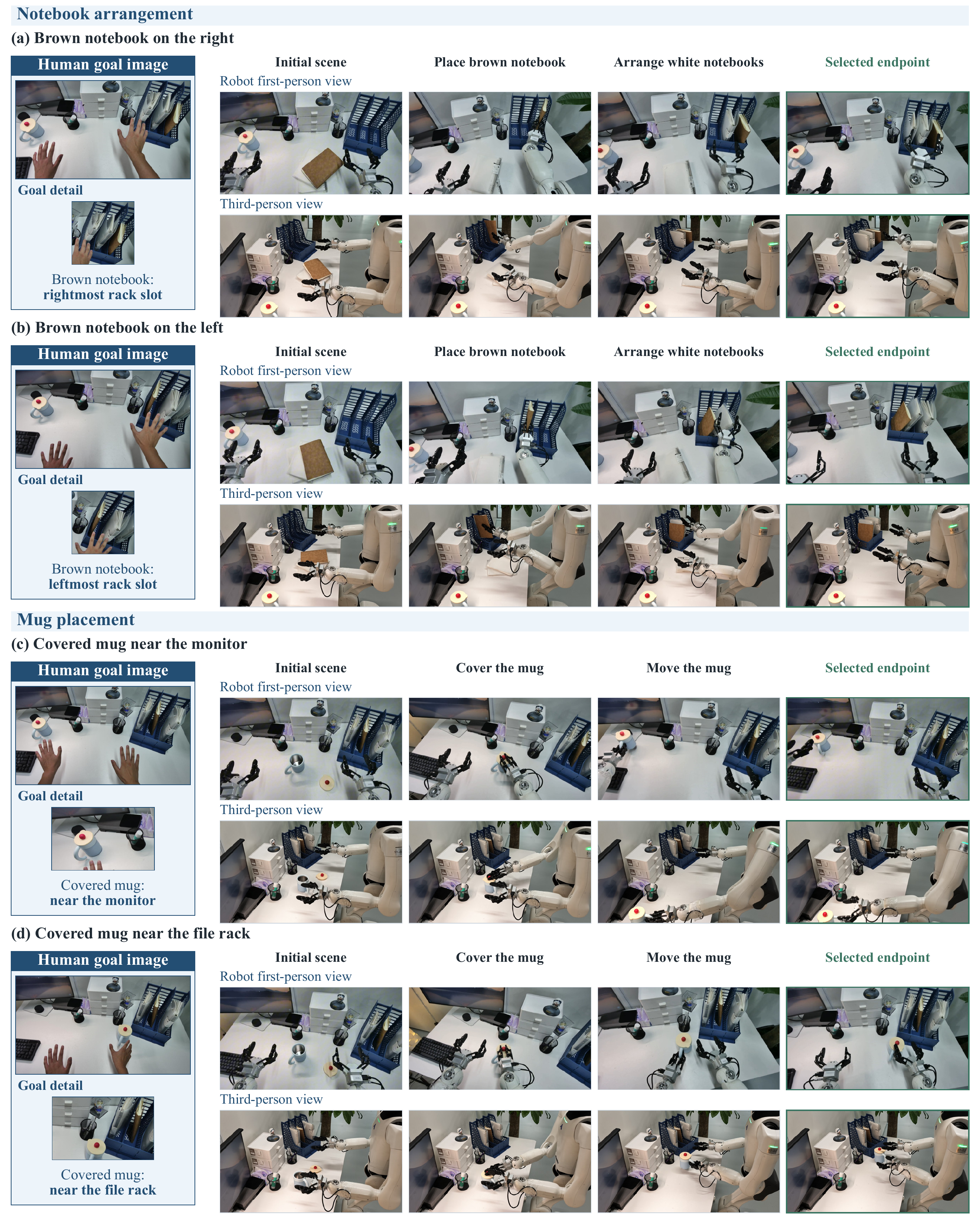}
   \caption{\textbf{Cross-embodiment in-context learning from egocentric human goal images.}
    Each panel pairs a human-demonstrated goal image and enlarged detail view (left, showing human hands) with synchronized first-person and third-person robot execution frames (right). Panels (a, b) direct the robot to file the brown notebook into the rightmost or leftmost rack slot, respectively. Panels (c, d) specify placing the covered mug near the monitor or beside the file rack. Green borders highlight the completed placements upon gripper release.}
    \label{fig:real_robot_human_goal_icl}
\end{figure}

As shown in~\cref{fig:real_robot_human_goal_icl}, despite the visual domain gap and the presence of human hands in the conditioning images, the policy accurately interprets the underlying spatial relations and adapts its execution across both paired comparisons: it slots the brown notebook into either the rightmost or leftmost compartment before arranging the white notebooks (panels~a vs.~b), and it caps the mug before transporting it to either the monitor or the file rack (panels~c vs.~d). These results are consistent with the hypothesis that human--robot co-training facilitates the transfer of goal semantics across embodiments.

\subsection{Long-Horizon Execution}
\label{ssec:real_world_long_horizon}

While the paired-goal comparisons in \cref{ssec:goal_based_icl} isolate immediate responsiveness to visual context, real-world deployment often requires chaining diverse manipulation skills over extended horizons. In such settings, each action alters the scene state for subsequent subgoals, requiring the policy to transition smoothly across subtasks without disturbing previously completed arrangements. We evaluate HumanWAM-Goal after task-specific post-training across two representative long-horizon workflows: multi-stage desk organization within a shared tabletop workspace and mobile household tidying across multiple work areas.

\begin{figure}[p]
    \centering
    \includegraphics[width=\linewidth,height=0.85\textheight,keepaspectratio]{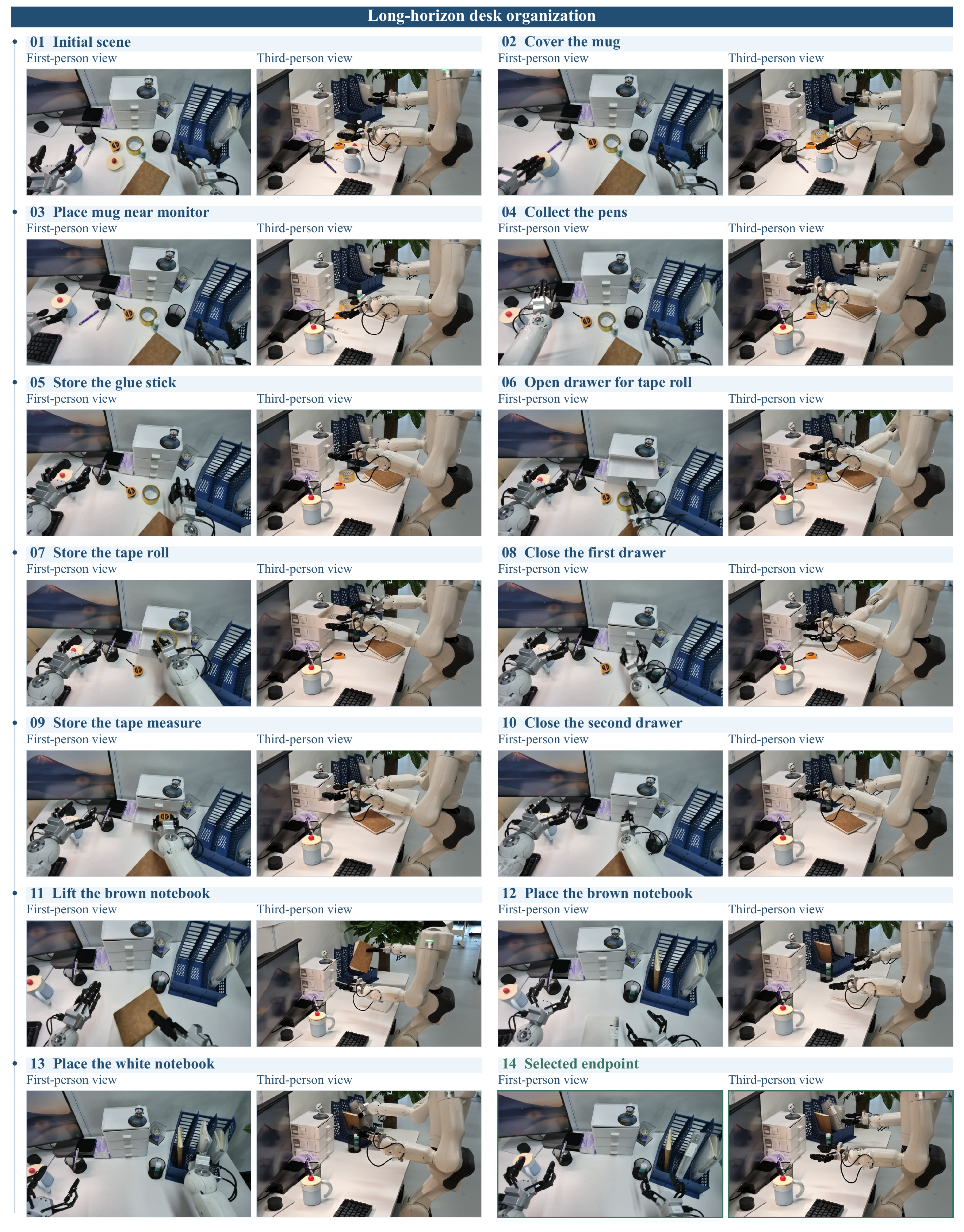}
    \caption{\textbf{Long-horizon desk organization with HumanWAM-Goal.}
    Fourteen chronologically ordered stages from a single continuous rollout, shown through paired first-person (left) and third-person (right) views. The trajectory sequentially executes mug capping and placement, pen and glue-stick sorting, articulated drawer storage of a tape roll and tape measure, and notebook filing. Green borders mark the terminal state after the final notebook is released into the rack.}
    \label{fig:real_robot_long_horizon_desk}
\end{figure}

\paragraph{Multi-stage tabletop organization.}
\Cref{fig:real_robot_long_horizon_desk} illustrates a continuous 14-stage rollout on a cluttered office desk. Across the trajectory, the policy coordinates distinct manipulation primitives—capping and relocating a mug (stages 01--03), sorting small stationery into holders (04--05), opening and closing two separate drawers to store a tape roll and tape measure (06--10), and filing notebooks into a vertical rack (11--14). Because all operations unfold within a confined workspace, later actions must navigate around earlier placements: the capped mug remains undisturbed near the monitor, and closed drawers stay shut while the robot manipulates adjacent objects. This rollout demonstrates coherent skill chaining across rigid-object placement, container insertion, and articulated-object manipulation.

\begin{figure}[p]
    \centering
    \includegraphics[width=\linewidth,height=0.88\textheight,keepaspectratio]{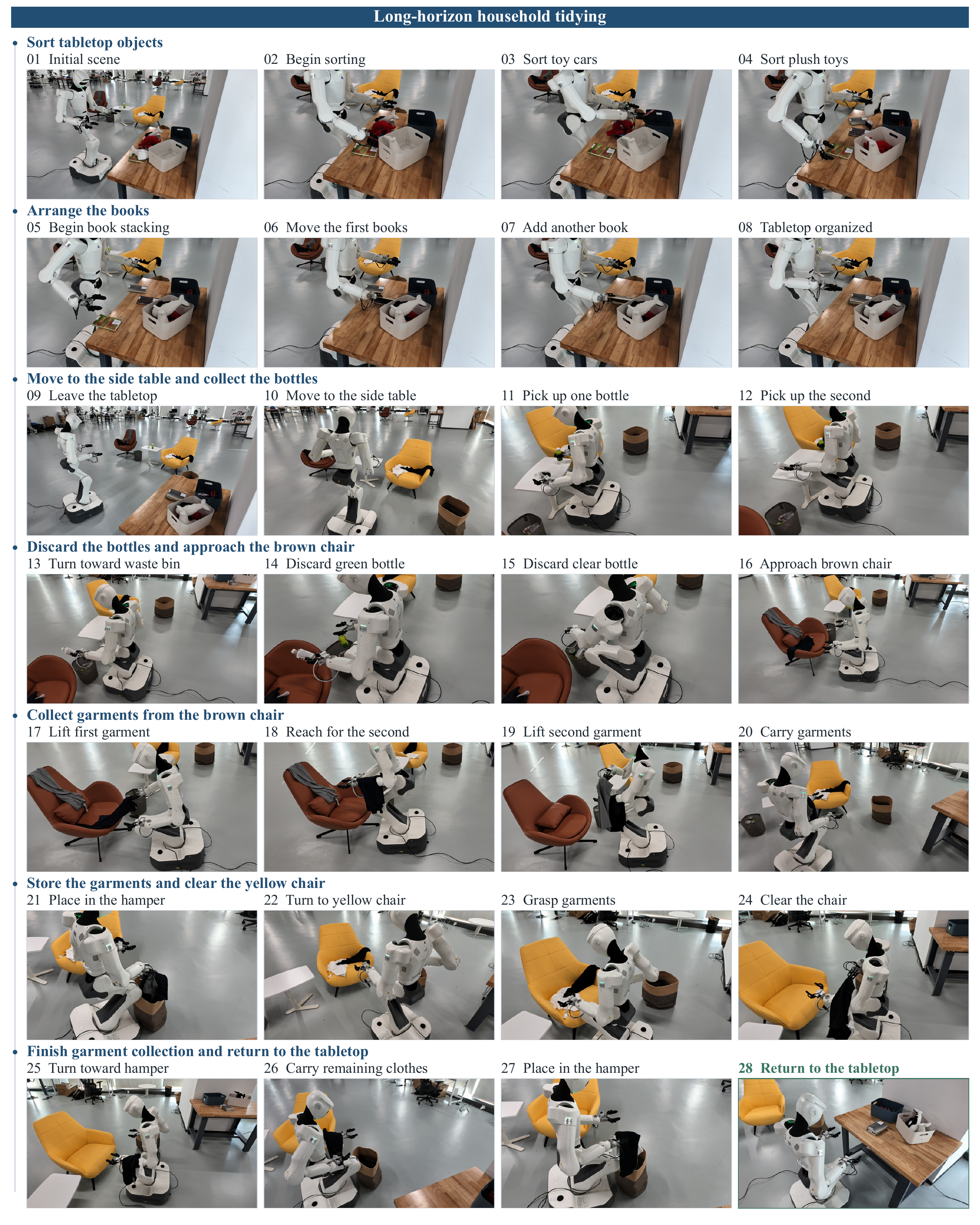}
    \caption{\textbf{Long-horizon mobile household tidying with HumanWAM-Goal.}
    Twenty-eight chronologically ordered keyframes from a single continuous rollout spanning multiple work areas: sorting tabletop toys and books, navigating to a side table to collect and discard empty bottles, gathering garments from two chairs into a laundry hamper, and returning to the main table.}
    \label{fig:real_robot_long_horizon_household}
\end{figure}

\paragraph{Multi-area mobile household tidying.}
\Cref{fig:real_robot_long_horizon_household} extends evaluation from a stationary tabletop to room-scale mobile manipulation on Astribot S1. Within a single uninterrupted sequence, the policy sorts toys into bins and stacks books at the main table (keyframes 01--08), navigates to a side table to collect and discard two plastic bottles into a waste bin (09--15), traverses to two chairs to gather deformable garments into a laundry hamper (16--27), and finally returns to the organized tabletop (28). By coupling base navigation with bimanual control in the unified 49-D hybrid action space (\cref{ssec:extended_action}), the model transitions seamlessly across spatial zones and object modalities—from rigid toys and bottles to deformable clothing. Together, these demonstrations show that HumanWAM-Goal can sustain temporally consistent, multi-skill execution across both dense tabletop layouts and extended mobile-manipulation workflows.
\section{Related Work}
\label{sec:related}

\subsection{World Action Models}
World--action models couple visual dynamics prediction with action generation, allowing robot policies to inherit manipulation priors from pretrained video models~\cite{gr1,ye2026worldactionmodelszeroshot}.
DreamZero~\cite{ye2026worldactionmodelszeroshot} learns control through joint video--action modeling, while LingBot-VA~\cite{li2026causal} integrates frame prediction and policy execution within an autoregressive diffusion framework.
The role of future visual information in action prediction remains an important architectural consideration~\cite{openwam,imagewam}.
Fast-WAM~\cite{yuan2026fastwamworldactionmodels} retains video prediction during training but omits future video generation during inference.
OpenWAM~\cite{openwam} systematically studies world--action information flow and embodied pretraining, demonstrating the benefits of explicit world-to-action interaction and synchronized joint denoising.
InternW0-$\Delta$~\cite{internw0delta} instead uses training-only future supervision to learn predictive representations for action generation without video rollout during inference.
Extending world--action modeling to compositional manipulation, ViGAR~\cite{gong2026rethinking} combines visual subgoal planning with goal-conditioned execution.
Our work complements these architectural developments by studying the effects of egocentric data scaling across different world--action configurations.

\subsection{Pure Egocentric Pretraining}
Pretraining on human manipulation provides visual representations and action priors for robot learning before embodiment-specific adaptation~\cite{r3m,egomimic}.
Being-H0~\cite{beingh0} develops human-motion pretraining for dexterous manipulation, while EgoScale~\cite{egoscale} scales action-labeled egocentric pretraining to over 20,000 hours and reports a log-linear relationship between data scale and validation loss.
EgoScale subsequently introduces aligned human--robot mid-training to facilitate embodiment transfer.
HumanScale~\cite{humanscale} establishes an empirical scaling law relating egocentric pretraining hours to validation action loss after robot post-training and finds that human data can outperform robot data at matched pretraining durations.
Dyna-2~\cite{dyna2} extends human-video pretraining to one million hours, reporting scaling in zero-shot action prediction on unseen robot data alongside improvements in post-trained robot task performance.
Building on these findings, Open-HumanWAM systematically evaluates egocentric scaling through closed-loop policy success under both in-distribution and out-of-distribution conditions.
Beyond data duration, our controlled comparisons establish diversity as an additional scaling dimension and quantify gains in the sample efficiency of robot adaptation.

\subsection{Zero-Shot Generalization and In-Context Learning}
Generalist robot policies increasingly extend beyond task-specific adaptation toward direct execution in unfamiliar tasks, environments, and embodiments~\cite{intelligence2025pi_,ye2026worldactionmodelszeroshot}.
Contextual guidance provides a complementary mechanism for specifying unfamiliar behaviors without task-specific parameter updates.
$\pi_{0.7}$~\cite{pi07} incorporates language, behavioral metadata, and visual subgoals into multimodal prompts, enabling compositional generalization and flexible control over task execution.
Demonstration-based in-context learning additionally conditions behavior on examples and interaction history~\cite{icrt,liu2026roboicl}.
RoboICL~\cite{liu2026roboicl} combines demonstration context with accumulated interaction memory while keeping model parameters fixed.
Our HumanWAM-Goal experiments connect these capabilities to the scale of human experience.
With robot training data held fixed, we quantify the effect of increasing egocentric data on zero-shot robot performance and qualitatively evaluate visual-goal conditioning before and after robot post-training.

\section{Conclusion}
\label{sec:conclusion}
We presented Open-HumanWAM, a fully open recipe for scaling robot learning with egocentric human experience. Our study of pure egocentric pretraining from 100 to 50,000 hours reveals an empirical log-linear scaling law in downstream out-of-distribution success. Expanding task and scene diversity strengthens this scaling trend, while larger-scale pretraining improves the sample efficiency of robot adaptation. Controlled experiments further demonstrate the value of human motion prediction and cross-source action alignment. Following this recipe, HumanWAM-Vanilla achieves competitive performance in simulation and improved robustness to visual distribution shifts on physical robots after robot post-training. Our release of annotated data, training code, checkpoints, and evaluation protocols provides a reproducible foundation for further research on egocentric scaling.

HumanWAM-Goal extends this exploration to joint human--robot co-training with visual-goal conditioning, improving zero-shot transfer in simulation and achieving state-of-the-art average success on RoboTwin 2.0. On physical robots, hybrid scaling with robot-balanced sampling boosts zero-shot success and expands task coverage without target-task demonstrations. Visual goals also enable embodied in-context learning with frozen parameters, steering policies across spatial and compositional variations as well as cross-embodiment human goal images after post-training. Finally, post-trained policies support coherent subgoal chaining across long-horizon desk organization and room-scale mobile household tidying.

\clearpage

\bibliographystyle{plainnat}
\setlength{\bibhang}{0pt}
\setlength\bibindent{0pt}
\bibliography{main}

\clearpage
\beginappendix

\lstdefinestyle{prompt}{
  basicstyle=\ttfamily\scriptsize,
  breaklines=true,
  breakatwhitespace=true,
  breakindent=0pt,
  columns=fullflexible,
  keepspaces=true,
  frame=single,
  framesep=4pt,
  xleftmargin=5pt,
  xrightmargin=5pt,
  aboveskip=0.8\baselineskip,
  belowskip=0.8\baselineskip,
  rulecolor=\color{black!30},
  backgroundcolor=\color{black!3},
  showstringspaces=false,
  escapechar=@,
}
\providecommand{\promptnote}[1]{\textcolor{black!55}{\textit{#1}}}

\section{Prompts for Goal Image Synthesis and Review}
\label{app:goal_prompt}

RoboTwin provides no reference goal images, so we synthesize one head-camera goal image for each scene. The only scene-specific inputs are the initial head-camera observation and the task instruction; no expert trajectory, terminal frame, or simulator state is used. Synthesis has two stages: a grounding stage localizes the task-relevant objects in the initial image, and an editing stage turns the initial observation into a completed-task observation. Each newly generated evaluation goal image is then reviewed and, if necessary, repaired.

\subsection{Pipeline Overview}

\paragraph{Inputs and resolution.}
The initial head image is letterboxed to $288{\times}224$ pixels (width$\times$height) and upsampled $4\times$ to $1152{\times}896$ for upload. GPT Image~2.5 generates a $1152{\times}896$ image at medium quality, which we downsample to $288{\times}224$ before restoring the original black letterbox pixels. The editing request contains three images: (1)~the upsampled initial head view, (2)~a task-independent robot appearance sheet, and (3)~a pixel-registered overlay of the initial view that marks the grounded objects.

\paragraph{Task-specific reference annotations.}
For each of the 50 RoboTwin tasks, we manually wrote one reference annotation on a reference scene. It lists the objects (identifier, description, bounding box, and whether the object moves or stays fixed), a natural-language description of the required final state, and visual acceptance checks. A few tasks also include optional geometric guides: hinge anchor lines for articulated objects, final-extent outlines, relocation cut-outs, or edit masks. These 50 annotations are fixed across all experiments; Stage~1 adapts them to each new scene.

\subsection{Stage 1: Scene-Specific Grounding}
For a new scene, a reasoning model (GPT-5.5, high reasoning effort) adapts the task's reference annotation to the current initial image and returns structured JSON with a fixed schema. Its inputs are the reference initial image, the current initial image, a copy of the current image overlaid with a pixel-coordinate grid, the reference instruction and annotation, the current instruction, and RoboTwin's public task description. The prompt template is:

\begin{lstlisting}[style=prompt]
Adapt an initial-scene annotation to a DIFFERENT seed of the same RoboTwin task.
Return JSON only. Image 1 is the reference seed's INITIAL head view. Image 2 is
the CURRENT seed's INITIAL head view. Image 3 is exactly the current initial
view with a coordinate grid and rulers labelled in NATIVE pixels. No image is
a completed task. Images 1 and 2 are enlarged 4x from a 288x224 native canvas.
ALL output coordinates must use native 288x224 pixels. Use Image 3's numbered
grid to locate each edge accurately, then confirm on Image 2. Do not estimate
coordinates from a resized display: interpolate the supplied ruler numbers.

Identify actual current object instances, including distractors. Adapt every
object description, box, arm, color, source/destination, number, direction and
numeric coordinate in the plan and checks to Image 2 and CURRENT INSTRUCTION.
Do NOT copy reference coordinates, scene sides, colors or asset names blindly.
Use the reference's correction principles, not its seed-specific scene geometry.
Only explicitly manipulated objects are fixed=false; distractors are fixed=true.
Preserve its physical task semantics and minimal-change rules. In an ambiguous
noun (container/object), select the current instance matching the instruction,
not an unrelated colorful distractor. If instruction names objects or arms,
those names take precedence over the reference plan. Some benchmark instructions
mention ONLY THE FIRST SUBSTEP (for example 'grab the red block' for handover_block).
The TASK ID and reference's complete final-state plan still specify the whole
task: preserve the transfer and final placement, not merely that first grab.
Do not drop the final task relation because an abbreviated instruction omits it.
Ground solely in initial RGB, instruction and the example. Do not infer hidden
simulator poses, consult future states, or fabricate an observed terminal image.

Keep the same context method: registered initial labels; stationary anchors
for articulated devices; proposed final extents only when the reference has
those; unscaled initial-pixel relocation guides only when the reference has
those; edit masks only when the reference has those. Empty arrays for unused
features. Objects may differ in count/identity between seeds. Tight boxes must
cover actual visible objects. Coordinates are integer pixels within x=0..288,
y=0..224. Bounds [x0,y0,x1,y1] have positive width/height. Point coordinates use
x=0..287,y=0..223. No more than 16 objects.

For hinged devices, mark the ACTUAL stationary base edge/corners and original
hinge from Image 2, not an axis-aligned imagined laptop. Rotate only the lid/door
about that hinge. Preserve existing plain surfaces, never add a keyboard.
For relocation guides trace a tight polygon around the COMPLETE current target
silhouette in Image 2; propose an integer translation [dx,dy] placing that same
unscaled silhouette on the actual support. Do not include unrelated object or
large background areas in the cutout. Destination is the usable platform, not
its display/handle. Keep the fixed support unchanged; natural overhang is okay.
Masks must include the WHOLE source, whole destination silhouette and a margin
for contact/shadows; avoid editing unrelated background. All dimensions/diagonal
yaw remain those in Image 2. For goals requiring rotation or changing orientation,
use final extent outlines rather than a translation that contradicts the task.
For stacking, preserve dimensions and ordering specified in current instruction.
Write plan and acceptance_checks with current coordinates if helpful. Annotations
are proposals from initial pixels; never describe them as ground truth.

TASK: {task}
REFERENCE INSTRUCTION: {reference_instruction}
REFERENCE ANNOTATION JSON: {reference_annotation}
CURRENT INSTRUCTION (authoritative for named instances/arms): {instruction}
{public_task_block}
\end{lstlisting}

\noindent The placeholder \texttt{\{public\_task\_block\}} is the same public-task block that closes the Stage-2 prompt below. It contains RoboTwin's public task description and placeholder schema, followed by a few lines of reviewed qualitative guidance for the task.

\subsection{Stage 2: Goal Image Editing}
The editing prompt for GPT Image~2.5 is a fixed template filled with the current instruction and the Stage-1 output (\texttt{\{objects\}}, \texttt{\{plan\}}, and \texttt{\{acceptance\_checks\}}). Paragraphs marked with a bracketed tag are included only for tasks whose reference annotation provides the corresponding guide. Gray italic text is explanatory and is not part of the prompt.

\begin{lstlisting}[style=prompt]
Edit Image 1 into ONE successful terminal HEAD-CAMERA observation of this exact simulated robot scene.

IMAGE ROLES:
1. Initial head view: authoritative camera, framing, geometry, appearance and scene layout.
2. Robot appearance sheet: robot construction only; its poses/background are not task targets.
3. REGISTERED geometry overlay: exactly the same initial head image at exactly the same pixel coordinates and scale as Image 1. Magenta boxes identify movable objects; cyan boxes/lines anchor fixed objects or base edges. The boxes are initial identity bounds, NOT requested object sizes or final locations. All annotations are explanatory only: output NO text, boxes, panels or arrows.

INSTRUCTION: {instruction}

GROUNDED INITIAL OBJECTS (pixel boxes use Image 1's 288x224 coordinates):
{objects}
@\promptnote{(expands to one line per object: ``- ID: description; initial box [x0, y0, x1, y1].'')}@

REQUIRED FINAL STATE:
{plan}

CONSERVATION RULES:
Apply the smallest physical change that satisfies the final state. Preserve camera extrinsics/intrinsics, table, background, light, object identities/materials/scale, every distractor, and fixed destination geometry. Translate/rotate only explicitly movable objects. Preserve exact object count; remove moved objects from their old locations. Reconstruct only newly exposed surfaces, with the existing simulator style. Do not canonicalize, beautify, replace or enlarge objects to improve visibility. A broad/awkward noun in the instruction refers to the specifically identified visible object above.
For articulated tasks, preserve the stationary body's original projected corners and yaw. Rotate only the moving part around its existing hinge; never swivel the whole device toward the viewer. For a press, retain the device's position and orientation and show the requested arm's gripper contacting the actual control specified by the current instruction. A control may face sideways; do not add a top button or rotate the device to expose it. For a hold/lift, show a real grasp with clearance above the support; for place/stack, show supported objects at the destination; release only grasps no longer required by the public complete task, and retain required support-holding grasps. Preserve the two-arm robot's construction, connected kinematic chains and plausible reach. An idle arm can stay outside the head frame. Do not invent extra hands/arms or use a floating disconnected gripper.

Before rendering, mentally match each ID to its original instance, distinguish movable and fixed parts, and check count, final relation, support/contact and stationary edges. Occlusion is natural: do not move fixed objects or change the camera just to reveal them. Output only the completed scene, at the exact aspect ratio and padding of Image 1. No goal image or future state is provided.

GEOMETRY PRIORITY: Image 1 is already the requested 1152x896 output size. Edit it in place: preserve unedited pixels and their exact coordinates. Image 3 aligns pixel-for-pixel with it. Do not reframe, regenerate or regularize the scene. Object boxes above use native coordinates: multiply by FOUR for input/output pixels. Preserve actual object dimensions under the fixed camera and perspective. Do not shrink targets to fit destinations or shrink whole stacks. Move targets with their original shape/volume. Preserve the named arm's mechanical identity and its connection to its original fixed mount. An arm may reach across the table; target position and the edge where a cropped gripper appears do not define arm identity. Do not mirror the robot or exchange arms.

[final-extent guides] Image 3 also includes ORANGE final-extent guides: a manually proposed geometric layout derived from initial object dimensions and the instruction, NOT an observed future image. Match the original object's size and shape within this guide at its destination. These outlines are approximate projections, never additional objects. Remove all guide lines/text from the output. Do not treat an orange guide as an initial object or duplicate the target.

[relocation guides] IMPORTANT IMAGE-3 SIZE / POSE GUIDE: the orange-outlined target at the destination is an exact unscaled cutout of the initial object, translated to propose the final placement. It is synthetic guidance from initial pixels, NOT a future observation or an additional object. This guide makes the required size and yaw explicit: match that FULL-SIZE silhouette and pose, keep the destination mat fixed underneath, and remove the old source object. Generate a coherent final scene with natural contact/shadows, without orange outlines or text. Do not shrink the target to fit inside the mat.

FINAL IMAGE CHECKS -- every item must be visibly satisfied:
{acceptance_checks}
@\promptnote{(expands to one bullet per check)}@

[edit masks] The attached alpha mask restricts editing to transparent regions containing the old target and proposed destination. Keep opaque regions exactly unchanged. Render the completed object placement inside the editable regions; do not move fixed supports or resize targets to fit the mask. No robot arm needs to enter a completed released-placement scene.

PUBLIC OFFICIAL COMPLETE TASK -- authoritative over the historical reference example:
Task: {task}
Full description: {robotwin_full_description}
Official placeholder schema: {robotwin_schema}
Preserve the exact current official instruction, including explicit release, put-down and negation clauses. Only after recorded compatibility review may the complete task fill in omitted steps. An explicit contradictory instruction is not an omission and must not be silently overridden. Unresolved language conflicts require review before generation. Left/right arm names refer to the robot's mechanical arms and their fixed mounts, not the target's horizontal image position or the edge where a cropped gripper appears. Preserve each arm's connection to its own mount, including cross-table reaches; do not mirror the robot or exchange arm identities. The appearance sheet illustrates construction only. Official {A}/{B} placeholders and historical annotation object IDs A/B are separate namespaces: rebuild their role mapping from current initial RGB and instruction, never equate letters. Preserve current object identity, size and background. The public complete task takes precedence over any reference plan, fixed flag, mask or acceptance check that would prevent required motion, lifting, holding, transfer or placement. Do not keep a support stationary when this task requires lifting or moving that support. Do not depict an intermediate placement as the completed task. Use no expert future RGB, trajectory, future pose, simulator state or hidden instance information.

REVIEWED QUALITATIVE GOAL QUALITY GUIDANCE:
This is reviewed image guidance derived from existing public task semantics; it is not a new official requirement or an assertion that a generated image proves physical success. It must remain compatible with the exact current instruction. Success may occur earlier in a trajectory; do not replace explicit current-language actions with a guessed terminal rule.
{task_guidance}
@\promptnote{(expands to one bullet per guidance line)}@
\end{lstlisting}

\paragraph{Example.}
For \texttt{blocks\_ranking\_size} with the instruction \emph{``Align large block, medium block, and small block by size from left to right at the center.''}, the scene-specific parts of the editing prompt are:

\begin{lstlisting}[style=prompt]
GROUNDED INITIAL OBJECTS (pixel boxes use Image 1's 288x224 coordinates):
- L: large purple block; initial box [52, 78, 90, 116].
- M: medium light green block; initial box [230, 92, 265, 131].
- S: small magenta block; initial box [106, 105, 126, 130].

REQUIRED FINAL STATE:
Move the three blocks to a central tabletop row and align them by physical size from left to right: L, the large purple block, on the left; M, the medium light green block, in the middle; S, the small magenta block, on the right. Keep the blocks upright, preserve their current colors and dimensions, leave small gaps between them, do not stack them, and release all blocks after placement. The completed row should be centered on the table around the middle of the image rather than leaving M far to the right or S between L and M.

FINAL IMAGE CHECKS -- every item must be visibly satisfied:
- Exactly the three visible blocks are arranged at the table center in one horizontal row.
- Left-to-right order is largest to smallest: large purple L, medium light green M, then small magenta S.
- All blocks remain upright, unstacked, non-overlapping, and released on the tabletop.
- No unrelated scene elements are moved or newly introduced.

Full description: there are three blocks on the table, the color of the blocks is random, move the blocks to the center of the table, and arrange them from largest to smallest, from left to right

REVIEWED QUALITATIVE GOAL QUALITY GUIDANCE:
- Identify large, medium and small using their true current physical dimensions, not color. Place the same three blocks in a central tabletop row with largest left, medium middle and smallest right, preserving all dimensions and colors.
- Small-then-medium-then-large wording specifies handling order; it does not reverse the final left-to-right size order given by full_description. Follow named per-block arms and depict completed released placement rather than stacking or holding a block.
\end{lstlisting}

\subsection{Goal Review and Repair}
\label{app:goal_review}
ChatGPT Astra High reviews every newly generated evaluation goal image. For each case it sees the initial and generated head images at native resolution, the instruction, the public task description, and the Stage-1 grounding. It treats the grounding as fallible and locates objects independently in the initial image, then returns \emph{pass}, \emph{fail}, or \emph{uncertain} with case-specific evidence. The review rubric is:

\begin{lstlisting}[style=prompt]
Check target/destination identity, exact instructed goal relation/arm when visually distinguishable, source removal/object counts, appearance/size, fixed background/destination/camera, real grasp/contact/support, articulated hinge, and coherent robot. Public full task only fills omissions compatibly, never overrides explicit instruction. Natural occlusion is legitimate. Do not demand an entire trajectory in a single goal frame, invent new acceptance rules, equate image-side with mechanical arm, or mark all cropped mounts uncertain. Pass means visible state is consistent with no clear defect; fail means concrete visible contradiction; uncertain identifies concrete unresolved evidence.

The audit concerns whether the visible generated goal is compatible with the exact task under the fixed head camera; it does NOT establish hidden physical state or execution history. If the requested endpoint is naturally outside the view or inside a clipped container, all relevant visible changes and fixed geometry are consistent, and no visible contradiction exists, use pass with an explicit visibility limitation. Retain uncertain when there is a specific visible ambiguity (which physical object, actual contact, orientation, scale) rather than merely unverifiable hidden physics.
\end{lstlisting}

\paragraph{Repair.}
Each non-passing verdict (fail or uncertain) comes with repair feedback and a repair stage. If the grounding misidentified the source, receiver, or contact region, the reviewer supplies a corrected target from the initial image and Stage~1 is rerun with this feedback. Otherwise, Stage~2 is rerun with the feedback inserted before the task block, and the grounding is kept. Every replacement image is reviewed again. Regenerated images are judged under a looser criterion than first-pass images: the reviewer accepts reasonable perspective variation and mild size differences that can be explained by projection, but still rejects wrong targets or receivers, duplicated or missing objects, wrong task relations, severe deformation, invented fixtures, and disconnected or extra arms.

\paragraph{Review statistics.}
Of the 7{,}032 evaluation goal images newly generated with this pipeline, 6{,}380 (90.7\%) passed the first review, 542 (7.7\%) failed, and 110 (1.6\%) were uncertain (Table~\ref{tab:goal_review}). After repair, all 7{,}032 passed review; repaired images were judged under the looser criterion described above. The review checks visual and instructional consistency only; it does not verify physical feasibility or task success in simulation. The remaining 2{,}968 of the 10{,}000 evaluation goals are earlier images that a separate audit approved for reuse.

\begin{table}[t]
\centering
\small
\caption{First-pass review outcomes for the 7{,}032 newly generated evaluation goal images.}
\label{tab:goal_review}
\begin{tabular}{lrr}
\toprule
Verdict & Images & Share \\
\midrule
Pass      & 6{,}380 & 90.7\%  \\
Fail      & 542     & 7.7\%   \\
Uncertain & 110     & 1.6\%   \\
\midrule
Total     & 7{,}032 & 100.0\% \\
\bottomrule
\end{tabular}
\end{table}
\clearpage

\section{Real-World Zero-Shot Task Suite}
\label{app:real_world_tasks}

Table~\ref{tab:real_world_tasks} specifies the 20 real-world tasks and their scoring rubrics. Each checkpoint is evaluated on 10 trials per task (200 rollouts), without target-task demonstrations or parameter updates. Tasks are specified by goal images with the shared instruction \textit{``Complete the task shown in the goal image.''}

\paragraph{Scoring conventions.}
Each task has three or four scored milestones. Parenthesized values are maximum stage scores and sum to 10; intermediate scores retain the partial credit recorded during evaluation. Binary success is assessed separately using the task outcome in the last column, and need not imply a score of 10. For example, the records deduct one point for using a different arm from the goal image (E1), and one or two points for a tipped cup or a lid grasp, respectively (H3), despite marking these trials successful. Both metrics are averaged over trials and then equally over tasks.

\begin{table*}[h]
  \centering
  \caption{\textbf{Real-world zero-shot tasks and scoring rubrics.}
  Milestones are listed in the order recorded in the evaluation sheets;
  arrows do not impose a universal prerequisite rule. Parentheses give
  maximum points, not cumulative scores. E1--E10 are easy tasks;
  H1--H10 are hard tasks.}
  \label{tab:real_world_tasks}
  \begingroup
  \footnotesize
  \setlength{\tabcolsep}{4pt}
  \renewcommand{\arraystretch}{1.18}
  \newcommand{\rwstep}{\ensuremath{\,\rightarrow\,}}
  \begin{tabularx}{\textwidth}{@{}l >{\raggedright\arraybackslash}p{0.205\textwidth} c >{\raggedright\arraybackslash}p{0.350\textwidth} >{\raggedright\arraybackslash}X@{}}
    \toprule
    \textbf{ID} & \textbf{Task} & \textbf{Family} & \textbf{Scored milestones (maximum points)} & \textbf{Task success outcome} \\
    \midrule
    \multicolumn{5}{@{}l}{\textbf{Easy tasks} \enspace (10 tasks; 100 trials)} \\
    \addlinespace[3pt]
    E1 & Lift a block & B & Approach (3) \rwstep grasp (4) \rwstep lift (3) & Target block held aloft. \\
    E2 & Compress a sponge & A & Approach (2) \rwstep contact (3) \rwstep press (5) & Sponge fully compressed. \\
    E3 & Open a hardcover book & C & Approach (2) \rwstep touch (2) \rwstep grasp (3) \rwstep open (3) & Front cover opened. \\
    E4 & Ring a desk bell & A & Approach (3) \rwstep contact (3) \rwstep press (4) & Bell actuated to ring. \\
    E5 & Press a click pen & A & Approach (3) \rwstep contact (3) \rwstep press (4) & Pen's push button depressed. \\
    E6 & Open a drawer & C & Contact (3) \rwstep grasp (4) \rwstep pull open (3) & Drawer pulled open. \\
    E7 & Push an object away & A & Contact (3) \rwstep manipulate (3) \rwstep complete (4) & Object displaced from its initial position. \\
    E8 & Lift and shake a bottle & B & Locate (3) \rwstep grasp (3) \rwstep lift (4) & Bottle lifted and shaken. \\
    E9 & Arrange blocks & B & Grasp (2) \rwstep place (3) \rwstep complete (5) & Blocks in the target arrangement. \\
    E10 & Place a block on a tray & D & Grasp (3) \rwstep place (3) \rwstep complete (4) & Block on the tray. \\
    \midrule
    \multicolumn{5}{@{}l}{\textbf{Hard tasks} \enspace (10 tasks; 100 trials)} \\
    \addlinespace[3pt]
    H1 & Stack blocks & D & Grasp (3) \rwstep place (3) \rwstep complete (4) & One block stacked on another. \\
    H2 & Place a can beside a book & D & Grasp (3) \rwstep place (3) \rwstep complete (4) & Can beside the book. \\
    H3 & Place a cup on a coaster & D & Grasp (3) \rwstep place (3) \rwstep complete (4) & Cup on the coaster. \\
    H4 & Place an egg in a bowl & D & Grasp (3) \rwstep place (3) \rwstep complete (4) & Egg inside the bowl. \\
    H5 & Place an object on a book stack & D & Grasp (3) \rwstep place (3) \rwstep complete (4) & Object on top of the book stack. \\
    H6 & Place a glue stick in a pen holder & B & Grasp (3) \rwstep place (3) \rwstep complete (4) & Glue stick inside the pen holder. \\
    H7 & Stack two bowls & E & Grasp (3) \rwstep place (3) \rwstep complete (4) & One bowl stacked on the other. \\
    H8 & Lift two bottles bimanually & B & Grasp left (2) \rwstep lift left (3)\newline \rwstep grasp right (2) \rwstep lift right (3) & Both bottles lifted, one per gripper. \\
    H9 & Open a drawer and place an object inside & C & Grasp drawer (3) \rwstep open drawer (3)\newline \rwstep grasp object (2) \rwstep place (2) & Drawer open with the object inside. \\
    H10 & Insert a plug into a power strip & E & Approach plug (2) \rwstep grasp (3) \rwstep insert (5) & Plug inserted into the target socket. \\
    \bottomrule
  \end{tabularx}
  \par\smallskip
  \begin{minipage}{\textwidth}
    \footnotesize
    \textit{Families:} A, contact/state changes (4); B, grasping/transport (5);
    C, articulated opening/closing (3); D, spatial placement/alignment (6);
    E, stacking/insertion (2). Family assignments and milestone labels follow
    the task records; overlapping skills are represented by one assigned family.
  \end{minipage}
  \endgroup
\end{table*}

\clearpage

\let\clearpage\relax 
\end{document}